\documentclass{IEEE-Con-Sys-mag}

\newcommand{\Ad}{\mathrm{Ad}}
\newcommand{\ad}{\mathrm{ad}}

\usepackage{amsmath,amssymb}
\usepackage{booktabs}
\usepackage{colortbl}
\usepackage{xcolor}
\usepackage{array}
\usepackage{algorithm}
\usepackage{algorithmic}
\usepackage{tikz}
\usepackage[most]{tcolorbox}
\usepackage{threeparttable}
\usepackage{etoolbox}

\usetikzlibrary{arrows.meta,positioning,calc}
\definecolor{figblue}{RGB}{55,100,165}
\makeatletter
\def\@IEEEproof[#1]{%
  \par\addvspace{12pt \@plus2pt \@minus2pt}%
  \noindent{\sffamily\itshape\fontsize{9pt}{10.5pt}\selectfont #1~}\par\noindent%
}
\makeatother

\usepackage[most]{tcolorbox}
\usepackage{algorithmic}
\usepackage{xcolor}

\definecolor{csmblue}{RGB}{0,92,184}
\definecolor{csmback}{RGB}{246,248,252}
\definecolor{csmhead}{RGB}{232,238,247}

\newcounter{algobox}
\renewcommand{\thealgobox}{\arabic{algobox}}

\newtcolorbox{algoboxbase}[2][]{%
  enhanced,
  colback=csmback,
  colframe=csmblue,
  boxrule=0.6pt,
  arc=2mm,
  left=2mm,
  right=2mm,
  top=2mm,
  bottom=2mm,
  title={Algorithm~\thealgobox: #2},
  fonttitle=\bfseries,
  colbacktitle=csmhead,
  coltitle=black,
  titlerule=0pt,
  title filled,
  toptitle=1mm,
  bottomtitle=1mm,
  borderline={0.6pt}{0pt}{csmblue},
  #1
}

\newenvironment{algobox}[2][]{%
  \refstepcounter{algobox}%
  \begin{algoboxbase}[#1]{#2}%
}{%
  \end{algoboxbase}%
}

\usepackage{etoolbox}

\newcounter{sidebarID}

\AtBeginEnvironment{sidebar}{\stepcounter{sidebarID}}

\usepackage{threeparttable}
\definecolor{CSMblue}{RGB}{220,230,245}   
\definecolor{CSMorange}{RGB}{255,239,214} 
\definecolor{CSMgray}{RGB}{245,245,245}   

\jvol{XX}
\jnum{XX}
\paper{8}
\jmonth{June}
\jname{IEEE CONTROL SYSTEMS}
\pubyear{2020}
\newcommand{\SE}{\mathbb{SE}_2(3)}
\newcommand{\SEE}{\mathbb{SE}_5(3)}
\newcommand{\SO}{\mathbb{SO}(3)}
\newcommand{\so}{\mathfrak{so}(3)}
\newcommand{\se}{\mathfrak{se}_2(3)}
\newcommand{\BAd}{\mathbf{Ad}}
\newcommand{\Bad}{\mathbf{ad}}
\renewcommand{\exp}{\mathop{\mathbf{exp}}\nolimits}
\renewcommand{\log}{\mathop{\mathbf{log}}\nolimits}
\title{Tutorial on Aided Inertial Navigation Systems\\
\stitle{A Modern Treatment Using Lie-Group Theoretical Methods}}
\author{{S}oulaimane Berkane}
\affil{}
\makeatletter
\providecommand{\headerps@out}{}
\makeatother
\begin{document}
\begin{titlepage}
    \centering
    
    {\Large \textbf{LaRSA Technical Report}\par}
    \vspace{0.5cm}
    {\large Robotics and Autonomous Systems Laboratory\par}
    {\large Universit\'e du Qu\'ebec en Outaouais\par}
    
    \vspace{1.5cm}
    
    {\huge \textbf{Tutorial on Aided Inertial Navigation Systems}\par}
    \vspace{0.4cm}
    {\Large A Modern Treatment Using Lie-Group Theoretical Methods\par}
    
    \vspace{1.2cm}
    
    {\large Soulaimane Berkane\par}
    
    \vspace{1cm}
    
    \begin{tabular}{rl}
        \textbf{Report number:} & LaRSA-TR-2026-01 \\
        \textbf{Date:} & October 2026 \\
        \textbf{Version:} & arXiv submission version 3 \\
        \textbf{Affiliation:} & Universit\'e du Qu\'ebec en Outaouais \\
        \textbf{Laboratory:} & LaRSA -- Robotics and Autonomous Systems Laboratory \\
        \textbf{Contact:} & soulaimane.berkane@uqo.ca \\
    \end{tabular}
    
    \vfill
    
    \begin{minipage}{0.9\textwidth}
    \textbf{Abstract.}
    This technical report contains the arXiv version of the manuscript entitled
    \emph{``Tutorial on Aided Inertial Navigation Systems: A Modern Treatment Using
Lie-Group Theoretical Methods''}. It is circulated as an internal LaRSA report
    for dissemination and archival purposes.
    \end{minipage}
    
    \vspace{0.8cm}
    
    \begin{minipage}{0.9\textwidth}
    \textbf{Note.}
    The pages that follow reproduce the manuscript itself.
    \par\medskip
    \textbf{Code.}
    MATLAB code for reproducing the numerical results is available at
https://github.com/sberkane/ins-tutorial (version 1.0.0).
    \end{minipage}
    
    \vfill
    
    {\small (c) 2026 Soulaimane Berkane / LaRSA, UQO\par}
\end{titlepage}

\setcounter{page}{1}
\makeatletter

\def\ps@headings{%
  \def\@oddhead{}%
  \def\@evenhead{}%
  \def\@oddfoot{\hfill\thepage}%
  \def\@evenfoot{\thepage\hfill}%
}

\def\ps@plain{%
  \def\@oddhead{}%
  \def\@evenhead{}%
  \def\@oddfoot{\hfill\thepage}%
  \def\@evenfoot{\thepage\hfill}%
}

\def\ps@tpage{%
  \def\@oddhead{}%
  \def\@evenhead{}%
  \def\@oddfoot{\hfill\thepage}%
  \def\@evenfoot{\thepage\hfill}%
}

\pagestyle{headings}
\thispagestyle{plain}
\renewcommand{\dois}[2]{}
\makeatother
\maketitle

\dois{}{}

\begin{sidebar}{Matrix Lie Groups}
\label{sidebar-MatrixLieGroup}
Material related to this sidebar can be found in standard textbook references
such as~[S1]. A \emph{group} is a pair $(\mathcal{G},\circ)$, where $\mathcal{G}$ is a set and
$\circ:\mathcal{G}\times\mathcal{G}\to\mathcal{G}$ is a binary operation such that,
for all $g_1,g_2,g_3\in\mathcal{G}$, the following properties hold:
\begin{itemize}
\item \emph{Closure:} $g_1\circ g_2\in\mathcal{G}$.
\item \emph{Associativity:} $(g_1\circ g_2)\circ g_3=g_1\circ(g_2\circ g_3)$.
\item \emph{Identity:} there exists $e\in\mathcal{G}$ with $e\circ g=g\circ e=g$ for all $g\in\mathcal{G}$.
\item \emph{Inverse:} for each $g\in\mathcal{G}$ there exists $g^{-1}\in\mathcal{G}$
such that $g\circ g^{-1}=g^{-1}\circ g=e$.
\end{itemize}
On the other hand, a \emph{smooth manifold} is a topological space that is locally homeomorphic to the Euclidean space $\mathbb{R}^n$ and equipped with a smooth atlas, which enables differentiation of maps between manifolds. Each point $g \in \mathcal{G}$ is associated with a tangent space $T_g\mathcal{G}$—a real vector space that intuitively represents all possible directions in which one can pass tangentially through $g$, whose elements are called tangent vectors.

Now, a \emph{Lie group} is a group $(\mathcal{G},\circ)$ that is also a smooth manifold,
such that the group operations (multiplication and inverse) are smooth maps.
This compatibility ensures that one can perform calculus on group-valued states.
The Lie algebra of a Lie group $\mathcal{G}$ is the tangent space
$\mathfrak{g}:=T_e\mathcal{G}$ at the identity element $e$, endowed with a bilinear operation
$[\cdot,\cdot]:\mathfrak{g}\times\mathfrak{g}\to\mathfrak{g}$ called the Lie bracket.
This operation is closed on $\mathfrak{g}$, bilinear, alternating
($[X,X]=0$), and satisfies the Jacobi identity; see~\cite{barfoot2024state}.

\textbf{Exponential and logarithm maps.}
For any $X\in\mathfrak{g}=T_e\mathcal{G}$, there exists a unique smooth curve
$\gamma_X:\mathbb{R}\to\mathcal{G}$ satisfying
\[
\gamma_X(0)=e, \qquad 
\gamma_X(t+s)=\gamma_X(t)\circ\gamma_X(s),
\qquad 
\dot\gamma_X(0)=X .
\]
Such a curve is called a \emph{one--parameter subgroup}.
The \emph{exponential map} is defined by
\[
\exp(X):=\gamma_X(1).
\]

The exponential map is smooth and locally invertible around the identity:
there exist neighborhoods $\mathcal{V}\subset\mathfrak{g}$ of $0$ and
$\mathcal{U}\subset\mathcal{G}$ of $e$ such that
$\exp:\mathcal{V}\to\mathcal{U}$ is a diffeomorphism.
Its local inverse is the \emph{logarithm map}
\[
\log:\mathcal{U}\to\mathcal{V}, \qquad \log(\exp(X))=X \ \text{for } X\in\mathcal{V}.
\]
Outside this neighborhood, $\log$ need not be uniquely defined.

\textbf{Matrix Lie group.}
A \emph{matrix Lie group} $\mathcal{G}$ is a Lie group that is a subgroup of the general linear group
\[
GL(n,\mathbb{R})
:=\{A\in\mathbb{R}^{n\times n}:\det(A)\neq 0\},
\]
viewed as an embedded submanifold of $\mathbb{R}^{n\times n}$.
The group operation is matrix multiplication, with identity element $I_n$.
Its Lie algebra is the matrix space
\[
\mathfrak{g}
=\{X\in\mathbb{R}^{n\times n}:\exp(tX)\in\mathcal{G}\ \forall t\in\mathbb{R}\},
\]
which coincides with the tangent space $T_e\mathcal{G}$ at the identity. The Lie bracket reduces to the matrix commutator
$[X,Y]=XY-YX$.

For matrix Lie groups, the exponential map 
$\exp:\mathfrak{g}\to\mathcal{G}$ coincides with the 
\emph{matrix exponential}, defined by the absolutely convergent power series
\[
\exp(X)
=\sum_{k=0}^{\infty}\frac{1}{k!}X^k.
\]

Conversely, the logarithm map (when defined locally around the identity)
coincides with a \emph{matrix logarithm}, i.e., a matrix $X$ satisfying
$\exp(X)=Y$. 
In a neighborhood of the identity, it admits the series expansion
\[
\log(Y)
=\sum_{k=1}^{\infty}\frac{(-1)^{k+1}}{k}(Y-I_n)^k,
\]
which converges whenever $\|Y-I_n\|<1$.

\end{sidebar}

\chapterinitial{I}nertial navigation systems (INS) form the backbone of modern positioning and navigation
solutions for autonomous vehicles, robotics, and defense applications. By integrating
measurements from accelerometers and gyroscopes, an INS can estimate the position,
velocity, and orientation of a moving body without relying on external infrastructure.
However, these estimates are inherently corrupted by sensor noise and biases, which
cause the navigation error to grow unbounded over time \cite{Woodman2007}. To mitigate this drift, inertial navigation systems (INS) are typically
\emph{aided} by incorporating measurements from external sensors within a
consistent inertial fusion framework. Such aiding sources may include
Global Navigation Satellite Systems (GNSS), magnetometers, cameras,
Light Detection and Ranging (LiDAR), radio-based ranging systems,
barometers, or other environment-dependent modalities~\cite{Titterton2004,Groves2013}.
The specific choice of aiding sensors is application-dependent and is
dictated by the operating environment, required accuracy, and system
constraints.

A central challenge in this fusion problem lies in the representation and estimation of
orientation. Unlike position or velocity, orientation evolves on a nonlinear manifold
associated with rotations \cite{Shuster1993}. A naive application of the extended Kalman filter (EKF),
based on additive corrections in Euler angles or quaternions, leads to well-known
difficulties such as singularities, normalization errors, and inconsistency in the
linearized error dynamics \cite{BarShalom2001,Crassidis2007}. These limitations motivated
the development of the multiplicative extended Kalman filter (MEKF), originally
introduced for spacecraft attitude estimation by Lefferts, Markley, and Shuster
\cite{Lefferts1982}. \begin{summary}
\summaryinitial{R}obots and autonomous vehicles rely on accurate estimates of their position, velocity, and orientation. Inertial sensors measure motion at high rates, but errors accumulate as these measurements are integrated. Satellite navigation, cameras, and other sensors help limit this drift. Combining these sources effectively requires estimation methods that account for the geometry of the navigation state.

This tutorial presents a unified introduction to aided inertial navigation from a Lie-group perspective. It connects motion prediction, uncertainty propagation, and measurement updates within a common geometric framework. Particular attention is given to invariant filtering, which uses system symmetries to simplify the description and evolution of estimation errors. The tutorial explains how choices of state and error representations shape the filter equations. It connects theoretical principles with formulas suitable for implementation and discusses recent extensions addressing convergence and sensor biases. By organizing established and emerging methods around their shared geometric foundations, the presentation provides students, researchers, and engineers with a basis for understanding, implementing, and extending navigation algorithms.
\end{summary}
The MEKF represents the nominal attitude using a unit quaternion,
while estimation errors are expressed multiplicatively through a minimal three-parameter
representation. This formulation preserves the unit-norm constraint automatically,
eliminates singularities, and significantly improves numerical robustness. Although
initially developed for attitude estimation, the MEKF was later extended to full inertial
navigation systems and remains a widely adopted approach in practice \cite{Wang2020}.

Despite these advantages, the MEKF retains a limitation shared with the classical EKF:
the linearized error dynamics depend on the estimated trajectory.
As a result, the filter relies on a local linearization that is valid only when
the estimation error remains sufficiently small.
Large initial errors, or prolonged periods of weak or intermittent aiding,
can therefore lead to inconsistency or even divergence \cite{Barrau2017}.
To address this issue,
researchers have increasingly exploited the underlying \emph{geometric structure} of
inertial navigation systems. Over the past two decades, a substantial body of work has
emerged on nonlinear observers and filters defined directly on matrix Lie groups (see ``\nameref{sidebar-MatrixLieGroup}'').
Among these approaches, the \emph{invariant extended Kalman filter (InvEKF)} has
established itself as a particularly powerful framework
\cite{Barrau2017,Bonnabel2007,Bonnabel2009,Barrau2018}.
By leveraging the \emph{group-affine structure} of the navigation dynamics, the InvEKF yields invariant error dynamics that are autonomous and independent
of the estimated trajectory. More remarkably, these error dynamics possess the
\emph{log-linear property}: when expressed in exponential coordinates on the Lie group,
the nonlinear error evolves according to an exact linear differential equation, rather
than merely a first-order linearization about the estimated trajectory.
This structural property fundamentally distinguishes the InvEKF from conventional
EKF-type formulations and contributes to its improved consistency and stability
properties. Closely related developments include
deterministic nonlinear observers with Lyapunov stability guarantees under various aiding
configurations and sensing modalities; see, e.g., \cite{Bryne2017,Wang2021,Berkane2021}.

More recently, symmetry-based filtering frameworks have been proposed to
extend these ideas beyond group-affine systems on Lie groups. In particular,
the \emph{equivariant filter} (EqF) framework
\cite{vanGoorHamelMahony2022} considers systems evolving on homogeneous
spaces equipped with a suitable symmetry. The EqF constructs a global
equivariant error whose local representation is linearized about a
\emph{fixed origin} of the error space, rather than about the time-varying
state estimate as in the classical EKF. Although some dependence on the
estimate generally remains through a symmetry-transformed input, the linearization no longer depends separately on the estimated state
and the system input, as in the classical EKF.

The InvEKF is recovered as a particular EqF realization when the state space
is itself a Lie group and the dynamics possess the group-affine property. In
that specialization, the stronger InvEKF properties---autonomous invariant
error dynamics and exact log-linear propagation---are recovered; these
properties are not generic to an EqF. The broader framework instead provides
additional freedom in the choice of symmetry, which is particularly useful
for augmented systems such as biased inertial navigation, where the classical
group-affine structure is lost. Moreover, when the measurement model is equivariant, the symmetry can also be exploited in the correction step to eliminate the second-order
output-linearization error, yielding an approximation with a third-order error. This
refined formulation, referred to as EqF$^\star$
\cite{vanGoorHamelMahony2022}, can significantly improve the transient
performance of the filter. Thus, the
EqF should be viewed not simply as a replacement for the InvEKF, but as a
broader framework that retains a symmetry-centered error linearization when
the stronger group-affine structure is unavailable, while recovering the
InvEKF in the latter case. Interested readers may consult \cite{MahonyAnnual2022} for a broader perspective on equivariant filtering and observer design.

The purpose of this tutorial is to provide a unified and accessible
control-oriented introduction to aided inertial navigation through a
Lie-group formulation centered on the \textit{extended Special Euclidean
group} $\SE$, first introduced in~\cite{Barrau2017}.
Rather than surveying a collection of estimation techniques, the tutorial
is organized around a single geometric backbone that captures the essential
structure of inertial navigation systems and their interaction with aiding
measurements.
Starting from $\SE$, the navigation equations, error representations, and
filtering mechanisms are developed in a form that is both rigorous and
implementation-oriented.

A key objective of this work is to provide a constructive
presentation of the main results developed within the $\SE$ framework and
to make the geometric mechanisms underlying them transparent and easy to
follow for readers less familiar with state estimation on Lie groups.
Several existing results are revisited and reformulated in a constructive
and unified manner, with an emphasis on clarity, geometric
interpretation, structural properties, and practical implementation.
Particular attention is given to showing how the underlying
group structure explains and connects results in invariant estimation,
discrete-time propagation, uncertainty propagation, and preintegration,
rather than presenting these developments as separate constructions.
Throughout, care is taken to preserve the system-theoretic foundations
required for rigorous analysis, while presenting the material at a level
suitable for use as supporting content in graduate-level courses on
inertial navigation and state estimation.

Building on this foundation, the tutorial discusses recent advances that
extend or enrich the $\SE$ formulation, including higher-order state
representations such as $\SEE$~\cite{Wang2021,Benahmed2025}, synchronous
observer designs~\cite{vanGoor2025Automatica}, and equivariant filtering
methods~\cite{Fornasier2025}.
These developments are presented as natural extensions of the same
geometric principles, highlighting how alternative constructions address
specific issues such as convergence guarantees and bias handling.
Throughout the tutorial, the focus is on conveying a coherent
system-theoretic perspective rather than cataloging methods or emphasizing
empirical performance.

\begin{pullquote}
A central challenge in this fusion problem lies in the representation and estimation of
orientation. 
\end{pullquote}

\begin{figure*}[t]
\centering
\begin{tcolorbox}[colback=red!10, arc=2mm, boxrule=0pt,
                  left=3mm, right=3mm, top=3mm, bottom=3mm]

\centering
\includegraphics[width=0.85\linewidth]{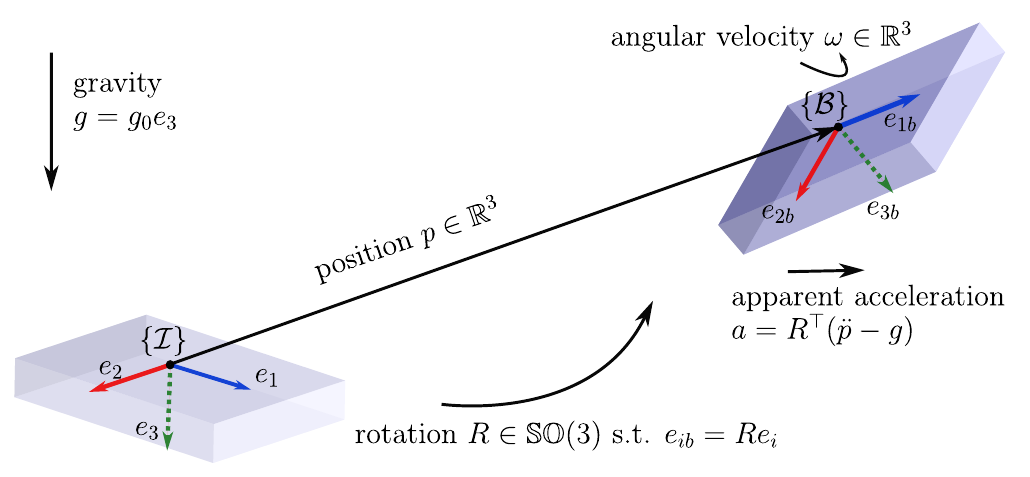}

\end{tcolorbox}

\vspace{4pt}

\caption{\textbf{INS kinematics setting.} 
The North--East--Down (NED) inertial frame $\{\mathcal I\}$ is defined by the basis 
$(e_1,e_2,e_3)$. The body-fixed IMU frame $\{\mathcal B\}$ has basis vectors 
$(e_{1b},e_{2b},e_{3b})$. The vehicle position expressed in $\{\mathcal I\}$ is $p$, 
and its orientation is represented by the rotation matrix $R \in \SO$, which maps 
body-frame vectors into the inertial frame.}
\label{fig:ins_kinematics}
\end{figure*}

\begin{sidebar}{Attitude Representation on $\SO$} 
\label{sidebar-Orientation}

\setcounter{sequation}{0}
\renewcommand{\thesequation}{S\arabic{sequation}}
\setcounter{stable}{0}
\renewcommand{\thestable}{S\arabic{stable}}
\setcounter{sfigure}{0}
\renewcommand{\thesfigure}{S\arabic{sfigure}}
\sdbarinitial{O}rientation is a central variable in inertial navigation and has been studied extensively for over half a century.  
Shuster’s seminal review~\cite{Shuster1993} offers a comprehensive comparison of classical attitude representations—Euler angles, direction cosine matrices (DCMs), and unit quaternions.  
Euler angles are intuitive but suffer from singularities (gimbal lock).  
Quaternions eliminate singularities and are numerically efficient, yet represent a \emph{double cover} of $\SO$, meaning they are not unique and must satisfy a unit-norm constraint.  
Rotation matrices (DCMs) are globally nonsingular and conceptually straightforward, though redundant (nine parameters for three degrees of freedom).  
In this tutorial, we adopt the \emph{group-theoretic view} of orientation through $\SO$, which offers a compact, globally valid, and geometrically consistent framework for analysis and estimation—a perspective now central to modern filtering and observer design.

Consider the \emph{North–East–Down (NED)} inertial frame $\{\mathcal I\}$ and the \emph{body frame} $\{\mathcal B\}$ attached to the moving platform (see Figure \ref{fig:ins_kinematics}). 
Let $\{e_i\}$ denote the canonical basis vectors of the inertial frame and $\{e_{ib}\}$ the axes of the body frame, expressed in inertial coordinates. 
The orientation of the body with respect to the inertial frame is described by a rotation matrix $R \in \SO$, satisfying
\begin{sequation}
e_{ib} = R\,e_i.
\end{sequation}
Hence, the columns of $R$ are simply the body-frame unit vectors written in inertial coordinates. 
Equivalently, for any vector $x$, its coordinates in the two frames are related by
\begin{sequation}
x^{\mathcal I} = R\,x^{\mathcal B}, 
\qquad \text{or inversely} \qquad 
x^{\mathcal B} = R^\top x^{\mathcal I}.
\end{sequation}
This relation clarifies the physical meaning of the rotation matrix and its role in transforming vectors between coordinate frames.

The set of all 3D rotation matrices forms the \emph{special orthogonal group}
\begin{sequation}
\SO:= \{ R \in \mathbb{R}^{3\times3} \mid R^\top R = RR^\top= I_3,\; \det(R) = 1 \}.
\end{sequation}
Each element $R \in \SO$ represents a proper rotation that maps vectors from the body frame to the inertial frame, with its transpose performing the inverse transformation.

\noindent\textbf{Lie Algebra and Skew-Symmetric Matrices.}
The tangent space of $\SO$ at the identity is the \emph{Lie algebra} $\mathfrak{so}(3)$, consisting of $3\times3$ skew-symmetric matrices. 
For a vector $\omega = (\omega_1,\omega_2,\omega_3)\in \mathbb{R}^3$, the skew-symmetric operator $(\cdot)^\wedge:\mathbb{R}^3 \to \mathfrak{so}(3)$ is defined as
\begin{sequation}
\omega^\wedge := 
\begin{bmatrix}
0 & -\omega_3 & \omega_2 \\
\omega_3 & 0 & -\omega_1 \\
-\omega_2 & \omega_1 & 0
\end{bmatrix},
\end{sequation}%
with the property $\omega^\wedge u = \omega \times u$ for any $u \in \mathbb{R}^3$.  The inverse map, denoted $(\cdot)^\vee: \mathfrak{so}(3) \to \mathbb{R}^3$, satisfies $(\omega^\wedge)^\vee = \omega$. Together, these operators provide a bijective correspondence between vectors in $\mathbb{R}^3$ and elements of $\mathfrak{so}(3)$.
\end{sidebar}

\begin{sidebar}{Attitude Representation on $\SO$ (Continued) \continuesidebar}
\noindent\textbf{Exponential and Logarithmic Map.}
The \emph{exponential map} provides a smooth correspondence between the Lie algebra $\mathfrak{so}(3)$ and the group $\SO$. 
For any rotation vector $\omega \in \mathbb{R}^3$, the matrix exponential of $\omega^\wedge$ yields a rotation matrix through the celebrated \emph{Rodrigues’ formula}:
\begin{sequation}
\exp(\omega^\wedge) 
= I_3
+ \frac{\sin\|\omega\|}{\|\omega\|}\,\omega^\wedge 
+ \frac{1 - \cos\|\omega\|}{\|\omega\|^2}\,(\omega^\wedge)^2.
\end{sequation}
This expression plays a central role in integrating continuous-time kinematics on $\SO$. The logarithmic map \( \log \colon \SO \to \mathfrak{so}(3) \) retrieves the corresponding Lie algebra matrix:
\begin{sequation}
\log(R) 
= \frac{\theta}{2 \sin\theta} (R - R^\top), \quad
\theta = \cos^{-1}\left( \frac{\mathrm{tr}(R) - 1}{2} \right),
\end{sequation}%
provided the rotation angle satisfies $\theta<\pi$. It defines local coordinates on $\SO$ around the identity, enabling local
linear analysis of rotational quantities.

\noindent\textbf{Jacobians of $\SO$.}
When discretizing the equations of motion or deriving filter updates, one frequently requires the left/right Jacobian of $\SO$, defined for any rotation vector $\omega \in \mathbb{R}^3$ as
\[
J_\ell(\omega) := \mathbf{J}(\omega^\wedge),
\qquad
J_r(\omega) := \mathbf{J}(-\omega^\wedge),
\]
where $\mathbf{J}(\cdot)$ is defined in \eqref{eq:left-jacobian-exp}.
The closed-form expressions of the (left) Jacobian and its inverse are given by
\begin{equation*}
\begin{aligned}
\mathbf{J}(\omega^\wedge) &= I_3
+ \frac{1 - \cos\|\omega\|}{\|\omega\|^2}\,\omega^\wedge
+ \frac{\|\omega\| - \sin\|\omega\|}{\|\omega\|^3}(\omega^\wedge)^2, \\[3pt]
\mathbf{J}^{-1}(\omega^\wedge) &= I_3
- \frac{1}{2}\,\omega^\wedge
+ \left(
\frac{1}{\|\omega\|^2}
- \frac{1 + \cos\|\omega\|}{2\|\omega\|\sin\|\omega\|}
\right)(\omega^\wedge)^2.
\end{aligned}
\end{equation*}
Jacobians relate infinitesimal perturbations in the Lie algebra to finite rotations in the group~\cite{barfoot2024state}. Sometimes, a higher-order term known as the \emph{second-order left Jacobian}, with ``second-order'' referring to the double integral below, is also required:
\begin{equation*}
  \begin{aligned}
Q_\ell(\omega) \;=\;&\; \tfrac{1}{2} I_3
+ \frac{\|\omega\| - \sin\|\omega\|}{\|\omega\|^{3}}\,\omega^\wedge \\[3pt]
&\qquad+ \frac{\|\omega\|^{2} + 2\cos\|\omega\| - 2}
       {2\,\|\omega\|^{4}}\,(\omega^\wedge)^{2}.
\end{aligned}  
\end{equation*}
which satisfies
\begin{sequation}
\int_0^\Delta\!\!\int_0^u 
\exp(\omega^\wedge s)\,ds\,du 
\;=:\; Q_\ell(\omega\Delta)\,\Delta^2.
\end{sequation}
\noindent\textbf{Summary.}
A wide range of attitude representations exist, each offering distinct advantages and limitations~\cite{Shuster1993}. 
The Lie-group formulation emphasizes the intrinsic geometry of rotations and supports consistent, coordinate-free estimation schemes. 
In particular, it underpins recent developments in \emph{invariant filtering} and geometric observer design~\cite{Barrau2017}. 
Throughout this tutorial, the Lie group $\SO$ and its algebra $\mathfrak{so}(3)$ will constitute the mathematical foundation for attitude modeling, discretization, and estimation in inertial navigation.
\end{sidebar}

\section{Basic Notation}
Throughout the paper, $\mathbb{R}^n$ denotes the $n$-dimensional Euclidean
space equipped with the standard inner product
$\langle x,y\rangle = x^\top y$ and norm $\|x\| = \sqrt{x^\top x}$. The identity and zero matrices of appropriate dimensions are denoted by
$I_n$ and $0_{m\times n}$, respectively. For a unit vector $x \in \mathbb{R}^n$
with $\|x\| = 1$, we define the orthogonal projection onto the normal plane as $
\pi_x := I_n - x x^\top$.
For any $x,y\in\mathbb{R}^3$, the vector cross product is written $x\times y$.
The determinant operator is denoted by $\det(\cdot)$.
For any matrices $A$ and $B$ of compatible dimensions, the matrix commutator
(Lie bracket) is defined as $[A,B]=AB-BA$.
The operator $\mathrm{diag}(\cdot)$ denotes a (block) diagonal matrix formed from its
arguments, and $\mathrm{tr}(\cdot)$ denotes the trace of a square matrix.
For matrices $A\in\mathbb{R}^{m\times n}$ and $B\in\mathbb{R}^{p\times q}$,
the Kronecker product is denoted $A\otimes B$.
For any matrix $A\in\mathbb{R}^{m\times n}$, the vectorization operator
$\mathrm{vec}(\cdot)$ stacks the columns of $A$ into a vector in
$\mathbb{R}^{mn}$. Parentheses are used to indicate vector stacking; for example, if
$x_1,x_2\in\mathbb{R}^3$, then $(x_1,x_2)\in\mathbb{R}^6$ denotes the stacked
$6\times 1$ vector $[x_1^\top\;x_2^\top]^\top$.
The notation $\mathcal{O}(k)$ denotes higher-order terms of order $k$ or
greater in a series expansion. Finally, the \emph{integrated exponential matrix function} is defined for any
$A \in \mathbb{R}^{n\times n}$ as
\begin{equation}\label{eq:left-jacobian-exp}
    \mathbf{J}(A)
    := \int_0^1 \exp(sA)\,ds
    \;=\; \sum_{n=0}^\infty \frac{A^n}{(n+1)!},
\end{equation}
which satisfies $\exp(A) = I_n + A\mathbf{J}(A)$.
This operator frequently appears in the analysis and implementation of
Lie-group-based estimation methods. Finally, basic notions related to matrix Lie groups and their associated Lie algebras,
used throughout this paper, are briefly recalled in
"\nameref{sidebar-MatrixLieGroup}".

\section{INS System Modelling}
This section develops the mathematical model of an inertial navigation system (INS) by combining rigid-body kinematics with sensor models. 
The state is defined in terms of orientation, velocity and position, with 
orientation represented by a rotation matrix
$
R\in\SO:=\left\{R\in\mathbb{R}^{3\times3}\mid
R^\top R=I_3,\ \det(R)=1\right\}.$
The system state is also expressed as an element of the Lie group $\SE$. Technical details of these group structures are summarized in ``\nameref{sidebar-Orientation}'' and in ``\nameref{sidebar-SE23}''.

\subsection{Rigid-Body Kinematics}
We consider the motion of a rigid body equipped with an IMU.
Let $\{\mathcal I\}$ denote the local-level inertial frame, chosen here as the
North–East–Down (NED) frame commonly used in navigation.
The body frame $\{\mathcal B\}$ is rigidly attached to the body at the IMU location.
The orientation of the body with respect to the inertial frame is represented by $R \in \SO$, which maps vectors from $\{\mathcal B\}$ to $\{\mathcal I\}$. 
The body’s translational motion is described by its velocity $v \in \mathbb{R}^3$ and position $p \in \mathbb{R}^3$, both expressed in $\{\mathcal I\}$. 
Gravity is modeled as a constant vector $g \in \mathbb{R}^3$ aligned with the down axis, e.g.,
\[
g := (0,\,0,\,g_0), \qquad g_0 \approx 9.81~\text{m/s}^2.
\]

Let $\omega \in \mathbb{R}^3$ denote the body angular velocity expressed in $\{\mathcal B\}$, and $a \in \mathbb{R}^3$ the specific force (non-gravitational acceleration) expressed in $\{\mathcal B\}$. 
The rigid-body kinematics are then
\begin{equation}
\dot R = R\,\omega^\wedge, 
\qquad \dot v = R\,a + g, 
\qquad \dot p = v,
\label{eq:ins_kinematics}
\end{equation}
where $(\cdot)^\wedge$ denotes the skew-symmetric operator such that $u^\wedge v = u\times v$ for any $u,v \in \mathbb{R}^3$. The inertial frame $\{\mathcal{I}\}$, the body-fixed IMU frame $\{\mathcal{B}\}$, together with the navigation states and IMU inputs, are illustrated in Fig.~\ref{fig:ins_kinematics}.

Equations~\eqref{eq:ins_kinematics} describe the continuous-time evolution of the
navigation states under ideal IMU inputs $(\omega,a)$.
Given perfect initial conditions, the orientation evolves through integration of
the angular velocity, the velocity results from integration of the specific force
expressed in the inertial frame together with gravity, and the position is obtained
by integrating the velocity.

In this tutorial, we adopt a local-level \emph{flat-Earth} model, in which the
North--East--Down (NED) frame is assumed inertial and both Earth rotation and
curvature effects are neglected.
This assumption is standard in robotics and short-duration navigation scenarios,
where it provides a sufficiently accurate description while significantly
simplifying the governing equations.
Nevertheless, most of the results presented here can be extended to models that
account for Earth rotation; see also \cite{brossard2022uncertainty}.

\begin{figure}[t]
\centering
\begin{tcolorbox}[colback=red!10, arc=2mm, boxrule=0pt,
                  left=3mm, right=3mm, top=3mm, bottom=3mm]
\includegraphics[height=2.5cm, keepaspectratio]{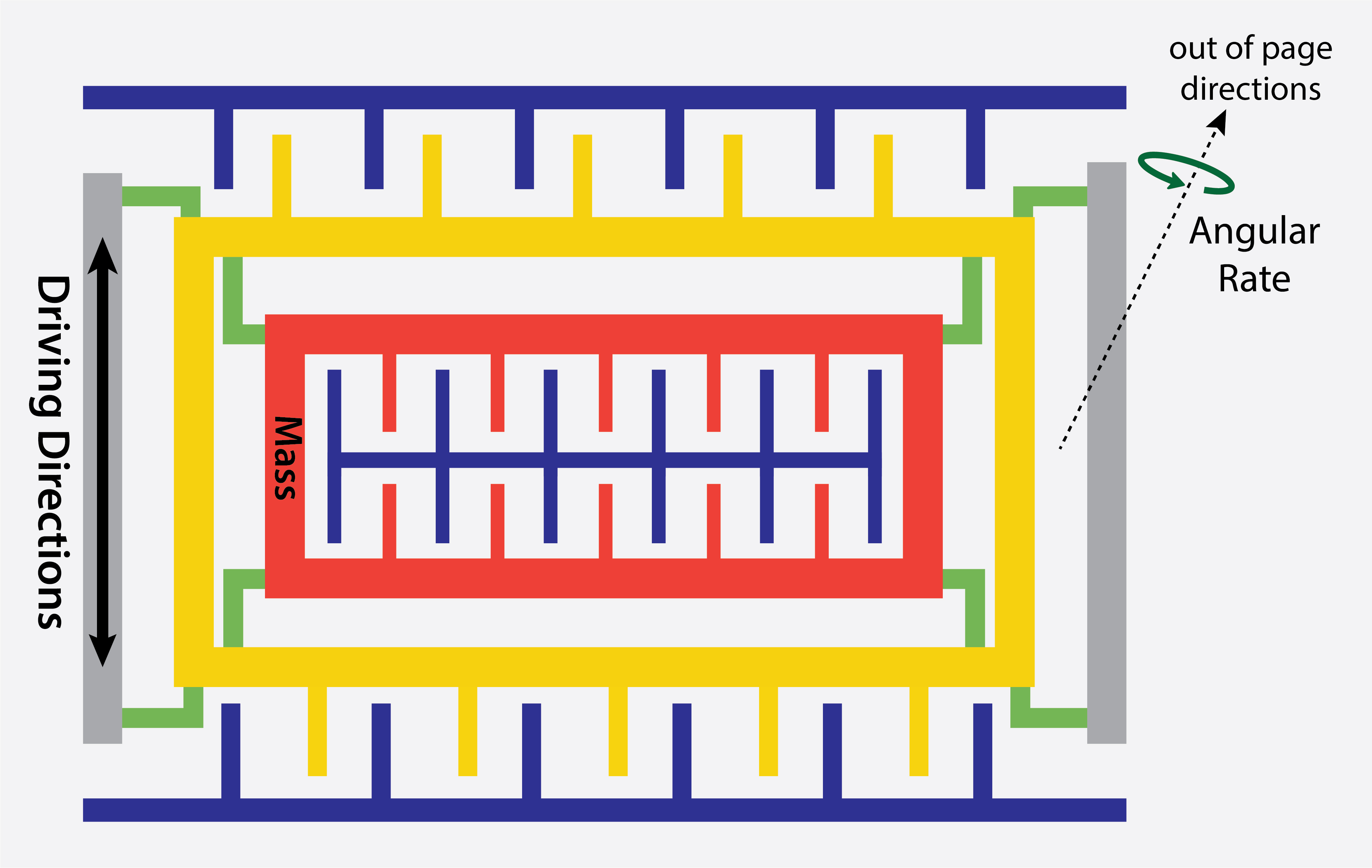}
\hfill
\includegraphics[height=2.5cm, keepaspectratio]{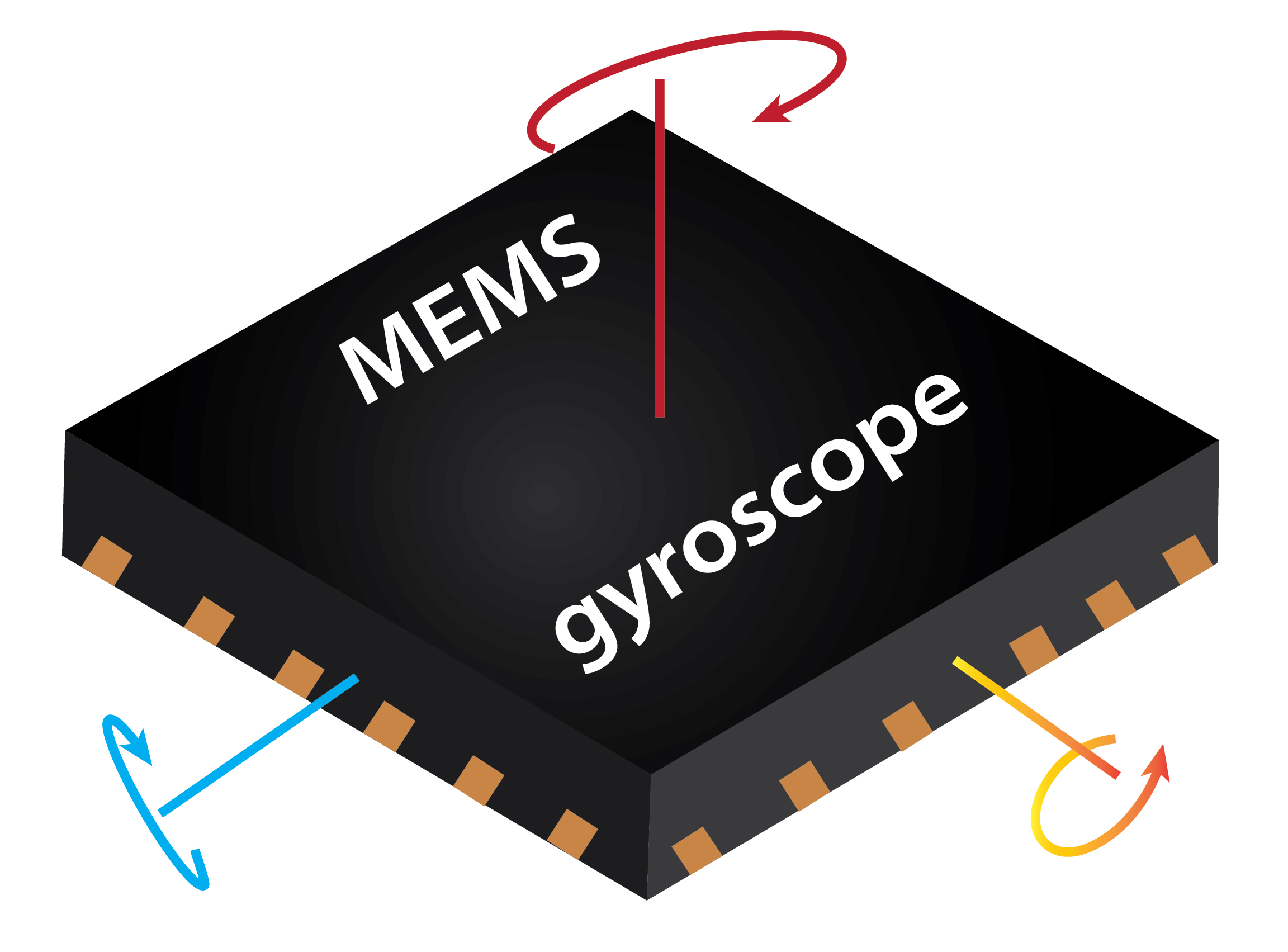}%
\end{tcolorbox}
\vspace{4pt}
\caption{{\bf MEMS gyroscope structure used in inertial sensing}. 
(a) A two-dimensional schematic of a vibrating MEMS gyroscope in which a driven 
proof mass experiences a Coriolis-induced motion proportional to the angular rate. 
(b) A tri-axis MEMS gyroscope with three orthogonal vibrating structures that 
measure the body-frame angular-velocity components about the $x$, $y$, and $z$ axes.}
\label{fig:mems-gyro}
\end{figure}
\begin{figure}[t]
\centering
\begin{tcolorbox}[colback=red!10, arc=2mm, boxrule=0pt,
                  left=3mm, right=3mm, top=3mm, bottom=3mm]

\includegraphics[height=2.5cm, keepaspectratio]{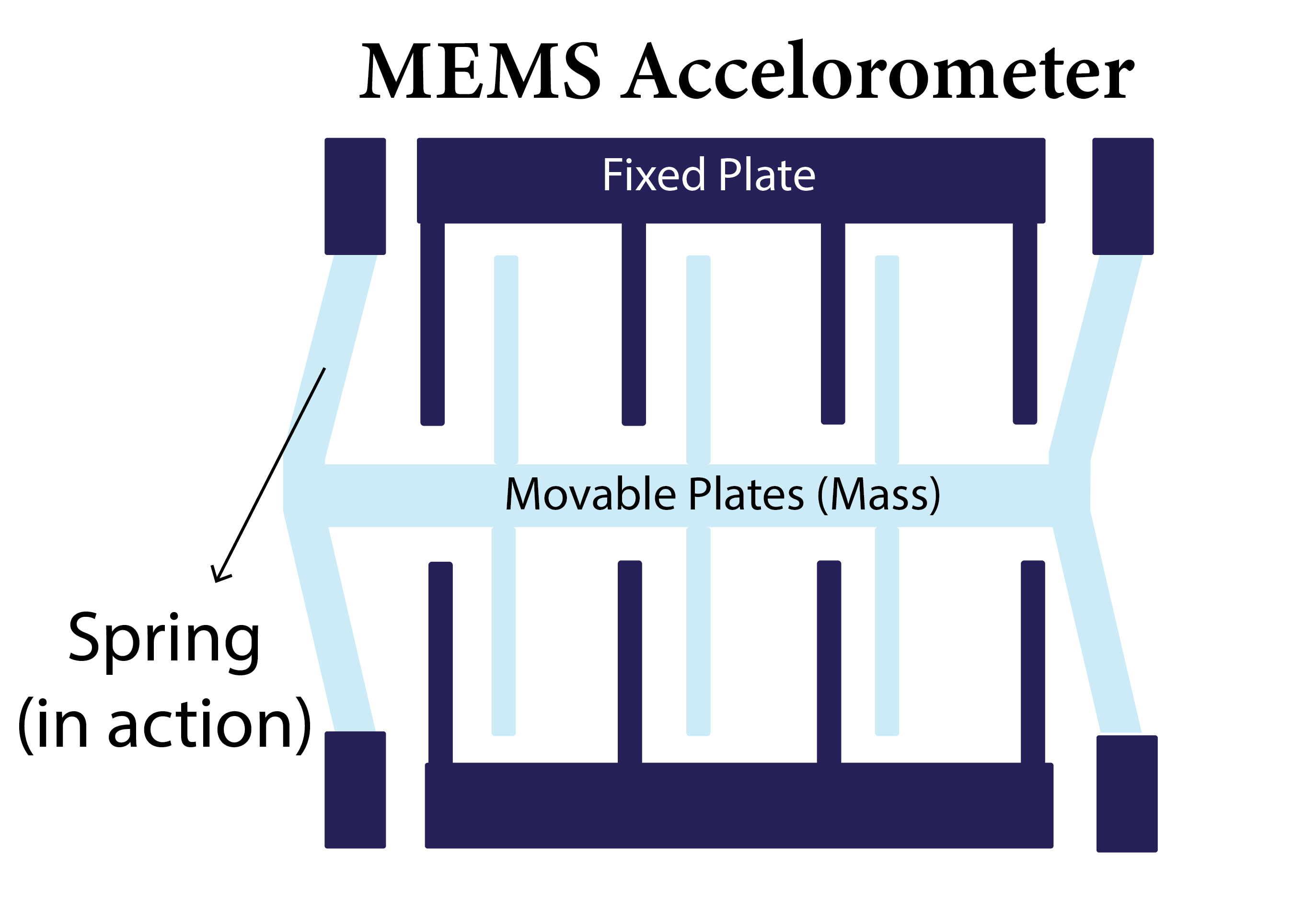}%
\hfill
\includegraphics[height=2.5cm, keepaspectratio]{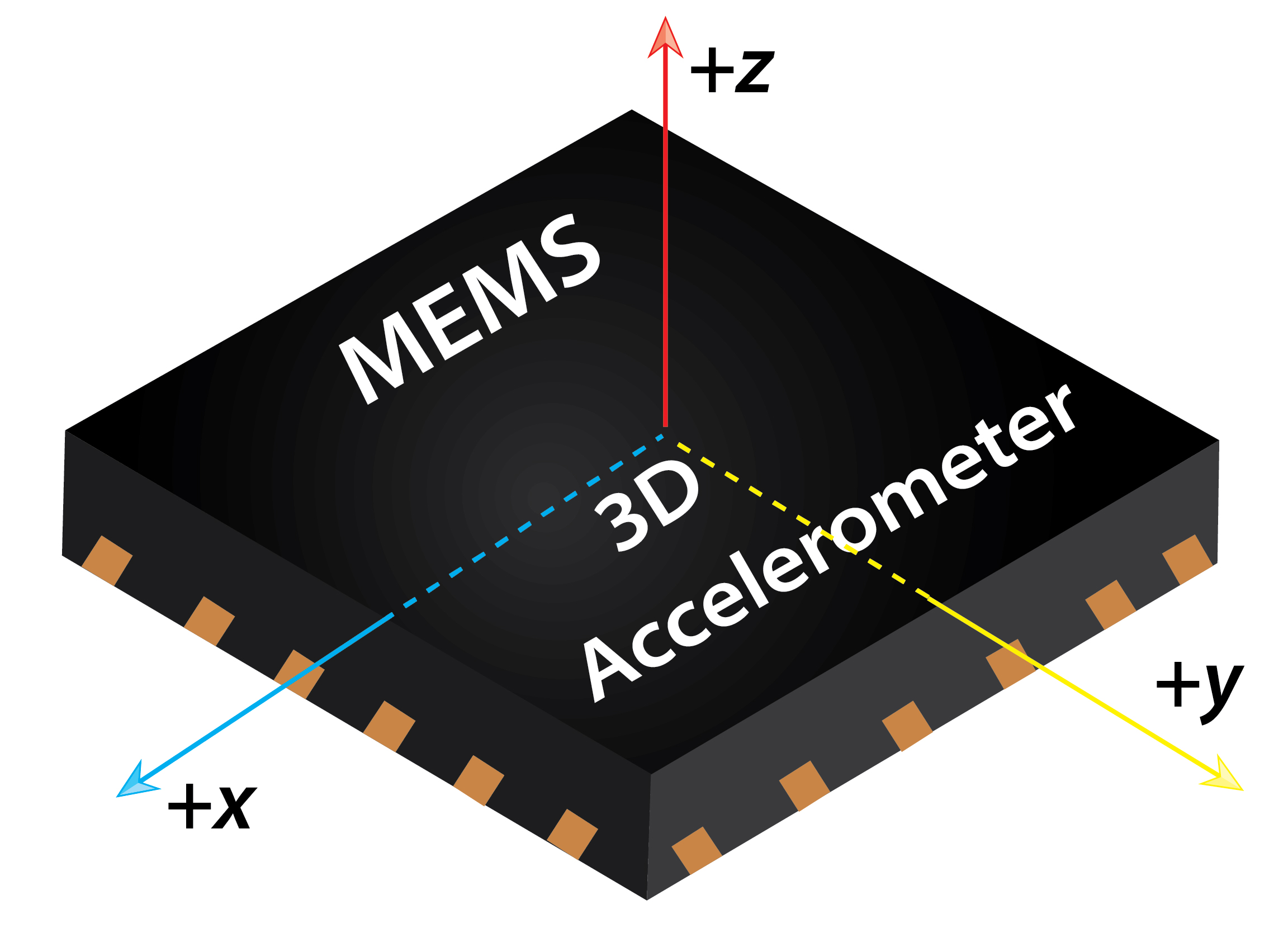}%

\end{tcolorbox}

\vspace{4pt}

\caption{\textbf{MEMS accelerometer structure used in inertial sensing}. 
(a) A two-dimensional view of a capacitive one-axis MEMS accelerometer in which 
a suspended proof mass deflects under acceleration, producing a differential 
change in capacitance. (b) A tri-axis accelerometer consisting of three orthogonal 
sensing axes that measure the specific-force components along the $x$, $y$, and $z$ 
body-frame directions.}
\label{fig:mems-accel}
\end{figure}
\subsection{Inertial Measurement Unit (IMU)}
The inertial measurement unit (IMU) is the core sensing device of an INS. 
It typically consists of three orthogonally mounted gyroscopes and three accelerometers that measure angular velocity and specific force, respectively, as illustrated in Figs.~\ref{fig:mems-gyro} and~\ref{fig:mems-accel}. 
These measurements provide the inputs needed to integrate the rigid-body kinematics of \eqref{eq:ins_kinematics}.

From a technological standpoint, IMUs are implemented using different physical principles \cite{Groves2013}. 
High-end navigation systems employ fiber-optic or ring-laser gyroscopes and high-grade mechanical accelerometers, offering very low noise and bias drift at the cost of size and expense. 
In contrast, most robotics and consumer platforms rely on microelectromechanical systems (MEMS) gyroscopes and accelerometers (see Figs.~\ref{fig:mems-gyro}--\ref{fig:mems-accel}), which are small, lightweight, and inexpensive but significantly more prone to bias instability and noise. 
This trade-off between accuracy, size, and cost strongly influences the performance of any INS.

Although some commercial IMUs integrate additional sensors such as magnetometers (to sense the Earth’s magnetic field) or barometers (to estimate altitude), in this tutorial we treat such information as \emph{aiding measurements} rather than part of the core IMU model. 
This separation emphasizes that the essential inertial navigation process relies only on gyroscopes and accelerometers, while other sensors provide external constraints to mitigate drift.

An ideal gyroscope would provide direct access to the body angular velocity 
$\omega \in \mathbb{R}^3$ expressed in the body frame~$\{\mathcal B\}$. 
In practice, the gyroscope measurement is corrupted by a slowly varying bias
$b_\omega \in \mathbb{R}^3$ and additive measurement noise
$n_\omega \in \mathbb{R}^3$, so that the measured angular velocity
$\omega_m$ satisfies
\begin{equation}
\omega_m = \omega + b_\omega + n_\omega .
\label{eq:gyro_model}
\end{equation}
The bias $b_\omega$ accounts for effects such as sensor miscalibration, temperature drift, 
and turn-on transients, and is typically modeled as a random walk. 
The noise $n_\omega$ represents high-frequency fluctuations, commonly approximated as 
zero-mean Gaussian white noise characterized by a known power spectral density.

Similarly, an ideal accelerometer would measure the specific force 
$a \in \mathbb{R}^3$, defined as the non-gravitational acceleration of the body 
expressed in $\{\mathcal B\}$. 
Relating this quantity to the inertial kinematics~\eqref{eq:ins_kinematics} gives
\[
a = R^\top(\dot v - g),
\]
that is, accelerometers measure the inertial acceleration of the platform 
after removing the gravity contribution. 
The practical measurement model reads
\begin{equation}
a_m = a + b_a + n_a,
\label{eq:accel_model}
\end{equation}
where $b_a \in \mathbb{R}^3$ denotes the accelerometer bias and 
$n_a \in \mathbb{R}^3$ an additive measurement noise.

The bias terms $(b_\omega,b_a)$ are usually modeled as random walks,
\begin{equation}
\dot b_\omega = n_{b_\omega}, 
\qquad 
\dot b_a = n_{b_a},
\label{eq:bias_rw}
\end{equation}
driven by zero-mean Gaussian white processes $n_{b_\omega}$ and $n_{b_a}$.  
This stochastic model captures the slow, unpredictable bias drift 
observed in low-cost MEMS sensors. 
All noise processes $\{n_\omega, n_a, n_{b_\omega}, n_{b_a}\}$ are assumed mutually independent, 
zero-mean, Gaussian white processes, characterized by continuous-time noise intensities 
(power spectral densities) $Q_\omega$, $Q_a$, $Q_{b_\omega}$, and $Q_{b_a}$, respectively.

\begin{sidebar}{The Pose--Velocity Group $\SE$}
\label{sidebar-SE23}

{\setlength{\abovedisplayskip}{2pt}
 \setlength{\belowdisplayskip}{2pt}
 \setlength{\abovedisplayshortskip}{0pt}
 \setlength{\belowdisplayshortskip}{0pt}
 \small

\setcounter{sequation}{0}
\renewcommand{\thesequation}{S\arabic{sequation}}
\setcounter{stable}{0}
\renewcommand{\thestable}{S\arabic{stable}}
\setcounter{sfigure}{0}
\renewcommand{\thesfigure}{S\arabic{sfigure}}
\sdbarinitial{W}hile $\SO$ captures the rotational part of motion, inertial navigation
requires a state that also includes translational quantities such as velocity and position. 
The \emph{pose–velocity group} $\SE$ (also known as the group of double direct spatial isometries or the extended Special Euclidean group \cite{Barrau2017}) provides a compact mathematical framework that unifies 
these variables into a single Lie group structure, allowing for consistent geometric modelling.

\medskip
\noindent\textbf{Group Definition and Notation.}
Each element of $\SE\subset\mathbb{R}^{5\times5}$ represents a triplet $(R,v,p)$,
where $R\in\SO$ denotes the orientation and $v,p\in\mathbb{R}^3$ are the
velocity and position expressed in the inertial frame.
To streamline notation, we define the embedding
\begin{equation*}
\Gamma:\SO\times\mathbb{R}^3\times\mathbb{R}^3\!\to\!\SE,\qquad
\Gamma(R,v,p)
:=\!
\begin{bmatrix}
R & v & p\\
0 & 1 & 0\\
0 & 0 & 1
\end{bmatrix},
\end{equation*}
so that any group element can be written simply as $X=\Gamma(R,v,p)$.
The group operation (matrix multiplication) follows the natural composition of two motions:
\begin{equation*}
\Gamma(R_1,v_1,p_1)\,\Gamma(R_2,v_2,p_2)
= \Gamma(R_1R_2,\; v_1 + R_1v_2,\; p_1 + R_1p_2),
\end{equation*}%
while the inverse corresponds to undoing the motion:
\begin{equation*}
\Gamma(R,v,p)^{-1} = \Gamma(R^\top,\; -R^\top v,\; -R^\top p).
\end{equation*}
Hence, $\SE$ can be viewed as a semidirect product 
$\SO\!\ltimes\!(\mathbb{R}^3\times\mathbb{R}^3)$ of dimension nine.

\medskip
\noindent\textbf{Lie Algebra.}
The Lie algebra associated with $\SE$ is the matrix Lie algebra
$\se \subset \mathbb{R}^{5\times5}$, defined as the tangent space of $\SE$
at the identity element.
Its elements admit the matrix representation
\begin{sequation}\nonumber
\xi^\wedge =
\begin{bmatrix}
\omega^\wedge & a & b\\[2pt]
0_{2\times3} & 0_{2\times1}& 0_{2\times1}
\end{bmatrix},
\qquad
\xi = (\omega,a,b) \in \mathbb{R}^9,
\end{sequation}
\noindent where $\omega,a,b \in \mathbb{R}^3$. Here, the $(\cdot)^\wedge$ operator, introduced in
``\nameref{sidebar-Orientation}'', is \emph{overloaded} to denote the canonical
embedding of vector coordinates into Lie algebra matrices:
for $\omega\in\mathbb{R}^3$, $\omega^\wedge\in\so$, and for
$\xi=(\omega,a,b)\in\mathbb{R}^9$, $\xi^\wedge\in\se$.
The inverse mapping $(\cdot)^\vee:\se\to\mathbb{R}^9$ satisfies
$(\xi^\wedge)^\vee=\xi$.

\noindent\textbf{Exponential Map and Logarithmic Map.}
The exponential map $\exp:\se\!\to\!\SE$ transforms an infinitesimal motion in the Lie algebra
into a finite motion on the group manifold.
For $(\omega,a,b)^\wedge\in\se$, the closed-form expression is
\begin{sequation}
\exp\!\big((\omega,a,b)^\wedge\big)
=
\Gamma\!\left(
\exp(\omega^\wedge),
\,J_\ell(\omega)a,
\,J_\ell(\omega)b
\right),
\end{sequation}
\noindent where $J_\ell(\omega)$ denotes the left Jacobian of $\SO$ (see "\nameref{sidebar-Orientation}'').
The logarithmic map \( \log \colon \SE \to\se \) retrieves the Lie algebra element corresponding to a finite group transformation. Given a matrix-form group element \( \Gamma(R, v, p) \in \SE \), such that $\mathrm{tr}(R)\neq -1$, the log map is:
\begin{sequation}
\log\left( \Gamma(R, v, p) \right)
=\left(
\phi,\,
J_\ell^{-1}(\phi)\, v,\,
J_\ell^{-1}(\phi)\, p
\right)^\wedge,
\end{sequation}
where:
\( \phi = \log(R)^\vee \in \mathbb{R}^3 \) is the rotation vector associated with \( R \in \SO \).

\noindent\textbf{Adjoint Operators.}
The adjoint representation characterizes how infinitesimal motions expressed in one tangent space
are transported to another through group composition.  
For $X \in \SE$, the adjoint map
$\BAd_X : \se \to \se$ is defined by
\[
\BAd_X(\xi^\wedge) = X\,\xi^\wedge\,X^{-1}.
\]
For a fixed $X = \Gamma(R, v, p)$, it is a linear operator whose matrix representation acts on $\xi = (\omega, a, b) \in \mathbb{R}^9$ as
\begin{sequation}\label{eq:adjoint}
\Ad_X =
\begin{bmatrix}
R & 0 & 0\\[2pt]
v^\wedge R & R & 0\\[2pt]
p^\wedge R & 0 & R
\end{bmatrix},
\qquad 
\Ad_X \xi = (\BAd_X(\xi^\wedge))^\vee.
\end{sequation}
Note that, for clarity, the bold symbol $\BAd_X$ denotes the adjoint operator acting on the Lie algebra,
while its matrix representation in $\mathbb{R}^9$ is written $\Ad_X$.
The \emph{small adjoint} (Lie algebra adjoint) is the linear operator
$\Bad_{\xi^\wedge}:\se\to\se$ defined by the Lie bracket
\[
\Bad_{\xi^\wedge}(\eta^\wedge) := [\xi^\wedge,\eta^\wedge],
\qquad \forall\,\eta^\wedge\in\se.
\]
Its matrix representation with respect to the vector coordinates
$\xi=(\omega,a,b)\in\mathbb{R}^9$ is denoted by $\ad_\xi\in\mathbb{R}^{9\times9}$ and
satisfies for all $\eta\in\mathbb{R}^9$
\begin{sequation}\label{eq:small_adjoint}
\ad_\xi =
\begin{bmatrix}
\omega^\wedge & 0 & 0\\[2pt]
a^\wedge & \omega^\wedge & 0\\[2pt]
b^\wedge & 0 & \omega^\wedge
\end{bmatrix},
\qquad
\ad_\xi\,\eta = (\Bad_{\xi^\wedge}(\eta^\wedge))^\vee.
\end{sequation}
The matrix representations of the two adjoint operators are linked by construction:
the small adjoint $\ad_\xi$ is the differential of $\Ad_{\exp(t\,\xi^\wedge)}$ at
$t=0$, namely,
\[
\Ad_{\exp(t\,\xi^\wedge)} = \exp\!\big(t\,\ad_\xi\big).
\]

This infinitesimal form is particularly useful for deriving linearized error dynamics and for expressing
Jacobian factors appearing in invariant filtering frameworks.
\normalsize} 
\end{sidebar}

\begin{sidebar}{The Pose--Velocity Group $\SE$ (Continued)\continuesidebar}
\noindent\textbf{Jacobians on $\SE$.} 
In analogy with the $\SO$ case, the left and right Jacobians on $\SE$ 
are defined in terms of the adjoint operator of the Lie algebra element 
$\xi \in \mathbb{R}^9$ as
\[
\mathcal{J}_\ell(\xi) := \mathbf{J}\!\big(\ad_{\xi}\big),
\qquad
\mathcal{J}_r(\xi) := \mathbf{J}\!\big(-\ad_{\xi}\big),
\]
where $\mathbf{J}(\cdot)$ denotes the matrix function introduced in "Basic Notation.’’
The left Jacobian admits the following block structure:
\begin{sequation}
\mathcal{J}_\ell(\omega,a,b) =
\begin{bmatrix}
J_\ell(\omega) & 0 & 0\\[3pt]
\mathcal{Q}_{\omega,a} & J_\ell(\omega) & 0\\[3pt]
\mathcal{Q}_{\omega,b} & 0 & J_\ell(\omega)
\end{bmatrix},
\end{sequation}
where the cross-coupling blocks $\mathcal{Q}_{\omega,\star}$ are given by
\begin{sequation}
\begin{aligned}
\mathcal{Q}_{\omega,\nu} 
={}&\;
\tfrac{1}{2}\,\nu^\wedge 
+ \tfrac{\theta - \sin\theta}{\theta^3}
\!\left(
\omega^\wedge \nu^\wedge + \nu^\wedge \omega^\wedge + \omega^\wedge \nu^\wedge \omega^\wedge
\right) \\[2pt]
&+ \tfrac{\theta^2 + 2\cos\theta - 2}{2\theta^4}
\!\left(
\omega^\wedge \omega^\wedge \nu^\wedge + \nu^\wedge \omega^\wedge \omega^\wedge - 3\,\omega^\wedge \nu^\wedge \omega^\wedge
\right) \\[2pt]
&+ \tfrac{2\theta - 3\sin\theta + \theta\cos\theta}{2\theta^5}
\!\left(
\omega^\wedge \nu^\wedge \omega^\wedge \omega^\wedge + \omega^\wedge \omega^\wedge \nu^\wedge \omega^\wedge
\right),
\end{aligned}
\end{sequation}
with $\theta := \|\omega\|$ and $\nu \in \{a,b\}$.
Since $\mathcal{J}_\ell(\omega,a,b)$ is block lower–triangular with identical diagonal blocks,
its inverse takes the simple form
\begin{sequation}
\mathcal{J}^{-1}_\ell(\omega,a,b) =
\begin{bmatrix}
J^{-1}_\ell(\omega) & 0 & 0\\[3pt]
-\,J^{-1}_\ell(\omega)\mathcal{Q}_{\omega,a}J^{-1}_\ell(\omega) & J^{-1}_\ell(\omega) & 0\\[3pt]
-\,J^{-1}_\ell(\omega)\mathcal{Q}_{\omega,b}J^{-1}_\ell(\omega) & 0 & J^{-1}_\ell(\omega)
\end{bmatrix}.
\end{sequation}
The right Jacobian admits similar expressions in view of 
$\mathcal{J}_r(\xi) = \mathcal{J}_\ell(-\xi)$.

These Jacobians describe how infinitesimal perturbations in the Lie algebra 
map to finite variations on the group—an essential ingredient for invariant 
filtering and linearization. For small $\eta\in\mathbb{R}^9$, the following first-order identities hold:
\begin{sequation}\label{eq:J:left:perturbation}
\begin{aligned}
\exp(\eta^\wedge)\,\exp(\xi^\wedge)
&\;\approx\;
\exp\!\Big(\big(\xi + \mathcal{J}_\ell^{-1}(\xi)\,\eta\big)^\wedge\Big),\\[2pt]
\exp(\xi^\wedge)\,\exp(\eta^\wedge)
&\;\approx\;
\exp\!\Big(\big(\xi + \mathcal{J}_r^{-1}(\xi)\,\eta\big)^\wedge\Big),
\end{aligned}
\end{sequation}
\noindent which express how left- and right-multiplicative perturbations propagate through 
the exponential map.

Finally, at the homogeneous-matrix level,  and for any element $(\omega,a,b)^\wedge\in\se$ the matrix function $\mathbf{J}$ admits the explicit closed-form expression
\begin{sequation}\label{eq:Jhom}
\mathbf{J}\!\big((\omega,a,b)^\wedge\big)
=
\begin{bmatrix}
J_\ell(\omega) & Q_\ell(\omega)a & Q_\ell(\omega)b\\[2pt]
0 & 1 & 0\\[2pt]
0 & 0 & 1
\end{bmatrix},
\end{sequation}
where $J_\ell(\omega)$ and $Q_\ell(\omega)$ denote, respectively,
the first- and second-order left Jacobians on $\SO$.
This matrix should not be confused with the Lie-group left Jacobian
$\mathcal{J}_\ell(\cdot)$,
which acts on tangent coordinates and is used for linearization or covariance propagation.
\end{sidebar}
\subsection{Overall INS Process Model}
Combining the rigid-body kinematics~\eqref{eq:ins_kinematics} 
with the sensor models~\eqref{eq:gyro_model}–\eqref{eq:bias_rw} 
yields the complete stochastic process model of the inertial navigation system.
The augmented navigation state
\[
(R,v,p,b_\omega,b_a)
\;\in\;
\SO\times\mathbb{R}^3\times\mathbb{R}^3\times\mathbb{R}^3\times\mathbb{R}^3
\]
evolves according to
\begin{equation}
\left\{
\begin{aligned}
\dot R &= R\,(\omega_m - b_\omega - n_\omega)^\wedge, \\[2pt]
\dot v &= R\,(a_m - b_a - n_a) + g, \\[2pt]
\dot p &= v, \\[2pt]
\dot b_\omega &= n_{b_\omega}, \\[2pt]
\dot b_a &= n_{b_a}.
\end{aligned}
\right.
\label{eq:ins_full}
\end{equation}
Equation~\eqref{eq:ins_full} summarizes the practical continuous-time INS dynamics
driven by measured IMU signals.
The attitude kinematics are propagated using the measured angular velocity
$\omega_m$ after bias and noise compensation, while the velocity dynamics are
governed by the measured specific force $a_m$, expressed in the inertial frame and
combined with gravity.
The position is obtained by integrating the velocity, and the inertial sensor
biases are modelled as slowly varying stochastic processes.

The state equations~\eqref{eq:ins_full} written in terms of $(R,v,p)$ constitute the
classical continuous-time INS model and have been successfully used for decades
in navigation and estimation.
More recently, it has been observed that these dynamics possess an intrinsic
geometric structure that can be exploited to obtain more compact representations
and to guide the design and analysis of estimation algorithms~\cite{Barrau2017}.
In particular, embedding the attitude, velocity, and position into the matrix Lie
group $\SE$ reveals invariance properties that are not apparent in the
component-wise formulation and that play a key role in the development of
geometry-aware observers and filters.

Using the $\SE$ representation recalled in "\nameref{sidebar-SE23}", the practical
INS dynamics~\eqref{eq:ins_full} can be expressed directly at the group level.
To this end, define
\[
\begin{aligned}
X &= \Gamma(R,v,p) \in \SE, 
\qquad U = (\omega_m,a_m,0)^\wedge\in\se,\\[2pt]
B &= (b_\omega,b_a,0)^\wedge,\qquad
G = (0,g,0)^\wedge,\qquad
N = (n_\omega,n_a,0)^\wedge,
\end{aligned}
\]
where $\Gamma(\cdot)$ and $(\cdot)^\wedge$ denote the group and Lie algebra
embeddings on $\SE$.
With these definitions, the INS process equations take the form
\begin{equation}
\left\{
\begin{aligned}
\dot X &= X\,(U - B - N) \;+\; G\,X \;+\; [X,D],\\[2pt]
\dot B &= (n_{b_\omega},n_{b_a},0)^\wedge,
\end{aligned}
\right.
\label{eq:ins_SE}
\end{equation}
where $D$ is a constant $5\times5$ matrix defined in
Table~\ref{tab:constant-matrices}.
This representation makes explicit the Lie--group structure of the INS dynamics:
the state evolves on $\SE$ through a combination of a left-invariant
IMU-driven term, a right-invariant gravity term, and the commutator
term $[X,D]$, which encodes the velocity--position kinematic coupling.
As shown later, this commutator term generates an automorphism of $\SE$
that plays a central role in the discrete-time propagation and
preintegration properties of the INS dynamics.
The dynamics are, therefore, written intrinsically on the group and its Lie algebra,
avoiding local coordinates and providing a natural foundation for invariant filtering
and observer design.

\begin{figure}[t]
\centering
\begin{tcolorbox}[
  colback=red!10,
  arc=2mm,
  boxrule=0pt,
  left=3mm,
  right=3mm,
  top=3mm,
  bottom=3mm
]
\centering
\begin{tikzpicture}[
  font=\small,
  block/.style={draw, rounded corners, align=center, minimum width=3cm, minimum height=1.15cm},
  io/.style={
  draw,
  rounded corners,
  align=center,
  minimum width=2.6cm,
  minimum height=0.95cm,
  fill=gray!15
}
,
  arrow/.style={-{Stealth[length=2mm]}, thick},
  dashedarrow/.style={-{Stealth[length=2mm]}, thick, dashed}
]

\node[io] (imu) {IMU\\$\omega_m,\ a_m$};

\node[block, below=12mm of imu] (prop) {%
Propagation on $\SE$\\[+2pt]
{\footnotesize $\dot{\hat X}=\hat XU+G\hat X+[\hat X, D]$}
};

\node[io, below=11mm of prop] (state) {%
State estimate\\
$\hat X=\Gamma(\hat R,\hat v,\hat p)$
};

\node[io, right=16mm of imu, minimum width=2.5cm] (aid) {%
Aiding sensors\\
GNSS, Vision, LiDAR\\
MAG, BAR
};

\node[block, below=10mm of aid, minimum width=2.5cm] (innovation) {%
Innovation\\
{\footnotesize $z_k=y_k-h(\hat X_k)$}
};

\node[block, below=10mm of innovation] (upd) {%
Correction on $\SE$\\[+2pt]
{\footnotesize $\hat X_k^+ = \exp\!\big(-(K_k z_k)^\wedge\big)\,\hat X_k$}
};

\node[font=\bfseries] (propcol) at ($(imu.north)+(0,7.5mm)$) {Prediction};
\node[font=\bfseries] (corrcol) at ($(aid.north)+(0,6mm)$) {Update};

\draw[arrow] (imu) -- (prop);
\draw[arrow] (prop) --node[right, midway, font=\footnotesize] {continuous}  (state);
\draw[arrow] (innovation) -- (upd);
\draw[arrow] (aid) -- (innovation);
\draw[dashedarrow] (upd) --node[below, midway, font=\footnotesize] {discrete}  (state);

\end{tikzpicture}
\end{tcolorbox}
\caption{\textbf{Schematic overview of the recursive bias-free INS estimation architecture.}
The state is propagated forward in time using high-rate IMU measurements through
the Lie-group process model~\eqref{eq:strapdown_group} (prediction step), whose
exact discrete-time update is given in~\eqref{eq:hatX_update}, together with the
associated covariance propagation in~\eqref{eq:Sigma-next-4th-A}.
The state is then intermittently corrected using measurements from aiding sensors
(update step) via the computation of the innovation~\eqref{eq:innovation}.
The Kalman gain $K_k$ is computed according to~\eqref{eq:gain1}--\eqref{eq:gain2}.}
\label{fig:ins_architecture}
\end{figure}
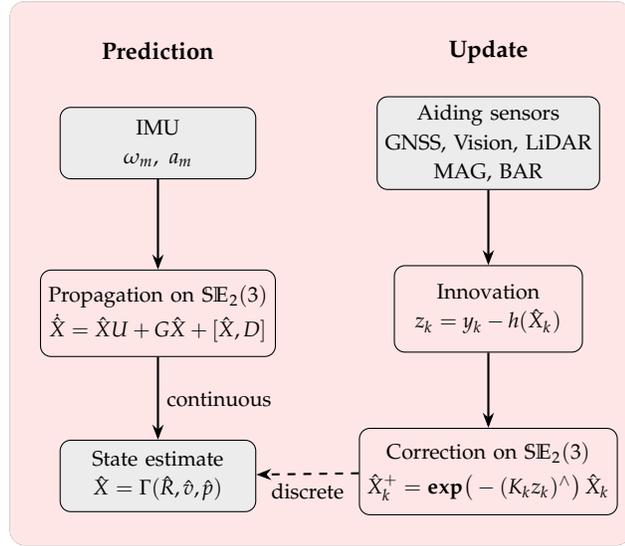

Throughout this tutorial, we adopt a recursive state estimation framework in which
the system state is continuously evolved forward in time using a process model driven by IMU
measurements (\emph{propagation} or \emph{prediction} step), and intermittently
corrected using measurements from aiding sensors such as GNSS, cameras, LiDAR,
magnetometers, or barometers (\emph{update} or \emph{correction} step).
This prediction--correction architecture is common to classical continuous-discrete Kalman filtering and its variants \cite{Jazwinski1970}.
Readers seeking a concise overview of the invariant filtering viewpoint adopted
in this work are referred to the \nameref{sidebar-InEKF}, which summarizes the key filtering
idea used throughout the paper.
A schematic overview of this propagation--correction architecture is shown in
Fig.~\ref{fig:ins_architecture}.

\section{Part I: Propagation of State and Uncertainty from IMU Measurements}
The propagation step of the estimator is governed by the continuous-time
INS equations of motion~\eqref{eq:ins_full}, which describe the ideal evolution
of the system state driven by inertial measurements.
In practice, however, the IMU provides discrete-time samples of angular velocity
and specific force at high frequency (typically 100--1000~Hz), which must be
integrated numerically to propagate the state estimate forward in time.
This operation is traditionally referred to as \emph{strapdown inertial navigation}
in the navigation literature, reflecting the fact that the sensors are rigidly
mounted (``strapped down'') to the vehicle body.
In the robotics and aerospace communities, the same process is often termed
\emph{inertial odometry} or \emph{inertial dead reckoning}, emphasizing its role as
an incremental motion estimator analogous to wheel odometry.
Despite the different terminology, all these viewpoints refer to the same
principle: recursive integration of IMU measurements to estimate orientation,
velocity, and position, based on the process model introduced in the previous
section.

In this part, we focus on the discrete-time propagation of the state and its
associated uncertainty from raw IMU measurements, forming the prediction step
used in the invariant filtering framework considered in this tutorial.

\subsection{Discrete-Time Propagation from Raw IMU Measurements}
In odometry, the gyroscope and accelerometer biases are either assumed to be known or estimated online. 
The bias-compensated angular velocity and specific force are obtained from the raw IMU measurements as
\begin{equation}
\hat{\omega} := \omega_m - \hat{b}_g, 
\qquad 
\hat{a} := a_m - \hat{b}_a,
\end{equation}
where $\hat{b}_g$ and $\hat{b}_a$ denote the estimated gyroscope and accelerometer biases, respectively. 
During propagation, these biases are typically modelled as constants over each integration interval:
\[
\dot{\hat{b}}_g = 0, 
\qquad 
\dot{\hat{b}}_a = 0.
\]
In view of \eqref{eq:ins_full}, the propagated state estimates $(\hat{R}, \hat{v}, \hat{p})$ satisfy the continuous-time \emph{strapdown equations}:
\begin{equation}
\dot{\hat{R}} = \hat{R}\,\hat{\omega}^\wedge, 
\qquad 
\dot{\hat{v}} = \hat{R}\,\hat{a} + g,
\qquad 
\dot{\hat{p}} = \hat{v},
\label{eq:strapdown_ct}
\end{equation}
for some initial estimates $(\hat R(0),\hat v(0),\hat p(0))\in\SO\times\mathbb{R}^3\times\mathbb{R}^3$.
The same propagation can be compactly expressed on the Lie group $\SE$.
Defining the estimated state $\hat X = \Gamma(\hat R,\hat v,\hat p) \in \SE$, the dynamics \eqref{eq:strapdown_ct} can be written as
\begin{equation}
\dot{\hat X} \;=\; \hat X\,\hat U \;+\; G\,\hat X \;+\; [\hat X, D], 
\qquad 
\hat X(0) \in \SE,
\label{eq:strapdown_group}
\end{equation}
where $\hat U := U - \hat B$, $\hat B:=(\hat{b}_g, \hat{b}_a, 0)^\wedge$, with $U$, $G$, and $D$ as introduced in~\eqref{eq:ins_SE}. 

{The INS dynamics~\eqref{eq:strapdown_group} are group affine
(see ``\nameref{sidebar-InEKF}''). This structure implies that their
finite-time flow admits a left--right factorization into group
translations and an automorphism
\cite[Corollary~9]{barrau2019linear}. For a prescribed input history,
the translation factors and the automorphism are independent of
the initial state.}

{In particular, for the INS decomposition in~\eqref{eq:strapdown_group}, the
velocity--position coupling $X\mapsto[X,D]$ generates a smooth
one-parameter group of automorphisms $\Phi_t:\SE\to\SE$, given by}
\begin{equation}
    \Phi_t(X)
    := \exp(-tD)\,X\,\exp(tD),
    \qquad t\in\mathbb{R}.
    \label{eq:Phi_automorphism}
\end{equation}
$\Phi_t$ has inverse $\Phi_{-t}$ and its infinitesimal generator is
\begin{equation}
    \left.\frac{\mathrm d}{\mathrm dt}\Phi_t(X)\right|_{t=0}
    = [X,D],
\end{equation}
and, since $\Phi_t$ is an automorphism, it satisfies
\begin{equation}
    \Phi_t(XY)=\Phi_t(X)\Phi_t(Y),
    \qquad
    \Phi_{t+s}=\Phi_t\circ\Phi_s.
    \label{eq:Phi_properties}
\end{equation}
{The resulting exact propagation over any interval $[t,t+\Delta t]$
is~\cite[Proposition~22]{barrau2019linear}
(see also \cite{brossard2022uncertainty})}
\begin{equation}\label{eq:Xnew}
    \hat X(t+\Delta t)
    =
    G_{\Delta t}\,
    \Phi_{\Delta t}\!\big(\hat X(t)\big)\,
    \hat M_t(\Delta t).
\end{equation}
Each factor in~\eqref{eq:Xnew} has a distinct geometric role:
\begin{itemize}
    \item $G_{\Delta t}\in \SE$ is the \emph{gravity increment}, representing the deterministic effect of gravity over $\Delta t$:
    \[
    G_{\Delta t} \;:=\; \Gamma\!\left(I_3,\; g\,\Delta t,\; \tfrac{1}{2}g\,\Delta t^2\right).
    \]
    \item $\Phi_{\Delta t}\!\big(\hat X(t)\big)$ captures the
\emph{velocity--position kinematic coupling} by propagating the position
according to the current velocity while leaving the attitude and velocity
unchanged:
\[
\Phi_{\Delta t}\!\big(\hat X(t)\big)
=
\Gamma\!\left(
\hat R(t),\;
\hat v(t),\;
\hat p(t)+\Delta t\,\hat v(t)
\right).
\]
    \item $\hat{M}_t(\Delta t)=\Gamma\!\big(\bar R_t(\Delta t),\bar v_t(\Delta t),\bar p_t(\Delta t)\big)\in \SE$ 
    is the \emph{IMU increment} capturing the contribution of IMU data over the interval, defined as the solution of
    \begin{equation}\label{eq:dbarX}
    \begin{cases}
    \dot{\hat{M}}_t(\tau) = \hat{M}_t(\tau)\,\hat U(t+\tau) + [\hat{M}_t(\tau),D], \\[0.25em]
    \hat{M}_t(0) = I_5,\quad \tau \in [0,\Delta t].
    \end{cases}
    \end{equation}
\end{itemize}
{Equation~\eqref{eq:Xnew} reflects directly the structure of the
continuous-time dynamics~\eqref{eq:strapdown_group}: the gravity term
$G\hat X$ gives rise to the increment $G_{\Delta t}$, the commutator
term $[\hat X,D]$ induces the automorphism
$\Phi_{\Delta t}$, and the left-invariant IMU-driven term
$\hat X\hat U$ is captured by the motion increment
$\hat M_t(\Delta t)$. Thus, the exact discrete-time factorization
preserves the geometric decomposition of the continuous-time INS
dynamics.}

The IMU increment $\hat M_t(\Delta t)$ depends only on the
bias-compensated IMU input over $[t,t+\Delta t]$ and is independent of
the initial navigation state $\hat X(t)$. This separation is particularly
useful when accumulating inertial motion between successive update instants,
since the IMU-driven contribution can be computed independently of the
initial pose, velocity, and position. {Nevertheless, computing the IMU factor can still require numerical
integration.}

Under the standard zero-order-hold assumption on the bias-compensated IMU
input, the group ODE~\eqref{eq:dbarX} admits the following compact
closed-form solution.
\begin{lemma}[Exact per-sample IMU increment]
\label{lem:per-sample-inc}
Assume that the bias-compensated IMU input is constant over
$[t_k,t_{k+1}]$, \textit{i.e.},
$\hat U(t)=\hat U(t_k)$ for $t\in[t_k,t_{k+1}]$, with
$\Delta t=t_{k+1}-t_k$. Then the solution of~\eqref{eq:dbarX} is
\begin{equation}
\label{eq:per-sample-solution}
\hat M_{t_k}(\Delta t)
=
\exp\!\big(\Delta t\,\hat U(t_k)\big)
+
\Delta t\,
\Big(
\mathbf J\!\big(\Delta t\,\hat U(t_k)\big)-I_5
\Big)D .
\end{equation}
\end{lemma}

{Using the explicit expressions of the exponential map and
$\mathbf J(\cdot)$ on $\SE$ (see ``\nameref{sidebar-SE23}''), the block
components of~\eqref{eq:per-sample-solution} are
\begin{align}
\label{eq:barR}
\bar R_{t_k}(\Delta t)
&=
\exp\!\big(\hat\omega(t_k)^\wedge\Delta t\big),
\\[2pt]
\bar v_{t_k}(\Delta t)
&=
J_\ell\!\big(\hat\omega(t_k)\Delta t\big)
\,\hat a(t_k)\,\Delta t,
\\[2pt]
\bar p_{t_k}(\Delta t)
&=
Q_\ell\!\big(\hat\omega(t_k)\Delta t\big)
\,\hat a(t_k)\,\Delta t^2 .
\label{eq:barp}
\end{align}
Here $J_\ell(\cdot)$ and $Q_\ell(\cdot)$ denote the first- and
second-order left Jacobians of $\SO$, respectively
(see ``\nameref{sidebar-Orientation}'').}

{The component-wise expressions~\eqref{eq:barR}--\eqref{eq:barp} are
well established in the IMU integration and preintegration literature
\cite{eckenhoff2019closed,hartley2020contact}. The compact group-level
expression~\eqref{eq:per-sample-solution} provides a direct derivation of
these three increments simultaneously, without separately integrating
the attitude, velocity, and position equations. For small $\|\hat\omega(t_k)\Delta t\|$, the approximations
$J_\ell\approx I$ and $Q_\ell\approx \tfrac{1}{2}I$ recover the
classical IMU propagation equations~\cite{Groves2013}. Retaining the full
Jacobian expressions yields the exact propagation under the
zero-order-hold assumption, without requiring the rotation increment
$\hat\omega(t_k)\Delta t$ to be small.
}

Finally, for compactness, let $\hat X_k:=\hat X(t_k)$ and
$\hat M_k:=\hat M_{t_k}(\Delta t)$. {Combining~\eqref{eq:Xnew} with
Lemma~\ref{lem:per-sample-inc}, the exact one-step propagation under the
zero-order-hold assumption reads}
\begin{equation}
\hat X_{k+1}
=
G_{\Delta t}\,\Phi_{\Delta t}(\hat X_k)\,\hat M_k .
\label{eq:hatX_update}
\end{equation}
In block form, the update~\eqref{eq:hatX_update} expands to
\begin{align}
\hat{R}_{k+1} &= 
\hat{R}_k\,\exp\!\big(\hat{\omega}(t_k)^\wedge \Delta t\big), \\[2pt]
\hat{v}_{k+1} &= 
\hat{v}_k + g\,\Delta t 
+ \hat{R}_k\,J_\ell\!\big(\hat{\omega}(t_k)\Delta t\big)\,
\hat{a}(t_k)\,\Delta t, \\[2pt]
\hat{p}_{k+1} &= 
\hat{p}_k + \hat{v}_k \Delta t + \tfrac{1}{2} g \Delta t^2 
+ \hat{R}_k\,Q_\ell\!\big(\hat{\omega}(t_k)\Delta t\big)\,
\hat{a}(t_k)\,\Delta t^2.
\end{align}
{Equation~\eqref{eq:hatX_update} provides the compact Lie-group form of
the discrete-time propagation, while its component-wise expansion gives
the corresponding orientation, velocity, and position updates.
This propagation model forms the basis for the invariant error and
uncertainty propagation developed next.}

\begin{figure}[t]
\centering
\begin{tcolorbox}[colback=red!10, arc=2mm, boxrule=0pt,
                  left=3mm, right=3mm, top=3mm, bottom=3mm]

\centering
\includegraphics[width=\linewidth]{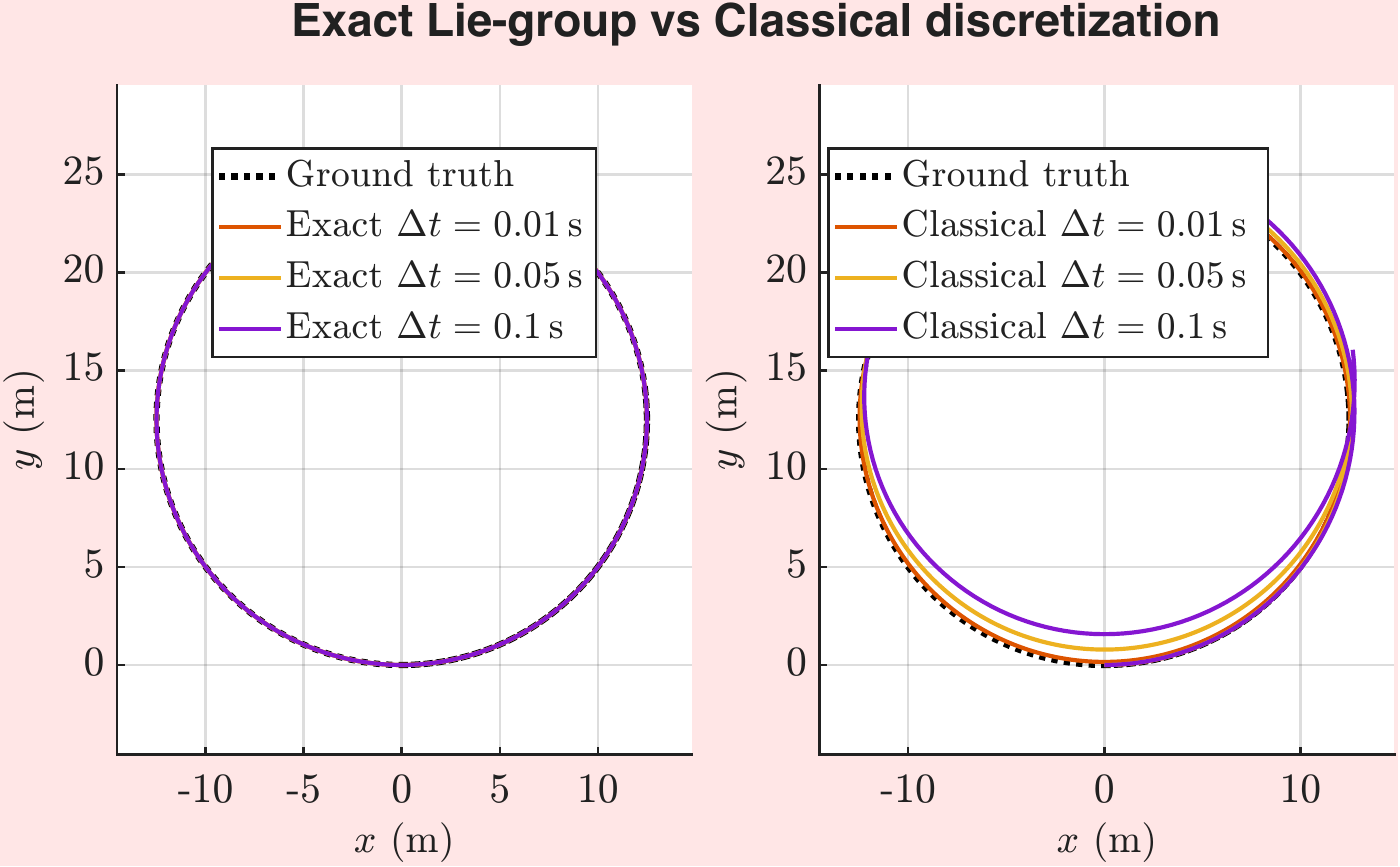}

\end{tcolorbox}

\vspace{4pt}

\caption{\textbf{Exact versus classical IMU discretization for circular motion}.
Comparison of discrete-time propagation under piecewise-constant IMU inputs.
The left panel shows the exact Lie-group propagation on $\SE$ for multiple
sampling intervals $\Delta t$, where the circular trajectory is preserved
independently of $\Delta t$.
The right panel shows the classical INS discretization
($J_\ell \approx I$, $Q_\ell \approx \tfrac{1}{2}I$), for which discretization
errors accumulate as $\Delta t$ increases, resulting in visible geometric
distortion of the trajectory.}
\label{fig:se23_discretization}
\end{figure}

\begin{example}[Exact versus Classical Discretization]\label{ex:discretization}
Figure~\ref{fig:se23_discretization} illustrates the effect of the proposed
Lie-group discretization on a simple circular motion.
The trajectory corresponds to planar motion with constant yaw rate of
$0.4~\mathrm{rad/s}$ and constant body-frame specific force
$a = [0,\;2,\;-g_0]^\top~\mathrm{m/s^2}$, yielding a circular path of radius
$V/\|\omega\| = 12.5~\mathrm{m}$ at speed $V=5~\mathrm{m/s}$.
The exact Lie-group propagation places the sampled positions on the same circular trajectory for
all sampling intervals~$\Delta t$, whereas the classical discretization
($J_\ell \approx I$, $Q_\ell \approx \tfrac{1}{2}I$) exhibits increasing geometric
distortion as~$\Delta t$ grows.
\end{example}

\begin{sidebar}{The Invariant Extended Kalman Filter (InvEKF)}
\label{sidebar-InEKF}
\setcounter{sequation}{0}
\renewcommand{\thesequation}{S\arabic{sequation}}
\setcounter{sfigure}{0}
\renewcommand{\thesfigure}{S\arabic{sfigure}}

\sdbarinitial{T}he Invariant Extended Kalman Filter (InvEKF) is a nonlinear
state estimator for systems evolving on Lie groups
\cite{Barrau2017,Potokar2024}.
For an important class of systems, known as \emph{group-affine systems},
the InvEKF exploits invariant estimation errors whose dynamics are
independent of the true and estimated trajectories. More remarkably, the
nonlinear invariant error admits an exact representation through Lie-algebra
coordinates evolving according to a linear differential equation. This
\emph{log-linear error property} is substantially stronger than merely having
a trajectory-independent Jacobian and is one of the fundamental features
distinguishing invariant filtering from the standard EKF.
Here we briefly recall these properties and the resulting InvEKF structure.

Consider a dynamical system evolving on a matrix Lie group $G$ driven by the input $u$,
\begin{sequation}\label{eq:dX:IEKF}
    \dot X = f_u(X), \qquad X \in G.
\end{sequation}
The dynamics are said to be \emph{group affine} if
\begin{sequation}
f_u(XY) = f_u(X)Y + X f_u(Y) - X f_u(I)Y,
\qquad \forall X,Y \in G.
\end{sequation}
This class includes inertial navigation systems modeled on $\SO$,
$\mathbb{SE}(3)$, and $\SE$, among other Lie groups commonly used in robotics and
navigation. 

Let $\hat X(t)\in G$ be a different (e.g., estimated) trajectory of system \eqref{eq:dX:IEKF} under the same input $u$.  The estimation error is defined as $\tilde X^r=\hat X X^{-1}$ or
$\tilde X^\ell=X^{-1}\hat X$, with $\tilde X$ denoting either form when the
distinction is not essential.  These errors are \textit{invariant} with respect to right (resp. left) multiplication by a constant element of $G$.

A first fundamental consequence of the group-affine property is that these
invariant errors evolve \emph{autonomously}: their dynamics depend on the
error and the known input, but not separately on $X$ or $\hat X$.
Specifically,
\begin{sequation}\label{eq:error-dynamics:IEKF}
\begin{aligned}
\dot{\tilde X}^r
&=
g_u^r(\tilde X^r)
:=
f_u(\tilde X^r)-\tilde X^r f_u(I),\\
\dot{\tilde X}^\ell
&=
g_u^\ell(\tilde X^\ell)
:=
f_u(\tilde X^\ell)-f_u(I)\tilde X^\ell .
\end{aligned}
\end{sequation}
This trajectory independence is already an important distinction from a
standard EKF, whose error dynamics are generally linearized about the current
state estimate.

To study these dynamics in
Euclidean coordinates, consider an invariant error sufficiently close to the
identity and express it using exponential coordinates as
\begin{sequation}\label{eq:exp-error:IEKF}
    \tilde X=\exp(\xi^\wedge),
    \qquad
    \xi\in\mathbb{R}^{\dim\mathfrak g},
\end{sequation}
where $\xi$ denotes the Lie-algebra coordinates of the invariant error and
$(\cdot)^\wedge:\mathbb{R}^{\dim\mathfrak g}\rightarrow\mathfrak g$ is the
wedge map. Linearizing the autonomous error dynamics about the identity
$\tilde X=I$, or equivalently about $\xi=0$, defines the matrix $A_u$ through
\begin{sequation}\label{eq:A:IEKF}
g_u\!\big(\exp(\xi^\wedge)\big)
=
\big(A_u\xi\big)^\wedge
+\mathcal O(\|\xi\|^2).
\end{sequation}
At first sight, this gives the usual first-order approximation
\begin{sequation}
    \dot\xi\approx A_u\xi.
\end{sequation}
For group-affine systems, however, a remarkable and much stronger result
holds.
\begin{theorem}[Log-linear error property {\cite{Barrau2017}}]
\label{thm:log-linear:IEKF}
Suppose that the initial invariant error satisfies
\[
    \tilde X(0)=\exp(\xi_0^\wedge).
\]
Let $\xi(t)$ be the solution of the linear differential equation
\begin{sequation}\label{eq:linear-error:IEKF}
    \dot\xi(t)=A_{u(t)}\xi(t),
    \qquad
    \xi(0)=\xi_0.
\end{sequation}
Then, for all times for which the solutions are defined,
\begin{sequation}\label{eq:loglinear:IEKF}
    \tilde X(t)=\exp\!\big(\xi(t)^\wedge\big).
\end{sequation}
\end{theorem}
The significance of this result goes beyond trajectory-independent
linearization. Although $A_{u(t)}$ is identified from a first-order expansion,
the solution of~\eqref{eq:linear-error:IEKF}, when mapped back to the group through the
exponential map, reproduces the exact nonlinear invariant-error evolution.
Thus, unlike in a standard EKF, the linear error system is not merely a local
approximation about the estimated trajectory. For group-affine systems, the
invariant error dynamics are independent of both $X(t)$ and $\hat X(t)$ and,
more strongly, their nonlinear evolution is exactly captured by the linear
dynamics of $\xi$. The matrix $A_{u(t)}$ may still depend on the known input
$u(t)$ and hence be time varying; crucially, it does not depend on the true or
estimated trajectory.

The log-linear property concerns the deterministic invariant error between
trajectories driven by the same input. In the stochastic InvEKF, process noise
is expressed in the selected invariant error coordinates and incorporated
into the corresponding covariance propagation.

\sdbarfig{%
\begin{tcolorbox}[
    colback=red!10,
    arc=2mm,
    boxrule=0pt,
    left=1mm,
    right=1mm,
    top=1mm,
    bottom=1mm
]
\centering
\begin{tikzpicture}[
    x=0.48cm,
    y=0.48cm,
    >=Stealth,
    font=\normalsize,
    blue arrow/.style={->,>=Stealth,thick,draw=figblue},
    black arrow/.style={->,>=Stealth,semithick,draw=black},
    panel label/.style={font=\normalsize}
]


\node[panel label] at (2.3,-1.5) {(a)};

\coordinate (I) at (1.05,2.00);

\draw[thick]
    (-0.9,0.3)
    .. controls (-0.05,1.65) and (0.55,2.30) .. (1.05,2.50)
    .. controls (2.10,2.85) and (3.70,2.60) .. (5.0,1.2)
    -- (3.4,-0.5)
    .. controls (1.8,0.9) and (0.3,0.7) .. (-0.9,0.3)
    -- cycle;

\draw[thin,gray!70]
    (0.20,1.55)
    .. controls (0.50,1.72) and (0.78,1.92) .. (I)
    .. controls (1.85,2.25) and (3.35,2.05) ..
    coordinate[pos=0.80] (exp_target)
    (4.50,1);

\node at (0,0) {$G$};

\draw[thick]
    (-1.00,1.15) --
    (0.713,4.30) --
    (3.913,4.30) --
    (2.20,1.15) -- cycle;
\node[anchor=south] at (2.25,4.5)
    {$T_I G=\mathfrak g$};

\draw[fill=black,thick] (I) circle (1.3pt);
\node[below=2pt] at (I) {$I$};

\node[anchor=east] (xi_rn) at (0.50,3.70)
    {$\xi\in\mathbb{R}^n$};

\node[anchor=west] (xi_g) at (2,3.70)
    {$\xi^\wedge\in\mathfrak g$};

\draw[black arrow]
    (xi_rn) -- (xi_g)
    node[midway,below=0pt] {$(\cdot)^\wedge$};

\coordinate (xi_hat_pt) at (2.8,3.4);

\draw[blue arrow]
    (I) -- (xi_hat_pt);

\draw[blue arrow,densely dashed]
    (2.8,3.4)
    .. controls (3.55,3.4) and (4.10,2.65) ..
    (exp_target)
    node[pos=0.53,right=2pt,black] {$\exp(\cdot)$};

\fill (exp_target) circle (1.3pt);

\node[below=2pt] at (exp_target)
    {$\tilde X$};


\node[panel label] at (9,-1.5) {(b)};

\node (xi0) at (7.6,4.2) {$\xi(0)$};
\node (xit) at (11.3,4.2) {$\xi(t)$};

\draw[black arrow]
    (xi0) -- (xit)
    node[midway,above=1pt] {$\dot{\xi}=A_u\xi$};

\node (X0) at (7.6,0.5) {$\tilde X(0)$};
\node (Xt) at (11.3,0.5) {$\tilde X(t)$};

\draw[black arrow, densely dashed,shorten <=3pt, shorten >=6pt]
    (xi0) -- (X0)
    node[pos=0.38,right=2pt] {$\exp((\cdot)^\wedge)$};

\draw[black arrow, densely dashed,shorten <=3pt, shorten >=6pt]
    (xit) -- (Xt);

\draw[thick]
    (6.5,0.5)
    .. controls (8.2,2.5) and (10.7,2.5) ..
    (12.3,0.5);

\draw[black arrow]
    (X0) -- (Xt)
    node[midway,below=4pt,black]
    {$\dot{\tilde X}=g_u(\tilde X)$};

\end{tikzpicture}
\end{tcolorbox}
}{%
\textbf{Geometric interpretation of the log-linear error property.}
(a) The invariant error $\tilde X\in G$ is represented by exponential
coordinates $\xi\in\mathbb{R}^{\dim\mathfrak g}$ through
$\tilde X=\exp(\xi^\wedge)$, where
$\xi^\wedge\in\mathfrak g=T_I G$.
(b) For group-affine dynamics, the linear evolution
$\dot{\xi}=A_u\xi$ maps through the exponential map exactly onto the nonlinear
invariant-error evolution $\dot{\tilde X}=g_u(\tilde X)$ on $G$.
\label{sfig:log-linear-error}
}
\end{sidebar}

\begin{sidebar}{\continuesidebar}
\setcounter{sequation}{8}
\renewcommand{\thesequation}{S\arabic{sequation}}
The same symmetry principle applies to measurements.
A measurement is \emph{right invariant} if $y=X^{-1}b$ and \emph{left invariant}
if $y=Xb$, where $b$ is a known constant vector.
The associated invariant innovations are
$z^r=\hat Xy-b$ and $z^\ell=\hat X^{-1}y-b$. {In implementation, identically zero components of the innovation may be
removed using a selection matrix $P$ (typically $P=[\,I\;\;0\,]$ for
special Euclidean groups and $P=I$ for the special orthogonal group).
For simplicity, we omit this truncation below to avoid unnecessary notation.}
Expressing the invariant error as $\tilde X=\exp(\xi^\wedge)=I+\xi^\wedge+\mathcal{O}(\|\xi\|^2)$, both cases admit
the expansion
\begin{sequation}\label{eq:H:IEKF}
z^\star = H\,\xi + \mathcal{O}(\|\xi\|^2),
\end{sequation}
with $H\xi=\xi^\wedge b$ for $z^r$ and $H\xi=-\,\xi^\wedge b$ for $z^\ell$.
{Importantly, in either case, $H$ depends only on the measurement geometry
and not on the state estimate $\hat X$. Thus, while the invariant measurement relation is still linearized locally,
its Jacobian is independent of the estimated trajectory. Combined with the
exact log-linear propagation of the invariant error, this yields the
trajectory-independent linearized error model underlying the InvEKF and
enables local stability results under suitable observability assumptions.}

{%
The symmetry of these invariant measurements can be exploited further in
the measurement update. In particular, the EqF$^\star$ construction
\cite{vanGoorHamelMahony2022} uses a symmetrized output approximation that,
for the right-invariant measurement considered above, yields
\[
    z^r
    =
    \frac{1}{2}\xi^\wedge\bigl(b+\hat Xy\bigr)
    +\mathcal{O}(\|\xi\|^3).
\]
Compared with~\eqref{eq:H:IEKF}, this symmetrization eliminates the
second-order approximation error while retaining a measurement relation
that is linear in the local error coordinates. This refinement is
complementary to the exact log-linear propagation of group-affine systems:
the former exploits the symmetry of the measurement, whereas the latter
follows from the structure of the dynamics. For the purposes of this
tutorial, however, we retain the original InvEKF output
approximation~\eqref{eq:H:IEKF} in the sequel.}

Algorithm~\ref{alg:invEKF} summarizes the resulting InvEKF equations for either
right- or left-invariant error definitions.
{
To retain the trajectory-independent measurement Jacobian, the invariant-error
convention should be matched to the corresponding measurement type: a
right-invariant error is naturally paired with right-invariant measurements,
and a left-invariant error with left-invariant measurements \cite{Barrau2017}. This matching
should therefore guide the choice between the right and left InvEKF.
When measurements of both types are available, one may choose the error
convention that matches the most frequent, accurate, or critical measurements \cite{Potokar2024}.
Measurements of the opposite type can still be incorporated by expressing
their linearization in the selected error coordinates; however, the resulting
Jacobians generally depend on $\hat X$, and the trajectory-independent
measurement structure is then lost for those measurements although the filter still retains the invariant formulation and the
log-linear propagation structure associated with the chosen error.}
\begin{algobox}{Invariant Extended Kalman Filter}
\label{alg:invEKF}
\begin{algorithmic}[1]
\STATE \textbf{Input:}
$\hat X(t_{k-1})\in G$, $\Sigma(t_{k-1})\in\mathbb{R}^{n\times n}$,
$n:=\dim\mathfrak g$.

\STATE \textbf{Error convention:}
{Select a right- or left-invariant error, preferably matching
the dominant measurement type.}

\STATE \textbf{Jacobians:}
{Compute $A_u$ from~\eqref{eq:A:IEKF}, and obtain $H$ by
linearizing the measurement innovation in the selected invariant-error
coordinates (using~\eqref{eq:H:IEKF} for matched invariant measurements).}
\vspace{3pt}
\STATE \textbf{Propagation for $t_{k-1}\le t<t_k$:}
\begin{align*}
\dot{\hat X}(t) &= f_u\!\big(\hat X(t)\big),\\
\dot{\Sigma}(t) &= A_u\,\Sigma(t)+\Sigma(t)\,A_u^\top+Q,
\end{align*}
where $Q$ is the process noise covariance.

\vspace{3pt}
\STATE \textbf{Correction at $t=t_k$:}
\begin{align*}
K &= \Sigma\,H^\top\,(H\,\Sigma\,H^\top+V)^{-1},\\[2pt]
\hat X^+ &=
\begin{cases}
\exp\!\big(-(Kz^r)^\wedge\big)\,\hat X, & \text{right-invariant error},\\[2pt]
\hat X\,\exp\!\big(-(Kz^\ell)^\wedge\big), & \text{left-invariant error},
\end{cases}\\[2pt]
\Sigma^+ &= (I-KH)\,\Sigma,
\end{align*}
where $V$ is the measurement noise covariance.

\vspace{2pt}
\STATE \textbf{Output:} $\hat X(t_k)=\hat X^+$, $\Sigma(t_k)=\Sigma^+$.

\end{algorithmic}
\end{algobox}
{%
The InvEKF is presented here in continuous-discrete form, with continuous-time
propagation and discrete measurement updates, following the original formulation
in~\cite{Barrau2017}. In practical implementations, the propagation equations
must be discretized consistently with the system dynamics and the chosen
numerical integration scheme; further details, including fully discrete
formulations such as \cite{Barrau2018}, are beyond the scope of this sidebar.}

The state estimate $\hat X$ is propagated directly on the Lie group using the
original system dynamics. At measurement instants, an invariant innovation is
formed, mapped to the Lie algebra, and lifted back to the group through the
exponential map, thereby preserving the underlying geometry.

The process and measurement noise covariances $Q$ and $V$ must correspond to the
additive noise terms in the linearized invariant-error dynamics and innovation,
respectively. Consequently, these covariances may represent \emph{effective}
(or modified) process and measurement noises, which depend on the chosen
invariant error definition and the measurement model, rather than the original
physical noise statistics.
\end{sidebar}

\begin{figure}[t]
    \centering
    \includegraphics[width=\linewidth]{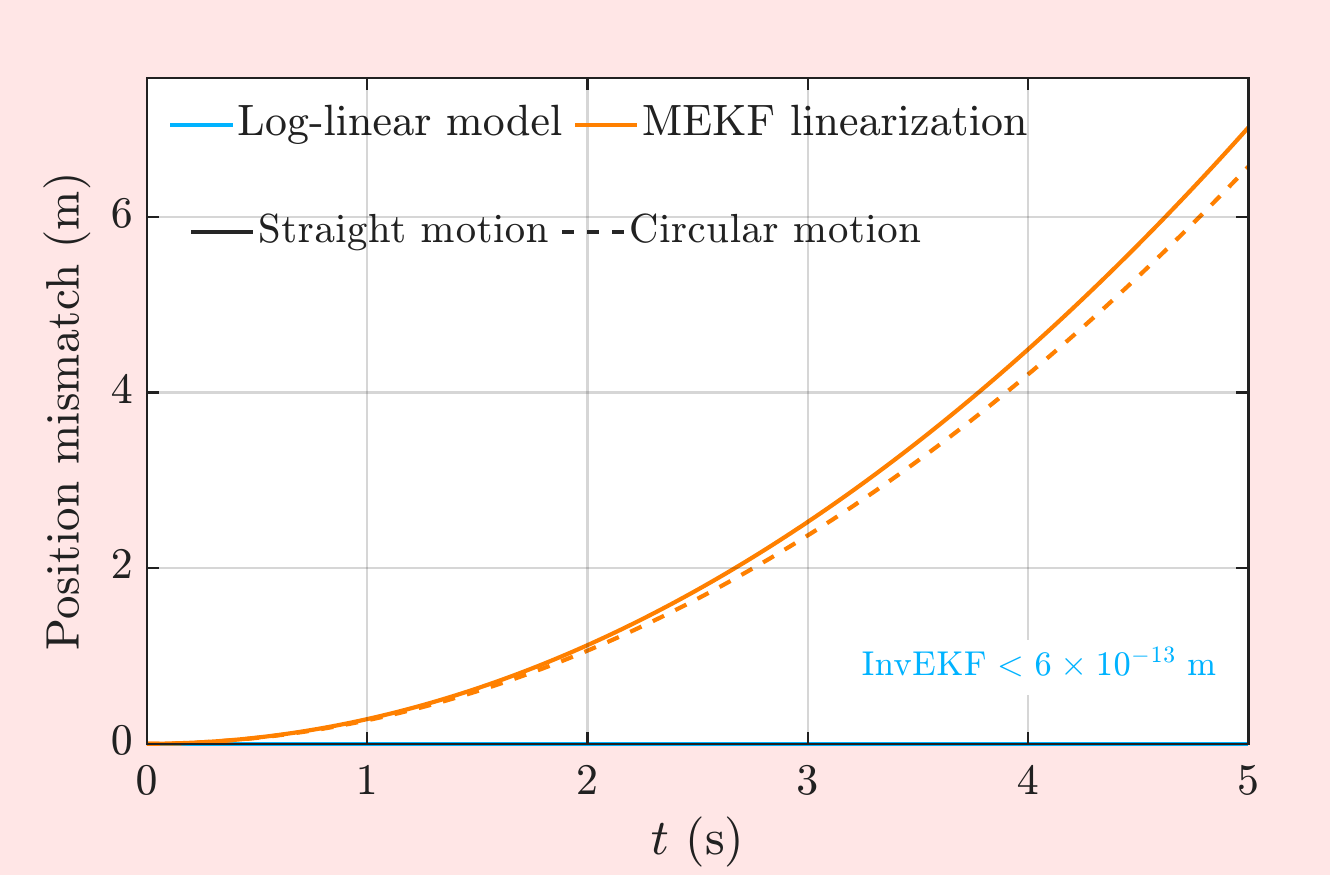}
    \caption{\textbf{Exact log-linear versus conventional linearized error
    propagation.}
    {Position mismatch between the exact nonlinear error and
    the corresponding error-propagation model for a $20^\circ$ initial
    attitude error. The invariant log-linear prediction agrees with the
    nonlinear error up to numerical precision, whereas the
    trajectory-dependent first-order model~\eqref{eq:mekf-error-propagation}
    exhibits an increasing mismatch. Solid and dashed lines denote straight
    and circular motion, respectively. No measurement updates are applied;
    the comparison concerns deterministic error propagation only.}}
    \label{fig:loglinear_example}
\end{figure}

\begin{figure*}[t] 
\begin{tcolorbox}[colback=red!10, arc=2mm, boxrule=0pt,
                  left=3mm, right=3mm, top=3mm, bottom=3mm]
\centering 
\includegraphics[width=0.85\linewidth]{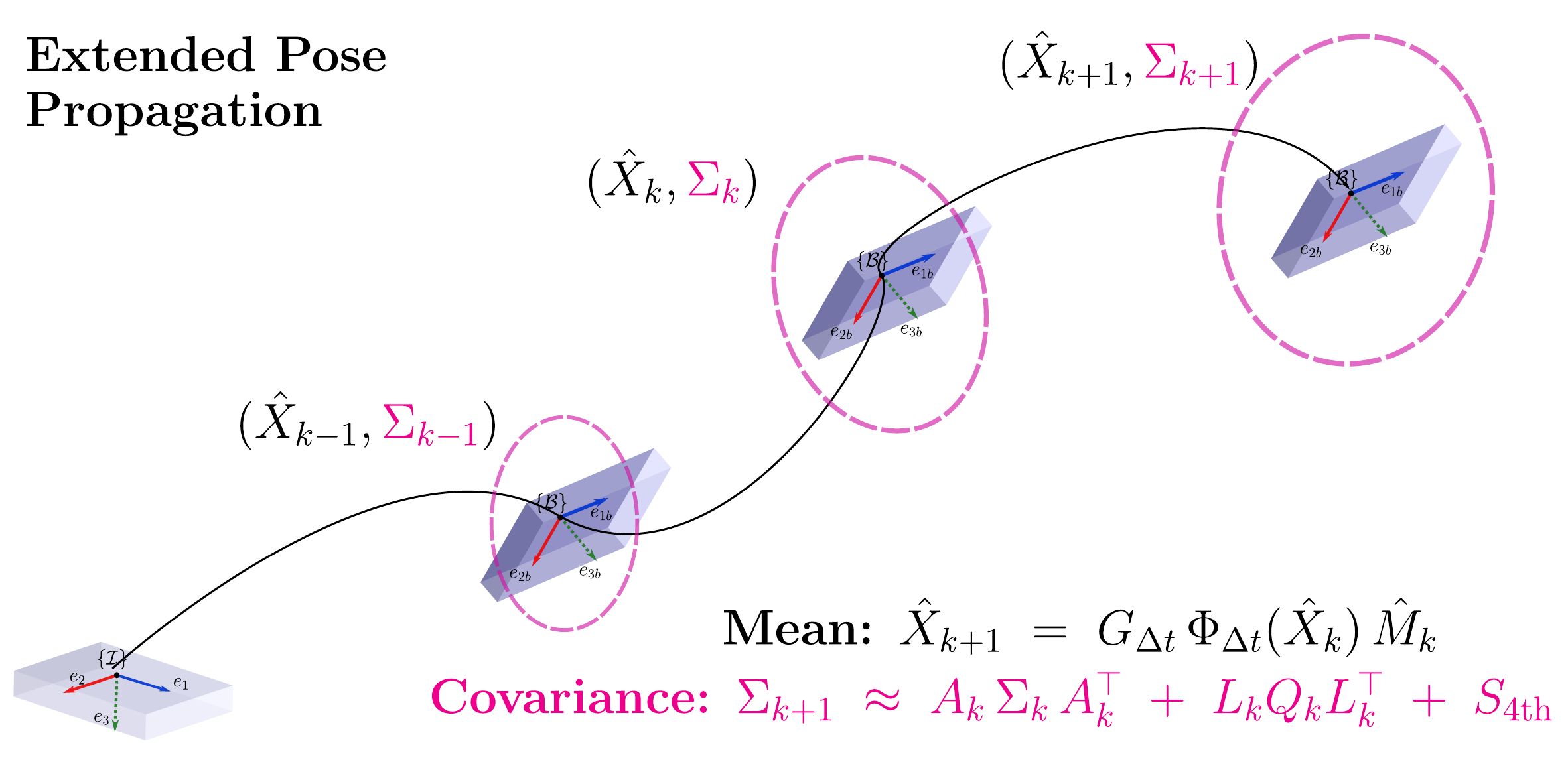} 
\end{tcolorbox}
\caption{\textbf{Extended-pose propagation on the Lie group $\SE$.} The mean estimate $\hat{X}_k$ evolves according to~\eqref{eq:hatX_update}, while the associated uncertainty follows~\eqref{eq:Sigma-next-4th-A}. Ellipsoids represent the error covariance, illustrating how uncertainty accumulates over successive IMU propagation steps.} 
\label{fig:propagation} 
\end{figure*}

\subsubsection{Error and Uncertainty Propagation}
Here we aim to study the propagation of the estimation error. 
The naive Euclidean error $X - \hat X$ is not geometrically consistent, since $X$ and $\hat X$ evolve on the Lie group $\SE$ while the error does not. 
Using such an error leads to linearized dynamics whose Jacobians depend explicitly on the unknown true trajectory, thereby introducing state-dependent gains and sensitivity to large initial errors—well-known limitations of classical Kalman filtering for nonlinear systems.
A more principled formulation defines the estimation error directly on the group, exploiting its intrinsic symmetries and yielding trajectory-independent dynamics.

For this purpose, two canonical error definitions are introduced
\begin{equation}
\tilde X^r := \hat X X^{-1}, 
\qquad 
\tilde X^\ell := X^{-1} \hat X,
\label{eq:inv_errors}
\end{equation}
referred to as the \emph{right-invariant} and \emph{left-invariant} errors, respectively. 
These definitions are termed invariant because $\tilde X^r$ (resp.\ $\tilde X^\ell$) remains unchanged when both $X$ and $\hat X$ are right-multiplied (resp.\ left-multiplied) by the same group element.
In inertial navigation, the right-invariant formulation is often preferred (although both are valid) because it leads to \textit{time-invariant autonomous} error dynamics. 
In the sequel, we focus on the right-invariant error and therefore write $\tilde X = \tilde X^r$ for simplicity.

{\paragraph{Deterministic Error Propagation}
We first consider propagation in the absence of IMU noise and bias
mismatch. In view of~\eqref{eq:ins_SE} and~\eqref{eq:strapdown_group},
the right-invariant error satisfies
}
\[
    \dot{\tilde X}=[\tilde X,D-G].
\]
{
These dynamics are autonomous and time-invariant because $D-G$ is
constant. Under the same assumptions, the left-invariant formulation
yields autonomous but generally input-dependent error dynamics.
}

\begin{pullquote}
A more principled formulation defines the estimation error directly on the group, exploiting its intrinsic symmetries and yielding trajectory-independent dynamics.
\end{pullquote}

{These properties are a consequence of the fact that the INS dynamics
considered here belong to the class of \emph{group-affine systems}
\cite{Barrau2017}. For such systems, the invariant error dynamics are
autonomous and, more remarkably, in the deterministic case, the logarithmic
coordinates of the invariant error satisfy an \emph{exact linear differential
equation}, a property known as the log-linear error property; see also
``\nameref{sidebar-InEKF}''. This is fundamentally different from the
trajectory-dependent local linearization underlying a classical EKF.}

Writing the error exactly as $\tilde X=\exp(\xi^\wedge)$, the log-linear
property implies that its logarithmic coordinates satisfy
\[
    \dot{\xi}^\wedge=[\xi^\wedge,D-G],
\]
without requiring a small-error approximation. To express these dynamics in vector coordinates, we first
identify the coordinate representation of the commutator with $D$.
This also reveals how the automorphism $\Phi_t$ introduced
in~\eqref{eq:Phi_automorphism} acts on the error coordinates. Let
$\bar D\in\mathbb{R}^{9\times9}$ be defined by
\begin{equation}
    (\bar D\xi)^\wedge=[\xi^\wedge,D].
    \label{eq:Dbar_definition}
\end{equation}
{
By the classical Lie group--Lie algebra homomorphism correspondence,
the differential of a smooth group automorphism at the identity is a
linear map on the Lie algebra that intertwines the exponential maps
\cite[Theorem~15]{barrau2019linear}. For $\Phi_t$, this differential
admits a matrix representation $F_t$ under the identification
$\xi \mapsto \xi^\wedge$, yielding
\begin{equation}
    \Phi_t\!\big(\exp(\xi^\wedge)\big)
    =
    \exp\!\big((F_t\xi)^\wedge\big),
    \qquad
    F_t:=\exp(t\bar D).
    \label{eq:Phi_lie_algebra}
\end{equation}
Thus, $F_t$ represents the action of the automorphism $\Phi_t$ in
exponential coordinates.
}
{
Using~\eqref{eq:Dbar_definition} and the adjoint identities
\eqref{eq:adjoint} and~\eqref{eq:small_adjoint}, the above error dynamics have
the exact vector representation
\begin{equation}
    \dot\xi=\big(\bar D+\ad_{G^\vee}\big)\xi.
    \label{eq:xi_deterministic}
\end{equation}
The explicit form of $\bar D$ is provided in
Table~\ref{tab:constant-matrices}.
}

{
The finite-time factorization~\eqref{eq:Xnew} makes the connection to
the automorphism framework explicit. Since the true and estimated
propagations share the same IMU increment in the deterministic case
considered here, this increment cancels from the right-invariant error:
\begin{equation}
    \tilde X_{k+1}
    =G_{\Delta t}\,\Phi_{\Delta t}(\tilde X_k)\,G_{\Delta t}^{-1}.
    \label{eq:deterministic-error-automorphism}
\end{equation}
This map is itself an automorphism. Combining~\eqref{eq:Phi_lie_algebra}
with the adjoint identity therefore gives an exact
exponential-coordinate representation of its propagation:
\begin{equation}
    \xi_{k+1}=A_k\xi_k,
    \qquad A_k=\Ad_{G_{\Delta t}}F_{\Delta t}.
    \label{eq:deterministic-error-transition}
\end{equation}
Thus, the discrete log-linear propagation follows directly from the
same automorphism structure as the state-flow factorization.
}
{
Equivalently, integrating~\eqref{eq:xi_deterministic} gives the same
deterministic transition matrix:
}
\begin{equation}
A_k
=
\exp\!\big((\bar D+\ad_{G^\vee})\Delta t\big)
=
\begin{bmatrix}
I_3 & 0 & 0\\[3pt]
g^\wedge\Delta t & I_3 & 0\\[3pt]
\tfrac{1}{2}g^\wedge\Delta t^2 & \Delta t\,I_3 & I_3
\end{bmatrix}.
\label{eq:Ak_explicit}
\end{equation}
{Notably, $A_k$ is independent of both the estimated
trajectory and the IMU measurements, directly reflecting the exact
log-linear structure of the deterministic invariant error dynamics.
Although $A_k$ is time-invariant in the present formulation (and its
subscript could therefore be dropped), we retain the notation $A_k$
throughout the tutorial for consistency and to facilitate extensions to
alternative error definitions, such as left-invariant errors, for which
the transition matrix may become time-varying.}

\begin{example}[Exact log-linear error propagation]
{
To illustrate that the noise- and bias-free linear time-invariant (LTI)
error dynamics in~\eqref{eq:xi_deterministic} describe the logarithmic invariant
error exactly, rather than as a first-order approximation, we consider two
noise- and bias-free INS motions over
$5\,\mathrm{s}$ with sampling period $\Delta t=0.01\,\mathrm{s}$.
Both start from
$R_0=I_3$, $v_0=[5,0,0]^\top\,\mathrm{m/s}$, and $p_0=0$.
The first is a straight, level, constant-speed motion, with
$\omega=0$ and $a=-g$. The second is the circular motion introduced
in Example \ref{ex:discretization}.
For each motion, the true and estimated states are propagated independently
using exactly the same IMU inputs. The estimated position and velocity are
initialized at their true values, while the attitude estimate has a
$20^\circ$ error about the unit axis
$u_0=[1,-1,0.5]^\top/\|[1,-1,0.5]^\top\|$.
No measurement updates are applied.}

{
For comparison, consider the conventional error representation underlying
the MEKF. Define the multiplicative attitude error
$
\tilde R:=\hat R R^\top
\approx I_3+\tilde\theta^\wedge,
$
together with the additive velocity and position errors
$
\tilde v:=v-\hat v, \tilde p:=p-\hat p,
$
and let
$\tilde x:=(\tilde\theta,\tilde v,\tilde p)\in\mathbb{R}^9$.
Linearization of the deterministic navigation error dynamics about the
estimated trajectory gives
\begin{equation}
\dot{\tilde x}
\approx
A_{\mathrm MEKF}(t)\tilde x,
\qquad
A_{\mathrm MEKF}(t):=
\begin{bmatrix}
0_3 & 0_3 & 0_3\\
(\hat R\hat a)^\wedge & 0_3 & 0_3\\
0_3 & I_3 & 0_3
\end{bmatrix},
\label{eq:mekf-error-propagation}
\end{equation}
where $\hat a$ denotes the bias-compensated specific force.
Unlike the invariant transition matrix~\eqref{eq:Ak_explicit},
$A_{\mathrm MEKF}(t)$ depends explicitly on the estimated trajectory and
provides only a first-order approximation of the nonlinear error dynamics.}

{
Figure~\ref{fig:loglinear_example} shows the mismatch between the position
component of the exact nonlinear error and that predicted by the
corresponding error-propagation model. The invariant log-linear model
reproduces the nonlinear error up to numerical precision for both motions.
In contrast, the trajectory-dependent first-order model
\eqref{eq:mekf-error-propagation} progressively departs from the nonlinear
error, reaching position mismatches of $7.02\,\mathrm{m}$ and
$6.57\,\mathrm{m}$ after $5\,\mathrm{s}$ for the straight and circular
motions, respectively. Importantly, the same invariant transition matrix
governs the logarithmic error for both motions, whereas
$A_{\mathrm MEKF}(t)$ varies with the estimated trajectory.}

\end{example}

{\paragraph{Effect of IMU Noise and Bias Mismatch}
We now introduce IMU noise and bias mismatch. The full right-invariant
error dynamics obtained from~\eqref{eq:ins_SE} and
\eqref{eq:strapdown_group} are
}
\begin{equation}
\dot{\tilde X}
= [\tilde X, D-G] + \BAd_{\hat X}(N-\tilde B)\,\tilde X,
\label{eq:tildeX_dyn}
\end{equation}
with $\tilde B=(\tilde b_\omega,\tilde b_a,0)^\wedge:=\hat B-B$.
{
Neglecting products of the small state and input errors gives
}
\[
    \dot{\xi}^\wedge
    \approx
    [\xi^\wedge,D-G]
    +
    \BAd_{\hat X}(N-\tilde B).
\]
{In vector coordinates, this becomes}
\begin{equation}
\dot\xi
\approx
\big(\bar D+\ad_{G^\vee}\big)\xi
+\Ad_{\hat X}\Pi_9^{6\top}(n-\tilde b),
\label{eq:xi_compact}
\end{equation}
where $\tilde b:=(\tilde b_\omega,\tilde b_a)$ and
$n:=(n_\omega,n_a)$. The projection matrix $\Pi_n^m$ is provided in
Table~\ref{tab:constant-matrices}.

The discrete-time propagation of the error coordinates can then be written as
\begin{equation}\label{eq:xi_update}
\xi_{k+1}
\approx
A_k\xi_k+L_k(n_k-\tilde b_k),
\end{equation}
{where $A_k$ is the deterministic transition matrix
in~\eqref{eq:Ak_explicit}.}
Under a zero-order approximation of the perturbation term over the
sampling interval, the corresponding injection matrix is
\begin{equation}
L_k
=
\Delta t\,\mathbf{J}\!\big((\bar D+\ad_{G^\vee})\Delta t\big)
\Ad_{\hat X_k}\Pi_9^{6\top}
\approx
\begin{bmatrix}
\Delta t\hat R_k & 0\\[4pt]
\Delta t\,\hat v_k^\wedge\hat R_k & \Delta t\hat R_k\\[4pt]
\Delta t\,\hat p_k^\wedge\hat R_k & 0
\end{bmatrix}.
\label{eq:Lk_explicit}
\end{equation}
Here, $\mathbf{J}(\cdot)$ denotes the integrated exponential matrix
function defined in~\eqref{eq:left-jacobian-exp}, and
$n_k := (n_{\omega,k},n_{a,k})$ denotes the discrete-time IMU noise sample
associated with continuous-time white noise processes of power spectral
densities $Q_\omega$ and $Q_a$, respectively.

The expressions~\eqref{eq:xi_update}--\eqref{eq:Lk_explicit} characterize
the discrete-time propagation of the invariant error coordinates and
their sensitivity to input perturbations. In the following, we exploit
this structure to examine the statistical properties of the propagation
model under simplifying assumptions.

{\paragraph{Uncertainty Propagation under IMU Noise}
\label{sec:biasfree}
We now analyze the statistical consistency of the propagation model
under IMU measurement noise, assuming perfect bias compensation,
i.e., $\tilde B=0$.}
In a stochastic setting, the discrete update~\eqref{eq:hatX_update} provides a consistent mechanism to propagate the mean of the extended pose.  
This property was established for $\mathbb{SE}(3)$ in~\cite{barfoot2014associating} and later generalized to $\SE$ in~\cite{brossard2022uncertainty}.  
It is equally important, however, to characterize the evolution of uncertainty under this mean propagation.

Accordingly, we now treat the invariant error $\xi_k$ as a random variable.  
By definition of the right-invariant error, the true extended pose at time~$t_k$ can be expressed as
\begin{equation}
X_k \;=\; \exp(-\xi_k^\wedge)\,\hat X_k,
\qquad 
\xi_k \sim \mathcal{N}(0,\Sigma_k),
\end{equation}
where $\xi_k\in\mathbb{R}^9$ denotes the zero-mean extended pose error and
$\Sigma_k\in\mathbb{R}^{9\times 9}$ its associated covariance.
Although $\xi_k$ is Gaussian in the Lie algebra, the induced distribution of the pose
$X_k$ on the group is generally non-Gaussian when expressed in Euclidean coordinates.
\begin{figure*}[t]
\centering
\begin{tcolorbox}[colback=red!10, arc=2mm, boxrule=0pt,
                  left=3mm, right=3mm, top=3mm, bottom=3mm]

\centering
\includegraphics[width=\linewidth]{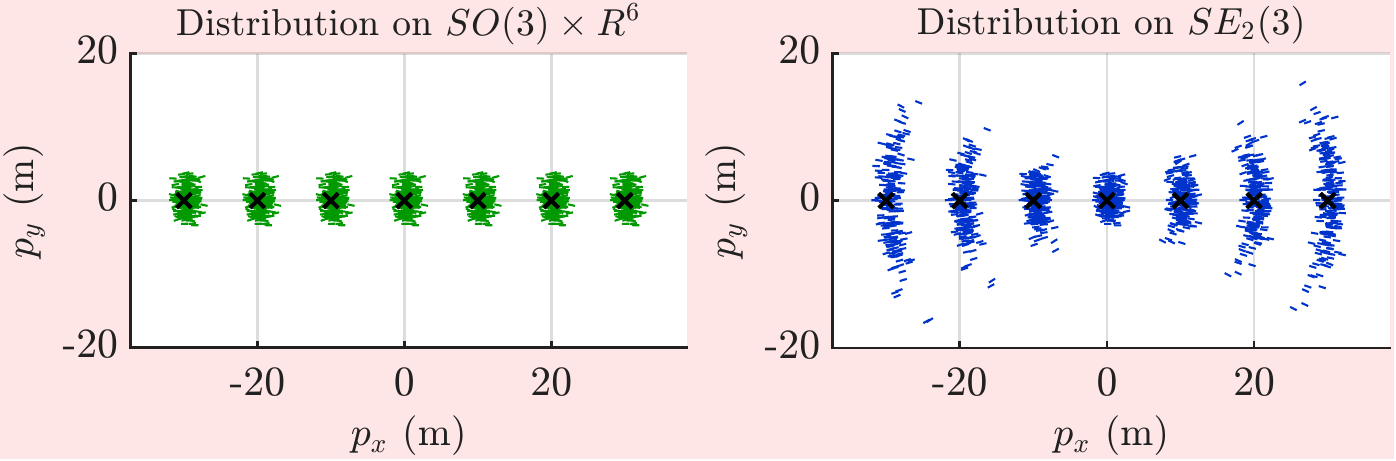}

\end{tcolorbox}
\caption{%
\textbf{Comparison of direct-product and right-invariant error distributions}.
Each panel shows multiple distributions with identical covariance and means located at
different positions along the $x$-axis.
For each mean pose, approximately $230$ Monte Carlo samples are drawn in the Lie algebra
from a zero-mean Gaussian distribution with standard deviations
$(10^\circ,\,0.8\,\mathrm{m},\,1.6\,\mathrm{m})$ on the yaw--$x$--$y$ components and mapped to the group.
The position components are shown in the $x$--$y$ plane, while the yaw component is encoded
by the orientation of each line segment.
\emph{Left:} direct-product error model on $\SO\times\mathbb{R}^6$ \cite{forster2016manifold}, which produces
ellipsoidal position uncertainty independent of the distance from the origin.
\emph{Right:} right-invariant error model on $\SE$, where the nonlinear group mapping
yields curved (banana-shaped) distributions whose spread increases with the
distance from the origin due to the presence of the $\hat p_k\times\delta\phi_k$ term in \eqref{eq:pk:distribution}.
}
\label{fig:banana_growth}
\end{figure*}

To gain insight into the geometric effect induced by the right-invariant perturbation,
consider writing
$\xi_k = (\delta\phi_k,\delta v_k,\delta p_k)\in\mathbb{R}^9$,
with $\delta\phi_k\in\mathbb{R}^3$ and $\delta p_k\in\mathbb{R}^3$ denoting the rotational
and positional components, respectively. A first-order approximation of the position yields
\begin{equation}\label{eq:pk:distribution}
    p_k \;\approx\; \hat p_k - \delta p_k +  \hat p_k\times\delta\phi_k.
\end{equation}
The additional term $\hat p_k\times\delta\phi_k$ reveals a coupling between orientation
and position induced by the group action: small orientation perturbations generate
translational errors whose magnitude depends on mean position $\hat p_k$.
In contrast, under a direct-product error model on $\SO\times\mathbb{R}^6$ the position evolves simply as
\[
p_k = \hat p_k - \delta p_k ,
\]
leading to uncertainty regions that, for this distribution, are independent of rotational uncertainty.
This geometric distinction is illustrated in Fig.~\ref{fig:banana_growth}, where the
resulting uncertainty on $\SE$ takes a curved, \textit{banana-like shape}.

Such banana-shaped dispersion patterns have been previously observed in mobile robot localization under sensor noise~\cite{Thrun2000,Long2012}. These effects are accurately captured by Gaussian distributions expressed in Lie exponential coordinates on $\mathbb{SE}(2)$~\cite{Chirikjian2009}. Related formulations for uncertainty propagation on $\mathbb{SE}(3)$ have been developed in~\cite{WangChirikjian2008,barfoot2014associating}, and more recently extended to the Lie group $\SE$ in~\cite{brossard2022uncertainty}.

The effect of IMU measurement noise $n_k$ on the measured IMU increment $\hat M_k$
is modeled as
\begin{equation}\label{eq:imu_increment_noise}
M_k \;=\; \exp(-\eta_k^\wedge)\,\hat M_k,
\end{equation}
where $M_k\in\SE$ denotes the true IMU increment and
$\eta_k^\wedge\in\mathfrak{se}_2(3)$ represents a small perturbation in the Lie algebra.
The following result characterizes the perturbation $\eta_k$.
{\begin{lemma}[IMU increment noise and error composition]
\label{lem:imu-increment-noise}
Assume a bias-free IMU with additive measurement noise, and let
$n_k := (n_{\omega,k},n_{a,k})$ denote the discrete-time IMU noise
sample. Then, to first order in the noise and under the small-sampling-interval
approximation, the true IMU increment $M_k$ admits the representation
\eqref{eq:imu_increment_noise} with $\eta_k = \bar L_k\,n_k$, where the
noise-injection matrix $\bar L_k$ is given by
\begin{equation}
\bar L_k=
\mathcal{J}_\ell\!\big(\log(\hat M_k)^\vee\big)
\begin{bmatrix}
\Delta t\,I & 0\\[2pt]
0 & \Delta t\,I\\[2pt]
0 &
J_\ell^{-1}(\omega_m\Delta t)
Q_\ell(\omega_m\Delta t)\Delta t^2
\end{bmatrix}.
\label{eq:Gk}
\end{equation}
Moreover, combining this increment perturbation with the exact
discrete group propagation yields
\begin{equation}
\exp(-\xi_{k+1}^\wedge)
=
\exp\!\big(-(A_k\xi_k)^\wedge\big)
\exp\!\big(-(L_k^{\mathrm g}n_k)^\wedge\big),
\label{eq:error_composition}
\end{equation}
where $A_k$ is given in~\eqref{eq:Ak_explicit} and
$
L_k^{\mathrm g}
:=
\Ad_{G_{\Delta t}}
\Ad_{\Phi_{\Delta t}(\hat X_k)}
\bar L_k$.
\end{lemma}}

{{Lemma~\ref{lem:imu-increment-noise} makes explicit the
connection between the discrete group propagation and the first-order
error model~\eqref{eq:xi_update}: the first factor in
\eqref{eq:error_composition} propagates the existing state error through
$A_k$, whereas the second injects the new IMU uncertainty through
$L_k^{\mathrm g}$. For small sampling intervals,
$L_k^{\mathrm g}$ is approximated by the injection matrix $L_k$ in
\eqref{eq:Lk_explicit}. Importantly, however, the two perturbations
compose multiplicatively on the group rather than additively in
exponential coordinates, see also \cite{barrau2020mathematical}.}}

{A Baker--Campbell--Hausdorff analysis of
\eqref{eq:error_composition} shows that, when $\xi_k$ and $n_k$ are
independent and zero-mean, the propagated invariant error remains
zero-mean up to third order, while higher-order terms introduce
corrections to its covariance
\cite{barfoot2014associating,brossard2022uncertainty}. Retaining terms
up to fourth order gives
\begin{equation}
\label{eq:Sigma-next-4th-A}
\Sigma_{k+1}
\approx
A_k\Sigma_kA_k^\top
+
L_kQ_kL_k^\top
+
S_{\mathrm{4th}},
\end{equation}
where
$Q_k:=\mathrm{diag}(Q_\omega/\Delta t,Q_a/\Delta t)$
is the covariance of the discrete IMU noise sample $n_k$, with
$Q_\omega$ and $Q_a$ denoting the corresponding continuous-time noise
power spectral densities. The matrix $S_{\mathrm{4th}}$ collects the
fourth-order corrections arising from the nonlinear composition of the
state and IMU-induced perturbations. In the second-order covariance
propagation adopted throughout this tutorial, these higher-order
corrections are neglected, i.e., $S_{\mathrm{4th}}=0$.}

\begin{figure}[t]
\centering
\begin{tcolorbox}[colback=red!10, arc=2mm, boxrule=0pt,
                  left=3mm, right=3mm, top=3mm, bottom=3mm]
\centering
\includegraphics[width=\linewidth]{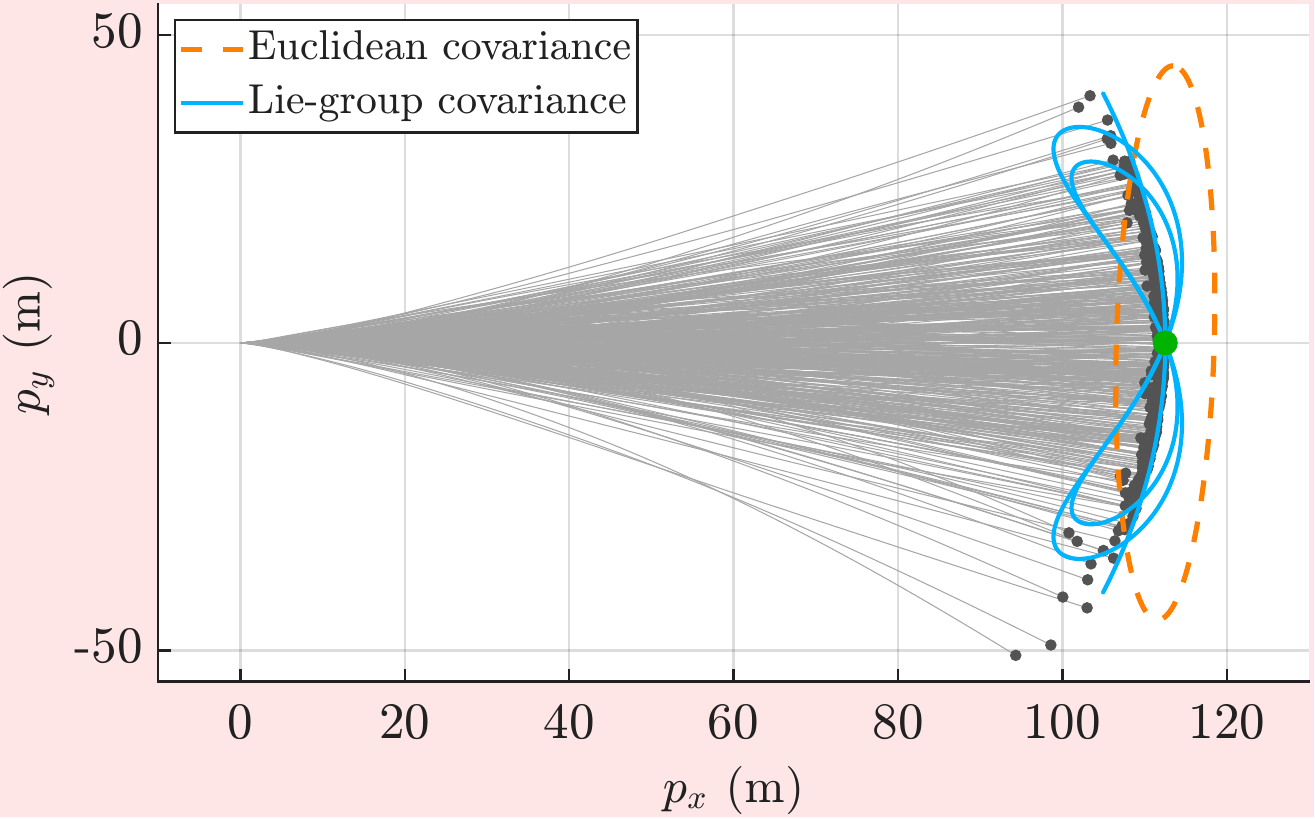}
\end{tcolorbox}
\caption{Simple propagation on $\SE$ with
$(\hat X_0,\Sigma_0)=(I_5,0_{9\times9})$,
$(\omega,a)=(0_3,e_1-g)$, sampling period
$\Delta t=0.05\,\mathrm{s}$, and propagation duration $15\,\mathrm{s}$.
{The gyroscope is corrupted by discrete-time noise along the $z$-axis
with standard deviation $\sigma_{\omega_z}=20^\circ/\mathrm{s}$ per
sample, while accelerometer noise is set to zero.}
Black dots show $500$ Monte Carlo endpoints in the $(p_x,p_y)$ plane,
while the filled green marker indicates the mean endpoint.
The blue curves represent projected Lie-group covariance contours
(principal great circles of the $3\sigma$ covariance ellipsoid in the
Lie algebra, mapped to the group and projected onto the $(p_x,p_y)$
plane).
The dashed orange curve is the fitted Euclidean $3\sigma$ covariance
ellipse obtained from the sample covariance of the endpoints in
$\mathbb{R}^2$.}
\label{fig:simple_prop_mc_vs_cov}
\end{figure}

{In summary, IMU measurements propagate the \emph{mean} extended pose
through~\eqref{eq:hatX_update}, while the associated uncertainty evolves
according to~\eqref{eq:Sigma-next-4th-A}, with $S_{\mathrm{4th}}=0$ in
the approximation adopted here.
This interplay between mean and covariance propagation is illustrated in
Fig.~\ref{fig:simple_prop_mc_vs_cov}, which compares Monte Carlo realizations
of the stochastic strapdown dynamics with the covariance predicted by the
invariant error model.
Despite the second-order covariance approximation, the resulting
Lie-group uncertainty already captures the characteristic curved,
\emph{banana-shaped} dispersion observed in the samples.
This effect is a direct consequence of representing uncertainty in exponential
coordinates of $\SE$.
More refined features of the uncertainty distribution, such as the finite
spread in the straight-ahead direction, require higher-order covariance
corrections, as discussed in~\cite{brossard2022uncertainty}.}

\paragraph{Imperfect Bias Compensation}
\label{sec:bias-propagation}

The derivation in Section "\nameref{sec:biasfree}" assumed that the IMU
measurements were perfectly compensated for bias, yielding a
bias-free propagation on~$\SE$.  
In practice, however, gyroscope and accelerometer biases can only be
estimated approximately and tend to drift slowly over time.
These residual errors slightly perturb the propagation model and lead to
a coupling between the pose uncertainty and the bias uncertainty.

To capture this effect, we augment the Lie-algebra error state $\xi_k$ with the
bias estimation error $\tilde b_k$ whose dynamics are
\[
\tilde b_{k+1} = \tilde b_k -\Delta t\,n_{b,k},
\]
where $n_{b}:=(n_{b_\omega},n_{b_a})$.
The overall error vector then becomes
\[
\zeta_k :=
\begin{bmatrix}
\xi_k\\[2pt]\tilde b_k
\end{bmatrix}
\in \mathbb{R}^{15},
\]
combining the extended-pose error~$\xi_k\in\mathbb{R}^9$
with the bias error. Using the discrete model \eqref{eq:xi_update} derived earlier
and the bias random-walk model above, the joint error propagation reads
\begin{equation}
\zeta_{k+1}
\;\approx\;
\underbrace{\begin{bmatrix}
A_k & -L_k\\[2pt]
0 & I
\end{bmatrix}}_{\displaystyle A_k^{\mathrm{ext}}}
\zeta_k
\;+\;
\underbrace{\begin{bmatrix}
L_k & 0\\[2pt]
0 & -\Delta t\,I
\end{bmatrix}}_{\displaystyle L_k^{\mathrm{ext}}}
\begin{bmatrix}
n_k\\[2pt] n_{b,k}
\end{bmatrix}.
\label{eq:xi_bias_augmented}
\end{equation}
The corresponding covariance propagation becomes
\begin{equation}
\begin{aligned}
\Sigma_{k+1}^{\mathrm{ext}}
&=
A_k^{\mathrm{ext}}\,
\Sigma_k^{\mathrm{ext}}\,
A_k^{\mathrm{ext}\top}
+
L_k^{\mathrm{ext}}\,
Q_k^{\mathrm{ext}}\,
L_k^{\mathrm{ext}\top},
\\
Q_k^{\mathrm{ext}}
&=
\frac{1}{\Delta t}
\mathrm{diag}
\big(
Q_\omega,\,
Q_a,\,
Q_{b_\omega},\,
Q_{b_a}
\big).
\label{eq:Sigma_bias_augmented}
\end{aligned}
\end{equation}
This augmented formulation extends the bias-free propagation by
accounting for imperfect IMU calibration while preserving the same
geometric structure for the pose dynamics on~$\SE$.

\begin{figure*}[t]
\begin{tcolorbox}[
    colback=red!10,
    arc=2mm,
    boxrule=0pt,
    left=3mm,
    right=3mm,
    top=3mm,
    bottom=3mm
]
\centering
\includegraphics[width=0.85\linewidth]{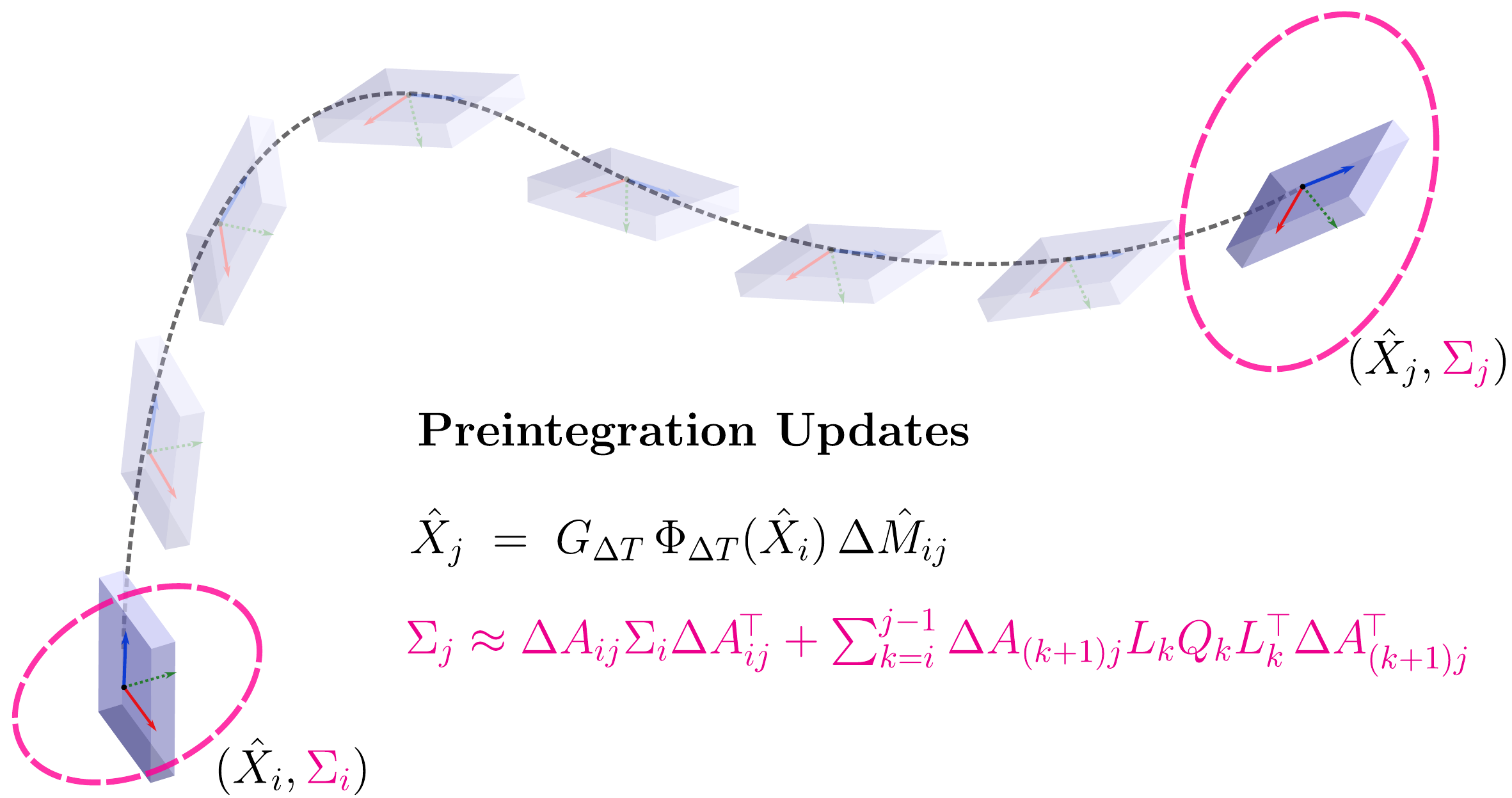}
\end{tcolorbox}
\caption{\textbf{Batch propagation between keyframes.}
The figure illustrates the propagation of the extended state
$(\hat X_i,\Sigma_i)$ through high-rate IMU integration to obtain the
keyframe estimate $(\hat X_j,\Sigma_j)$.
Equations~\eqref{eq:Xj_batch} and~\eqref{eq:Sigma_batch} summarize the
second-order propagation of the mean and covariance over the interval
$[t_i,t_j]$, accounting for all IMU measurements between the two keyframes.}
\label{fig:preintegration}
\end{figure*}

\subsection{Preintegration of IMU Measurements Over a Time Interval}
{When the state is updated only at sparse instants (e.g., keyframes), the
high-rate IMU effects over each interval can be accumulated directly on the
group. This yields the \emph{preintegrated increments} widely used in
visual--inertial odometry and factor-graph
SLAM~\cite{forster2016manifold,eckenhoff2019closed}.
Their geometric interpretation follows from the preintegration theory
in~\cite{barrau2019linear}: composing the
finite-time flows in~\eqref{eq:hatX_update} preserves their separation into an automorphism
acting on the initial state and initial-state-independent increments.
Consequently, these increments can be reused when the initial navigation
state is changed, provided that the bias-compensated input history is held
fixed. Beyond optimization-based estimators, preintegration is also useful
for filters operating at a lower update rate while still leveraging all
IMU samples.}
\begin{lemma}[Group preintegration factorization]\label{lem:group-preint}
Consider the one--sample update on $\SE$ given in~\eqref{eq:hatX_update}.
For integers $i<j$, define the time interval $\Delta T \triangleq (j-i)\Delta t$
and introduce the cumulative preintegration factor $\Delta\hat M_{ij}\in\SE$
recursively as
\begin{equation}\label{eq:recursion:Mij}
    \Delta\hat M_{i(k+1)}
\;=\;
\Phi_{\Delta t}\!\big(\Delta\hat M_{ik}\big)\,\hat M_k,
\end{equation}%
with 
$\Delta\hat M_{ii}=I_5$ and for all $k=i,\ldots,j-1$.
Then the state update between $\hat X_i$ and $\hat X_j$ admits the factorized form
\begin{equation}\label{eq:Xj_batch}
\hat X_j
\;=\;
G_{\Delta T}\,\Phi_{\Delta T}(\hat X_i)\,\Delta\hat M_{ij}.
\end{equation}
\end{lemma}
{The factorization in
Lemma~\ref{lem:group-preint} follows naturally from the automorphism
properties of $\Phi_t$ established in~\eqref{eq:Phi_properties}. The factor \(G_{\Delta T}\,\Phi_{\Delta T}(\hat X_i)\) carries the initial state and gravity contribution, whereas \(\Delta\hat M_{ij}\) depends only on the prescribed bias-compensated IMU history and the sampling times. Changing the bias estimates generally changes this history and requires updating the preintegrated factor.}

Moreover, unfolding the recurrence purely on the group gives the closed form
\begin{equation}\label{eq:DeltaX_prod}
\Delta\hat M_{ij}
\;=\;
\prod_{k=i}^{j-1}\Big(\Phi_{(j-1-k)\Delta t}(\hat M_k)\Big),
\end{equation}
{where the ordering (right-hand accumulation) follows the temporal sequence of the IMU increments.} Besides, writing
$\Delta\hat M_{ij}=\Gamma(\Delta R_{ij},\Delta v_{ij},\Delta p_{ij})$
and invoking \eqref{eq:barR}--\eqref{eq:barp} together with
\eqref{eq:DeltaX_prod}, we obtain the IMU motion increments
\begin{align}\label{eq:DRij}
\Delta R_{ij} &= \prod_{k=i}^{j-1} \exp\!\big(\hat\omega(t_k)^\wedge \Delta t\big), \\
\Delta v_{ij} &= \sum_{k=i}^{j-1} \Delta R_{ik}\,J_\ell\!\big(\hat\omega(t_k)\Delta t\big)\hat a(t_k)\,\Delta t, \\
\Delta p_{ij} &= \sum_{k=i}^{j-1} \left( \Delta v_{ik}\,\Delta t 
+ \Delta R_{ik}\,Q_\ell\!\big(\hat\omega(t_k)\Delta t\big)\,\hat a(t_k)\,\Delta t^2 \right),\label{eq:Dpij}
\end{align}
where $\Delta R_{ik}$ and $\Delta v_{ik}$ are defined analogously over $[t_i,t_k]$.

Substituting these blocks into the factorization of Lemma~\ref{lem:group-preint} gives the par-block preintegration formula:
\begin{align}\label{eq:Rj}
\hat R_j &= \hat R_i\,\Delta R_{ij}, \\
\hat v_j &= \hat v_i + g\,\Delta T + \hat R_i\,\Delta v_{ij}, \\
\hat p_j &= \hat p_i + \hat v_i \Delta T + \tfrac{1}{2} g \Delta T^2 + \hat R_i\,\Delta p_{ij}. \label{eq:pj}
\end{align}
{
These are the familiar component-wise preintegration relations
\cite{forster2016manifold,eckenhoff2019closed}, whose common group-level
structure and initial-state independence can also be understood through
the automorphism formulation in~\cite{barrau2019linear}.
Here, $J_\ell$ and $Q_\ell$ yield exact per-sample increments under the
stated zero-order-hold input model, whereas the preintegration
factorization \eqref{eq:Rj}--\eqref{eq:pj} itself holds more generally.
}

{
To propagate uncertainty over the same keyframe interval, we extend
the one-step group-error composition in
Lemma~\ref{lem:imu-increment-noise} to an arbitrary number of IMU samples.
The automorphism structure of $\Phi$ allows the successive error
perturbations to be transported through the discrete propagation and
accumulated multiplicatively on the group. This yields the following
result, closely related to~\cite[Proposition~5]{barrau2020mathematical}.
}
{
\begin{proposition}[Multiplicative error accumulation]
\label{prop:error-accumulation}
Under the per-sample IMU perturbation model of
Lemma~\ref{lem:imu-increment-noise}, let $i<j$ and define
$\Delta A_{ab}:=A_{b-1}\cdots A_a, \Delta A_{aa}:=I_9$.
Then the invariant error accumulated between $t_i$ and $t_j$ satisfies
\begin{multline*}
\exp(-\xi_j^\wedge)
=
\exp\!\big(-(\Delta A_{ij}\xi_i)^\wedge\big)
\\\prod_{k=i}^{j-1}
\exp\!\big(
-(\Delta A_{(k+1)j}L_k^{\mathrm g}n_k)^\wedge
\big),
\end{multline*}
where the product is ordered from left to right with increasing $k$.
\end{proposition}}

{This result is noteworthy because it provides a
closed-form description of the accumulation of the nonlinear invariant
error over an arbitrary number of propagation steps. Although the
composition remains nonlinear in exponential coordinates, it also
provides a direct route to statistical approximations of the propagated
error~\cite{barrau2020mathematical}. In particular, applying the BCH
formula to first order, together with $L_k^{\mathrm g}\approx L_k$,
gives}
\begin{equation}
\xi_j
\approx
\Delta A_{ij}\,\xi_i
+
\sum_{k=i}^{j-1}
\Delta A_{(k+1)j}\,L_k n_k.
\label{eq:xi_unrolled_noise}
\end{equation}

{Assuming that $\xi_i$ and the IMU noise samples are
zero-mean and mutually independent, the covariance at keyframe $j$ is
therefore}
\begin{equation}\label{eq:Sigma_batch}
\Sigma_j
\approx
\Delta A_{ij}\Sigma_i\Delta A_{ij}^{\!\top}
+
\sum_{k=i}^{j-1}
\Delta A_{(k+1)j}L_kQ_kL_k^\top
\Delta A_{(k+1)j}^{\!\top}.
\end{equation}
This expression provides a consistent second-order approximation of the
{uncertainty propagated between keyframes.}
In practice, the {transition products
$\Delta A_{(k+1)j}$ can be accumulated recursively with
$\mathcal{O}(j-i)$ computational cost.}
Higher-order corrections
{, such as} the fourth-order term $S_{\text{4th}}$
{, could also be included but are omitted here for clarity.}

\medskip
\noindent\textit{Remark.}
If IMU biases are not perfectly compensated, the same formulation can be extended
by augmenting the error state with the bias error and using the block-triangular
propagation matrix introduced in~\eqref{eq:xi_bias_augmented}.
The resulting covariance between keyframes then follows
\eqref{eq:Sigma_batch} with $A_k, L_k$ and $Q_k$ replaced by their augmented counterparts.
This provides a straightforward way to account for imperfect bias compensation
without altering the underlying preintegration framework.

\medskip
\noindent{\textit{Remark.}
The general form~\eqref{eq:Sigma_batch} accommodates time-varying
transition matrices, as may arise with augmented states or alternative
error definitions. For the right-invariant error considered here,
however, $A_k=A(\Delta t)$ is state-independent and satisfies the
semigroup property
$A(\Delta t)^n=A(n\Delta t)$.
Consequently, the transition factors $\Delta A_{ab}$ depend only on the
elapsed time and can be evaluated directly, without explicitly forming
products of the intermediate transition matrices.}



\section{Part II: Fusion of Aiding Measurements and Invariant Filtering on $\SE$}

Part~I of this tutorial addressed inertial propagation: given high-rate IMU
measurements, we showed how to evolve the nominal state
$(\hat R,\hat v,\hat p)$ on the Lie group $\SE$ and how to propagate the
associated uncertainty.
While essential, inertial propagation alone is insufficient for sustained
navigation, as IMU noise and unmodeled effects inevitably lead to drift.
To mitigate this behavior, inertial navigation systems incorporate additional
sensor information through measurement updates.

In this part, we focus on the fusion of such \emph{aiding measurements}.
These measurements provide partial,
sensor-dependent information that constrains specific components of the state
and its uncertainty.
Our goal is to place the measurement update step on the same geometric footing
as propagation by expressing observation models directly in terms of the
invariant error variables introduced earlier.

This perspective leads to an innovation model that is consistent with the
underlying Lie-group structure and naturally yields the correction equations
of the invariant extended Kalman filter (InvEKF) \cite{Barrau2017}.
We first derive a generic update for arbitrary observation models.
We then specialize the framework to right-invariant and left-invariant
measurements, before illustrating how common aiding sensors fit within this
setting (see Table \ref{tab:aiding-invariance}).
\begin{figure}[t]
\centering
\begin{minipage}{\columnwidth}
\begin{algobox}{Right-Invariant EKF (RIEKF) on $\SE$}
\label{alg:rieKF-biasfree}
\begin{algorithmic}[1]

\STATE \textbf{Prior state and covariance:}
$\hat{X}_k=\Gamma(\hat R_k,\hat v_k,\hat p_k)\in\SE$,
$\Sigma_k\in\mathbb{R}^{9\times9}$.

\STATE \textbf{Inputs:}
aiding measurement $y_k$ at time $t_k$;
IMU $\hat{U}_k := (\hat{\omega}_k,\hat{a}_k,0)^\wedge
\in\se$, held over $[t_k,t_{k+1})$;
gravity $g\in\mathbb{R}^3$ and timestep
$\Delta t=t_{k+1}-t_k$.

\vspace{4pt}
\STATE \textbf{Measurement and linearization:}
\[
z_k := y_k-h(\hat{X}_k)
\approx H_k\xi_k+\nu_k,
\qquad
\nu_k\sim\mathcal{N}(0,V_k),
\]
\[
H_k :=
\left.
\frac{\partial}{\partial\xi}
h\!\big(\exp(-\xi^\wedge)\hat{X}_k\big)
\right|_{\xi=0}.
\]

\vspace{4pt}
\STATE \textbf{Kalman gain and correction on the group:}
\begin{align*}
K_k &=
\Sigma_k H_k^\top
\big(H_k\Sigma_k H_k^\top+V_k\big)^{-1},\\[3pt]
\hat{X}_k^+ &=
\exp\!\big(-(K_kz_k)^\wedge\big)\hat{X}_k,\\[3pt]
\Sigma_k^+ &=
(I_9-K_kH_k)\Sigma_k.
\end{align*}
Write
$\hat{X}_k^+=\Gamma(\hat R_k^+,\hat v_k^+,\hat p_k^+)$.

\vspace{4pt}
\STATE \textbf{Compute increments and propagation matrices:}
\begin{align*}
&\text{Gravity:}\quad
G_{\Delta t}:=
\Gamma\!\left(
I_3,\,g\Delta t,\,\tfrac12g\Delta t^2
\right),\\[3pt]
&\text{Kinematics:}\quad
\Phi_{\Delta t}(\hat{X}_k^+):=
\Gamma\!\left(
\hat R_k^+,\,\hat v_k^+,\,\hat p_k^++\hat v_k^+\Delta t
\right),\\[3pt]
&\text{IMU:}\quad
\hat{M}_k:=
\exp\!\big(\Delta t\,\hat{U}_k\big)
+\Delta t\,
\big(\mathbf{J}(\Delta t\,\hat{U}_k)-I_5\big)D.
\end{align*}

\noindent Noise-injection matrix, evaluated after correction:
\[
L_k=\Delta t
\begin{bmatrix}
\hat R_k^+ & 0\\[3pt]
(\hat v_k^+)^\wedge\hat R_k^+ & \hat R_k^+\\[3pt]
(\hat p_k^+)^\wedge\hat R_k^+ & 0
\end{bmatrix}.
\]
The transition matrix $A_k$ is given in
\eqref{eq:Ak_explicit} and is constant for fixed $\Delta t$.

\vspace{4pt}
\STATE \textbf{Prediction to the next measurement time:}
\begin{align*}
\hat{X}_{k+1}
&=G_{\Delta t}\,
\Phi_{\Delta t}(\hat{X}_k^+)\,\hat{M}_k,\\[3pt]
\Sigma_{k+1}
&=A_k\Sigma_k^+A_k^\top+L_kQ_kL_k^\top,
\end{align*}
where
\[
Q_k=\frac{1}{\Delta t}\,
\operatorname{diag}(Q_\omega,Q_a)
\]
is the covariance of the IMU
noise sample. 

\vspace{4pt}
\STATE \textbf{Output:}
Next prior $(\hat{X}_{k+1},\Sigma_{k+1})$.

\end{algorithmic}
\end{algobox}
\end{minipage}
\end{figure}

\subsection{Generic Invariant Correction Step}
The goal of the measurement update is to incorporate the information contained in an aiding sensor output~$y_k$ into the current estimate~$\hat X_k$ and its covariance. 
We consider a general measurement model of the form
\begin{equation}
y_k = h(X_k) + \nu_k,
\qquad
\nu_k \sim \mathcal{N}(0,\,V_k),
\end{equation}
where $h:\SE\!\to\!\mathbb{R}^m$ is a smooth sensor function and $\nu_k$ is zero-mean measurement noise with covariance~$V_k$. 
To keep the update compatible with the Lie-group structure, the linearization is carried out not in Euclidean coordinates but directly with respect to the invariant error variable defined in Part~I.

Specifically, recall that the true state can be written as
\[
X_k = \exp(-\xi_k^\wedge)\,\hat X_k,
\]
where $\xi_k\in\mathbb{R}^9$ denotes the logarithm of the right-invariant error
on~$\SE$.
Substituting this expression into the measurement model and performing a
first-order expansion of $h(\cdot)$ about $\hat X_k$ yields
\begin{equation}
\begin{aligned}
y_k
&= h\!\big(\exp(-\xi_k^\wedge)\hat X_k\big) + \nu_k\\[3pt]
&\approx h(\hat X_k) + H_k\,\xi_k + \nu_k,
\end{aligned}
\label{eq:meas_lin}
\end{equation}
where the Jacobian~$H_k$ is defined geometrically as
\begin{equation}
H_k
:= 
\left.\nabla_\xi h(\exp(-\xi^\wedge)\hat X_k)\right|_{\xi=0}\in\mathbb{R}^{m\times 9}.
\label{eq:H-def}
\end{equation}
Equation~\eqref{eq:H-def} provides a coordinate-free definition of the measurement sensitivity with respect to infinitesimal variations of the state on the group. 
In many cases, one may simply use the first-order approximation
$\exp(-\xi^\wedge)\approx I-\xi^\wedge$ inside the composite map
$h(\exp(-\xi^\wedge)\hat X_k)$ to obtain the corresponding Jacobian.

The linearized residual (innovation) is then
\begin{equation}
z_k 
:= y_k - h(\hat X_k)
\;\approx\;
H_k\,\xi_k + \nu_k,
\label{eq:innovation}
\end{equation}

which serves as the basis for the correction step. 
When $h(\cdot)$ preserves the same symmetry as the chosen invariant error—such as a right-invariant observation— it is possible to obtain a Jacobian~$H_k$ that is trajectory-independent after forming the corresponding invariant innovation, yielding a \emph{perfect} invariant update (see next subsection "\nameref{sec:right-invariant-observation}"). 
Otherwise, it depends on~$\hat X_k$, leading to an \emph{imperfect} but still geometrically consistent formulation.

Now, given the prior mean $(\hat X_k,\Sigma_k)$ and the linearized residual~\eqref{eq:innovation}, the standard Kalman covariance update applies:
\begin{align}\label{eq:gain1}
K_k &= \Sigma_k H_k^\top (H_k \Sigma_k H_k^\top + V_k)^{-1},\\[3pt]
\Sigma_k^+ &= (I_9 - K_k H_k)\,\Sigma_k.\label{eq:gain2}
\end{align}
The mean estimate is then retracted on the group through the multiplicative update
\begin{equation}
\hat X_k^+ = \exp\!\big(-(K_k z_k)^\wedge\big)\,\hat X_k,
\label{eq:retraction}
\end{equation}
which guarantees that the corrected state remains on the manifold~$\SE$ and that
the update is consistent with the underlying group geometry.
At the error level, this retraction induces the local update
\begin{equation}
\xi_k^+ \;\approx\; (I_9 - K_kH_k)\xi_k-K_k\nu_k,
\label{eq:error_update}
\end{equation}
thereby recovering the familiar additive correction structure in the Lie algebra.
Equations~\eqref{eq:innovation}–\eqref{eq:retraction} thus define a generic
invariant correction step, applicable to any aiding model that can be linearized
with respect to the chosen invariant error.
The next subsection specializes this general formulation to right-invariant
observation models, which encompass many body-frame sensing geometries commonly
encountered in aided inertial navigation.

Together, the invariant propagation equations \eqref{eq:hatX_update},\eqref{eq:Sigma-next-4th-A} derived in Part~I and the
correction equations~\eqref{eq:innovation}–\eqref{eq:retraction} define what is
commonly referred to as the right-invariant extended Kalman filter (RIEKF).
Algorithm~\ref{alg:rieKF-biasfree} summarizes the resulting prediction and
update steps, ensuring that the state estimate evolves on the manifold~$\SE$
and that both propagation and correction are consistent with the underlying
group geometry.

\subsection{Right-Invariant Measurement Models}\label{sec:right-invariant-observation}
Right-invariant observation models arise when the sensor output represents an inertial quantity expressed in the local (body) frame. 
Such measurements depend on the inverse of the state, since transforming a fixed global vector or point into body coordinates involves multiplication by~$X^{-1}$. 
This encompasses a broad class of aiding sensors: a magnetometer measuring the Earth's magnetic field in the body frame or visual sensors measuring the position of fixed landmarks relative to the vehicle.

The right-invariant measurement model can be written in homogeneous form as
\begin{equation}
\bar y_k = X_k^{-1} r + \bar\nu_k,
\qquad 
\bar\nu_k \sim \mathcal{N}(0,\bar V_k),
\label{eq:rightinv-meas}
\end{equation}
where both $\bar y_k$ and $r$ are expressed in homogeneous coordinates (i.e.,
$5{\times}1$ vectors in $\mathbb{R}^5$).
In practice, $\bar y_k$ does not necessarily coincide with the raw sensor output.
Accordingly, the physical measurement noise and its covariance are embedded into
the homogeneous representation.

Now, rather than linearizing~\eqref{eq:rightinv-meas} directly as in \eqref{eq:meas_lin}, we follow \cite{Barrau2017} and define the \textit{processed output}
\begin{equation}\label{eq:yk-right}
\begin{aligned}
y_k 
    &:= \Pi_5^3\,\hat X_k\,\bar y_k=\Pi_5^3\,\hat X_k\bigl(X_k^{-1}r + \bar\nu_k\bigr) \\
    &=: h(X_k) + \Pi_5^3\,\hat X_k\bar\nu_k
\end{aligned}
\end{equation}
where $ \Pi_5^3$, defined  in Table \ref{tab:constant-matrices} projects the homogeneous vector onto its 3D components.  
In this form, the predicted output is simply $h(\hat X_k)= \Pi_5^3 r$.  
Substituting $X_k = \exp(-\xi_k^\wedge)\hat X_k$ and using the first-order approximation $\exp(\xi_k^\wedge)\!\approx\!I+\xi_k^\wedge$ yields
\[
z_k := y_k - h(\hat X_k)
    \approx H^r\,\xi_k + \Pi_5^3\,\hat X_k\bar\nu_k,
\qquad 
H^r\xi := \Pi_5^3\,\xi^\wedge r.
\]
The Jacobian~$H^r$ is constant, and the estimated state appears only in the transformed noise term 
$\Pi_5^3\hat X_k\bar\nu_k$, whose covariance is $V_k^r = \Pi_5^3\,\hat X_k \bar V_k \hat X_k^\top(\Pi_5^3)^\top$.
Consequently, the linearized model is autonomous, and the same invariant correction step developed earlier can be directly applied with $H_k = H^r$ and $V_k = V_k^r$. 
Under the classical conditions of linear Kalman filter convergence~\cite{deyst1968conditions}, the right-invariant IEKF is locally asymptotically stable~\cite[Theorem~4]{Barrau2017}.

\subsection{Left-Invariant Measurement Models}
On the other hand, many aiding sensors provide measurements expressed directly in the inertial frame rather than in body coordinates (e.g., GPS for global position). These naturally follow a \emph{left-invariant} observation model:
\begin{equation}
    \bar y_k = X_k\,r + \bar\nu_k, 
    \qquad 
    \bar\nu_k \sim \mathcal{N}(0,\bar V_k),
    \label{eq:leftinv-meas}
\end{equation}
where both $\bar y_k$ and $r$ are expressed in homogeneous coordinates, with $r$ denoting a known body-frame reference (e.g., the lever arm from the body origin to the receiver).

One could alternatively define a processed measurement
\[
y_k := \Pi_5^3\,\hat X_k^{-1}\bar y_k,
\]
so that the resulting residual takes a form consistent with the previously
introduced right-invariant output structure.
However, under the right-invariant error definition adopted here, the
corresponding measurement Jacobian becomes dependent on the estimated
trajectory, thereby breaking the desirable property that the linearized
measurement model be independent of the state estimate.

In the original invariant filtering framework~\cite{Barrau2017}, left-invariant
measurement models are paired with left-invariant error definitions (i.e.,
matching handedness), which ensures that the resulting Jacobians remain
constant and preserves the structural advantages of the invariant formulation.
{{For a unified treatment of the different aiding modalities,
the present tutorial nevertheless adopts a common right-invariant error
definition throughout.}}

To recover the structural benefits while maintaining the right-invariant error,
we instead \emph{transform the fixed-frame output into an equivalent body-frame
(right-invariant) expression}.
{This reformulation preserves the underlying measurement
information while expressing it in a form compatible with the adopted
right-invariant error, thereby recovering a measurement linearization
that does not explicitly depend on the estimated state
\cite{Fornasier2025,Benahmed2025}.}

Specifically, from~\eqref{eq:leftinv-meas}, we can rewrite the processed output as
\begin{equation}\label{eq:left-to-right}
    y_k := \Pi_5^3\,\hat X_k r 
    \overset{\eqref{eq:leftinv-meas}}{=} 
    \Pi_5^3\,\hat X_k X_k^{-1}(\bar y_k - \bar\nu_k),
\end{equation}
whose estimated output is $h(\hat X_k) = \Pi_5^3\bar y_k$.  
This processed signal can be interpreted as a pseudo body-frame measurement enforcing a nonlinear constraint~\cite{julier2007kalman}.  
Linearizing around $\hat X_k$ yields
\[
z_k := y_k - h(\hat X_k)
    \approx H_k^\ell\,\xi_k - \Pi_5^3\bar\nu_k,
    \qquad 
    H_k^\ell\xi := \Pi_5^3\,\xi^\wedge \bar y_k,
\]
with effective noise covariance $V_k^\ell:= \Pi_5^3\,\bar V_k(\Pi_5^3)^\top.$
The resulting Jacobian $H_k^\ell$ depends on the current measurement $\bar y_k$ but not on the estimated state $\hat X_k$, which constitutes a major structural advantage.  
{This allows the fixed-frame measurement to be incorporated
within the common right-invariant correction framework adopted throughout
the tutorial without introducing explicit dependence of the measurement
Jacobian on the estimated state.}

\subsection{Mixed and Non-Invariant Measurement Models}

In practice, navigation systems often combine multiple sensor types. The unified correction framework developed here accommodates all of them consistently:
\begin{itemize}
    \item[(i)] \textbf{Right-invariant (body-frame) measurements} — such as magnetometers or 3D landmarks  — use the constant Jacobian $H^r$;
    \item[(ii)] \textbf{Left-invariant (global-frame) measurements} — such as GPS position or velocity — are rewritten in right-invariant form as in \eqref{eq:left-to-right}, leading to the measurement-dependent Jacobian $H_k^\ell$;
    \item[(iii)] \textbf{Non-invariant measurements} — such as bearing or range measurements — are treated using the generic Jacobian obtained from standard linearization at $\hat X_k$.
\end{itemize}
All cases are seamlessly integrated within a common correction and covariance
update mechanism, ensuring a consistent estimation framework across different
sensor modalities.
A catalogue of commonly used sensors, along with their corresponding processed
outputs and Jacobians, is provided in Table~\ref{tab:aiding-invariance}.

\begin{table*}[t]
\centering
\caption{\textbf{Jacobians for common aiding sensors.}
All Jacobians are expressed with respect to the right-invariant error.
For non-invariant measurement models, the generic Jacobian $H_k$ is used
directly together with the corresponding measurement noise covariance $V_k$.
For right- (resp.\ left-)invariant measurement models, the corresponding invariant Jacobian
$H_k^{r}$ (resp.\ $H_k^{\ell}$) is used, provided that the associated noise covariance
is consistently transformed (denoted $V_k^{\star}$ when paired with
$H_k^{\star}$).
In all cases, the noise covariance must correspond to the processed output.}
\renewcommand{\arraystretch}{1.15}
\setlength{\tabcolsep}{4pt}
\begin{threeparttable}
\rowcolors{2}{CSMgray}{white}
\begin{tabular}{p{3.0cm}p{3.2cm}p{5cm}p{4cm}}
\toprule
\rowcolor{CSMblue}
\textbf{Sensor} & \textbf{Measurement} &
\textbf{Processed Output $y_k$} & \textbf{Jacobian $H_k^\star$}\\
\midrule
GPS position with fixed lever arm $b$&$p_{gps}=p+Rb$ &\eqref{eq:left-to-right} with $r=(b,0,1)$&$H_k^\ell=[-p_{gps}^\wedge\;\;~0~\;\;I]$\\

GPS velocity with fixed lever arm $b$&$v_{gps}=v+R(\omega\times b)$ &
\eqref{eq:left-to-right} with $r=(\omega\times b,1,0)$&$H_k^\ell=[-v_{gps}^\wedge\;\;~I~\;\;0]$\\

Magnetometer&
$R^\top r_m$ &
\eqref{eq:yk-right} with $r=(r_m,0,0)$&
$H^r=[-r_m^\wedge\;\;~0~\;\;0]$\\

Landmark position&
$R^\top(p_\ell-p)$ &
\eqref{eq:yk-right} with $r=(p_\ell,0,1)$&
$H^r=[-p_\ell^\wedge\;\;~0~\;\;I]$\\

DVL &
$R^\top v$ &
\eqref{eq:yk-right} with $r=(0_{3\times 1},-1,0)$&
$H^r=[0\;\;~-I~\;\;0]$\\

Barometer (altitude) &
$e_3^\top p$ &Same
&$H_k=[e_3^\top\hat p^\wedge\;\;\;~0~-e_3^\top\,]$\\

Range to anchor &
$\|p_a-p\|$ &$\frac{1}{2}\|p_a-p\|^2$
&$H_k=(p_a-\hat p)^\top[-\hat p^\wedge\;\;\;~0~\;\;\;I\,]$ \\[2pt]

Pitot tube&
$e_1^\top R^\top(v-v_{\mathrm{wind}})$ &Same
&$H_k=e_1^\top[\hat R^\top v_{\mathrm{wind}}^\wedge\;\;~-\hat R^\top\;\;~0]$\\

Optical flow&
$f=\frac{R^\top v}{\|R^\top v\|}$ & $\pi_{f}R^\top v=0_{3\times 1}$ 
&$H_k=\pi_f[0\;\;~-\hat R^\top~\;\;0]$\\

Bearing to feature & $b=\frac{R^\top(p_\ell-p)}{\|R^\top(p_\ell-p)\|}$ &$\pi_bR^\top(p_\ell-p)=0_{3\times 1}$&
$H_k=\pi_b[-\hat R^\top p_\ell^\wedge\;\;~0~\;\;\hat R^\top]$\\

Tilt angle \cite{Alnahhal2025Scalar}&$e_3^\top R^\top e_3$&Same&$H_k=[-e_3^\top\hat R^\top e_3^\wedge\;\;~0~\;\;0]$\\

Pseudo–measurement (zero lateral velocity)&$e_2^\top R^\top v\approx 0$&Same&$H_k=e_2^\top[0\;\;~-\hat R^\top\;\;~0]$\\
\bottomrule
\end{tabular}
\end{threeparttable}
\label{tab:aiding-invariance}
\end{table*}

\begin{table*}[t]
\centering
\caption{\textbf{Constant matrices used throughout the paper.} For convenience, we collect here the constant matrices that appear throughout the paper.  
They are independent of the state and time, and are used to compactly express 
the group operations, adjoint mappings, and linearized dynamics on $\SE$.  
}
\label{tab:constant-matrices}
\renewcommand{\arraystretch}{1.25}
\setlength{\tabcolsep}{6pt}
\begin{tabular}{lll}
\toprule
\textbf{Symbol} & \textbf{Definition} & \textbf{Description}\\
\midrule
$D$ &
$\begin{bmatrix}
0_{3\times3} & 0_{3\times2}\\[2pt]
0_{2\times3} &S
\end{bmatrix}$ &
Velocity–position coupling in the homogeneous model.\\[10pt]
$S$ &
$\begin{bmatrix}
0 & 1\\[2pt]
0 & 0
\end{bmatrix}$ &
Velocity–position coupling in the Euclidean model.\\[10pt]

$\bar D$ &
$(e_3 e_2^\top)\!\otimes\! I_3$ &
Velocity–position coupling in the Lie–algebra coordinates.\\[6pt]

$\Pi_n^m$ &
$\big[\,I_m\;\;0_{m\times(n-m)}\,\big]$ &
Projection extracting the first $m$–components.\\
\bottomrule
\end{tabular}

\end{table*}

\subsection{Observability of the Linearized Error System}
\label{subsec:observability}
In the invariant filtering framework, the matrices $A_k$ and $H_k$ introduced earlier correspond to the 
\emph{linearized error dynamics} rather than a direct linearization of the system itself.  
They describe how a small perturbation $\xi_k$ of the invariant error evolves and influences the measurements. In particular, their rank properties determine which error directions can be corrected from sensor information.

The local observability of the error system is assessed through the classical condition
\begin{equation}
H_k A_{k-1}\!\cdots A_0\,\xi_0 \;=\; H_k A^k \xi_0 \;=\; 0, 
\qquad \forall k \ge 0,
\label{eq:invariant_observability_condition}
\end{equation}
where $\xi_0$ denotes an initial perturbation.  
If a nonzero $\xi_0$ satisfies~\eqref{eq:invariant_observability_condition}, 
it represents an \emph{unobservable perturbation}: the estimated state may drift in this direction without altering any predicted measurement.  
Note that the convergence (asymptotic stability) of the invariant Kalman filter relies on a slightly stronger property known as \emph{uniform observability}, as detailed in~\cite[Theorem~3]{Barrau2017}.  
The rank condition above remains necessary, while uniform observability provides its stronger, time-uniform counterpart that ensures exponential convergence guarantees.
To illustrate these concepts and build geometric intuition, we now examine several representative aiding configurations, each highlighting a distinct mechanism by which sensor information restores lost observability.

\begin{example}[3D Landmarks, spatial diversity]\label{example:landmarks}
    For the right-invariant error dynamics derived earlier, the state-transition matrix $A$ is constant over the interval of interest.  
For landmark measurements expressed in the body frame, the associated right-invariant measurement Jacobian $H^r$ is also constant (see Table \ref{tab:aiding-invariance}).  
This makes it possible to inspect~\eqref{eq:invariant_observability_condition} explicitly.
For a fixed landmark $p_\ell$, the corresponding row of the observability matrix reads
\[
H^r A^k =
\begin{bmatrix}
-p_\ell^\wedge + \tfrac12(k\Delta t)^2 g^\wedge
& k\Delta t\,I_3
& I_3
\end{bmatrix}.
\]
Let $\xi_0 = (\xi_\theta,\xi_v,\xi_p)$ denote the initial perturbation of attitude, velocity, and position, respectively.
Enforcing $H^rA^k\xi_0 = 0$ for all integers $k \ge 0$ gives
\[
\left(-p_\ell^\wedge
+\tfrac12(k\Delta t)^2 g^\wedge\right)\xi_\theta
+k\Delta t\,\xi_v+\xi_p=0,
\qquad \forall k.
\]
Interpreting this as a polynomial in $k$, each coefficient must vanish independently, which yields
\[
g\times\xi_\theta = 0, \qquad \xi_v = 0, \qquad \xi_p = p_\ell\times\xi_\theta.
\]
 Hence, the null space associated with a single landmark is
\[
\mathrm{span}\!\left\{
\begin{bmatrix}
g\\[2pt]0\\[2pt]p_\ell\times g
\end{bmatrix}
\right\}.
\]
This direction corresponds to a yaw rotation about the gravity vector combined with a consistent position shift—an error mode that cannot be detected from a single landmark measurement.  
When multiple landmarks are available, the overall null space is nontrivial if and only if 
$(p_i - p_j)\times g = 0$ for all $i,j$, 
that is, if all landmarks are vertically aligned with gravity.  
This degeneracy disappears as soon as at least three landmarks are non-aligned (\textit{spatial diversity}), 
in which case the system becomes fully observable—an assumption widely adopted in landmark-based navigation 
(see, \textit{e.g.},~\cite{Vasconcelos2010Nonlinear,Khosravian2015Observers,Wang2020Hybrid}).  
This simple analysis highlights how the geometry of the aiding configuration directly determines which invariant error components can be corrected.
\end{example}

\begin{example}[GPS position, temporal diversity]
For a GPS-type sensor providing the vehicle position in the inertial
frame, the invariant linearized Jacobian (w.r.t.\ the right-invariant
error) is
\begin{equation}
H_k^\ell =
\begin{bmatrix}
-p_k^\wedge \;\; 0 \;\; I_3
\end{bmatrix},
\label{eq:H_gps}
\end{equation}
where $p_k$ denotes the GPS position, considered noise-free for this
analysis (see Table \ref{tab:aiding-invariance}).
Applying the rank condition, an unobservable initial perturbation
must satisfy
\[
\left(-p_k^\wedge+\tfrac12(k\Delta t)^2g^\wedge\right)\xi_\theta
+k\Delta t\,\xi_v+\xi_p=0,
\qquad \forall k.
\]
Taking second forward differences gives
\[
\bar f_k^\wedge\xi_\theta=0,
\qquad
\bar f_k :=
\frac{p_{k+2}-2p_{k+1}+p_k}{\Delta t^2}-g.
\]
Here $\bar f_k$ is a discrete, averaged inertial specific-force
vector. Any nonzero unobservable attitude direction must therefore
be parallel to every $\bar f_k$.
If two such vectors are noncollinear, then $\xi_\theta=0$;
the original equations at two distinct sampling instants then
give $\xi_v=\xi_p=0$, establishing full observability over the
corresponding measurement window. For instance, for constant-velocity motion, $p_k=p_0+k\Delta t\,v_0$ gives
$\bar f_k=-g$ for all $k$, leaving yaw unobservable.
Thus, \textit{temporal diversity} is provided by noncollinear
discrete inertial specific-force vectors over the observation window.
A related continuous-time excitation condition appears
in \cite{vanGoor2025Automatica}.
\end{example}
\begin{example}[Compass, directional diversity]
A common way to remove the yaw ambiguity of GPS/IMU systems is to
include a compass (magnetometer) providing the body-frame measurement
of the known inertial magnetic field $r_m$.
In the right-invariant formulation, the corresponding Jacobian is
constant and given by (see Table \ref{tab:aiding-invariance})
\[
H^r =
\begin{bmatrix}
-r_m^\wedge & 0 & 0
\end{bmatrix}.
\]
Combining this with the GPS unobservability condition derived in
the preceding example gives
\[
\bar f_k^\wedge\xi_\theta=0,
\qquad
r_m^\wedge\xi_\theta=0,
\qquad \forall k.
\]
Thus, $\xi_\theta$ must be parallel to both $r_m$ and every
$\bar f_k$. If $r_m$ is noncollinear with at least one $\bar f_k$
in the observation window, then $\xi_\theta=0$; the original GPS
conditions then imply $\xi_v=\xi_p=0$, establishing full
observability. This \textit{directional diversity} is consistent
with the continuous-time noncollinearity assumption in
\cite{Berkane2021}.

As a particular case, at rest $\bar f_k=-g$, so the condition
reduces to noncollinearity of $g$ and $r_m$---a standard assumption
in attitude estimation \cite{Mahony2008,Zlotnik2016,Berkane2017}.
The system is therefore fully observable over a measurement window,
even when the vehicle is at rest.
\end{example}

\begin{figure*}[t]
\begin{tcolorbox}[colback=red!10, arc=2mm, boxrule=0pt,
                  left=3mm, right=3mm, top=3mm, bottom=3mm]

\centering
\includegraphics[width=0.9\textwidth]{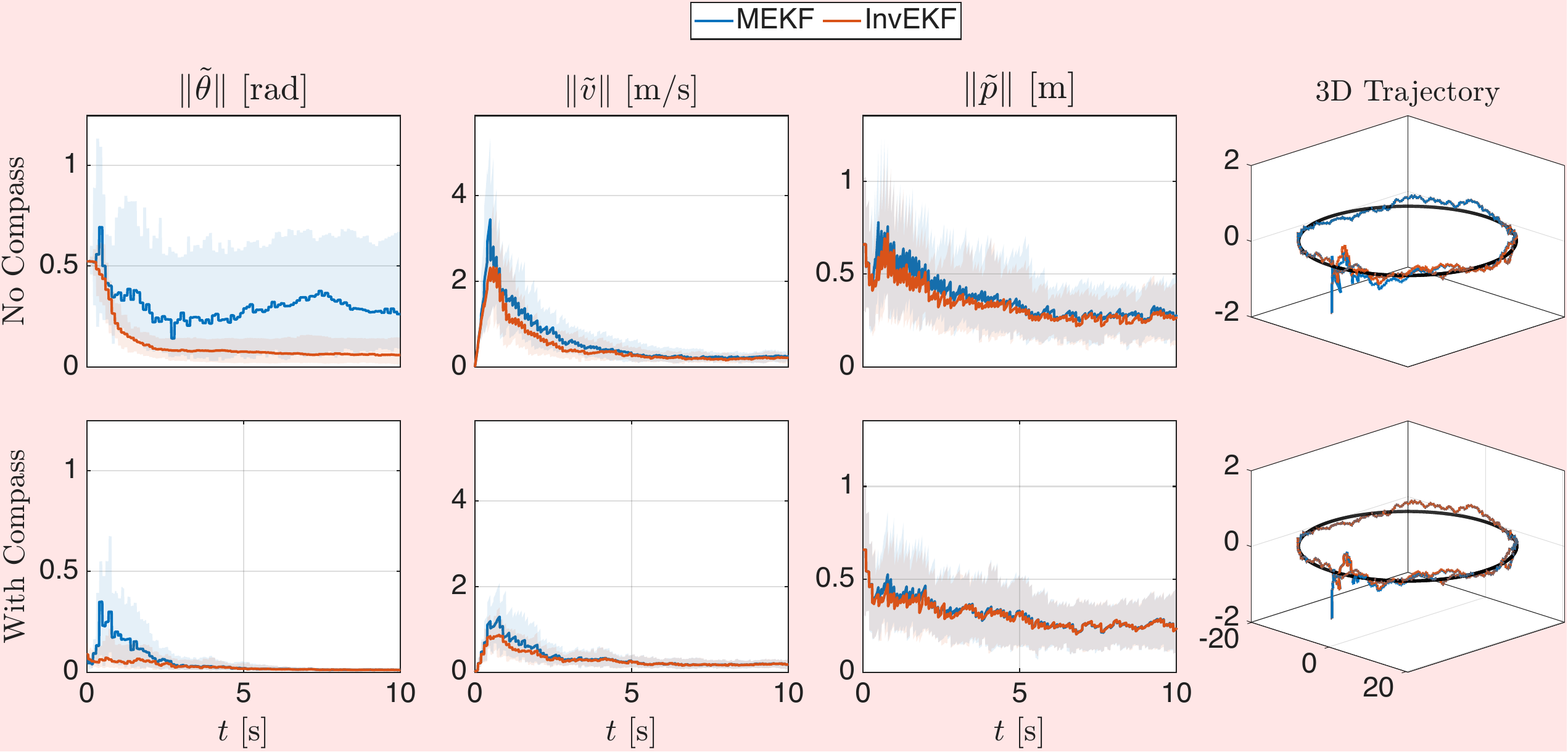}

\caption{\textbf{Comparison of the multiplicative EKF and the invariant EKF for GPS-aided inertial navigation, with and without magnetometer aiding.}
The filters are propagated using IMU measurements at $100\,\mathrm{Hz}$ and corrected whenever aiding measurements are available. Error panels show the first $10\,\mathrm{s}$, with median curves and shaded 10th--90th percentile bands over 50 paired noise realizations.
Trajectory panels show the full $60\,\mathrm{s}$ trajectory for one realization, with ground truth in black.
The \emph{top row} corresponds to IMU and GPS position updates only (no magnetometer),
whereas the \emph{bottom row} additionally incorporates body-frame magnetometer measurements providing heading information at $20\,\mathrm{Hz}$.
In the absence of magnetometer aiding, the invariant EKF exhibits smaller median attitude and velocity errors
during the initial transient, while the difference in position error is more modest.
Including magnetometer measurements improves the performance of both filters;
however, the invariant formulation retains an advantage during the initial transient,
particularly in attitude estimation.
}

\label{fig:example_circular}
\end{tcolorbox}
\end{figure*}

\begin{example}[Body-frame velocity, odometric aiding]
Consider now an aiding sensor that measures the vehicle’s linear velocity in its own body frame, 
as in Doppler velocity loggers (DVL) or wheel odometry.  
The corresponding measurement model is $y = R^\top v$, 
and the right-invariant Jacobian is therefore constant and given by
\[
H^r
=
\begin{bmatrix}
0 & -I_3 & 0
\end{bmatrix}.
\]
Substituting this into~\eqref{eq:invariant_observability_condition} yields
\[
H^r A^k =
\begin{bmatrix}
-k\Delta t\,g^\wedge & -I_3 & 0_{3\times3}
\end{bmatrix}.
\]
Enforcing $H^rA^k\xi_0 = 0$ for all integers $k \ge 0$ gives
\[
-k\Delta t\,g^\wedge\xi_\theta-\xi_v=0,
\qquad \forall k.
\]
which implies
\[
g\times\xi_\theta = 0,
\qquad
\xi_v = 0,\qquad\xi_p\in\mathbb{R}^3.
\]
This confirms that the system's global position is unobservable,
implying that such aiding provides only relative motion information
and therefore constitutes an {\it odometric measurement}.
Also, the first condition shows that attitude about the gravity
axis (yaw) is unobservable.
These four unobservable degrees of freedom (global position and yaw)
also occur in visual-inertial or LiDAR-inertial odometry without
absolute position or heading references; additional unobservable
directions may arise in degenerate configurations.
A deeper analysis of such relative-sensing frameworks is beyond
the scope of this tutorial.

\end{example}

\noindent\textbf{Summary.}
The above examples illustrate how different aiding modalities contribute complementary information to recover the unobservable modes of inertial navigation.  
Other sensors—such as altimeters, range finders, or cameras—follow the same principles, each introducing geometric constraints that progressively enhance the system’s observability (see Table~\ref{tab:aiding-invariance} for their Jacobians).  
Certain measurements can improve estimation performance even when
they do not expand the observable subspace; for example, when GPS
position aiding already ensures full observability, GPS velocity
can still improve numerical conditioning and estimation accuracy.

\begin{example}[GPS-aided inertial navigation]\label{ex:circular-ins}
For a numerical illustration, we consider a planar inertial navigation scenario designed to illustrate the practical impact of
trajectory-dependent versus invariant error formulations.
The vehicle follows a $60\,\mathrm{s}$ circular trajectory in the horizontal plane with radius
$r_0 = 20\,\mathrm{m}$ and constant tangential speed $v = 3\,\mathrm{m/s}$. 
The inertial measurement unit provides angular velocity and specific-force measurements at
$100\,\mathrm{Hz}$.
Both signals are corrupted by zero-mean continuous-time white noise processes with power spectral
densities
$Q_\omega = (0.2\,\mathrm{deg/s}/\sqrt{\mathrm{Hz}})^2 I_3$ and
$Q_a = (0.05\,\mathrm{m/s^2}/\sqrt{\mathrm{Hz}})^2 I_3$, respectively.
Global position measurements are obtained from a GPS receiver at $10\,\mathrm{Hz}$,
with isotropic Gaussian noise of standard deviation $0.5\,\mathrm{m}$ per axis.
In a second configuration, a magnetometer provides body-frame measurements of a fixed inertial
magnetic field direction at $20\,\mathrm{Hz}$, corrupted by zero-mean Gaussian noise with
standard deviation $0.05$.

Two estimators are implemented and compared under identical conditions:
a classical multiplicative extended Kalman filter (MEKF) and a right-invariant
extended Kalman filter (InvEKF).
{
The MEKF uses the multiplicative-attitude/additive-translation error
coordinates and the trajectory-dependent propagation matrix
$A_{\mathrm{MEKF}}(t)$ introduced in~\eqref{eq:mekf-error-propagation}.
Accounting for IMU measurement noise, its continuous-time error model becomes
\begin{equation}
\dot{\tilde x}
\approx
A_{\mathrm{MEKF}}(t)\tilde x
+
\begin{bmatrix}
\hat R & 0_3\\
0_3 & -\hat R\\
0_3 & 0_3
\end{bmatrix}n,
\label{eq:mekf-error-stochastic}
\end{equation}
so that both the deterministic error propagation and the noise injection
depend explicitly on the estimated attitude.}
In contrast, the InvEKF exploits the Lie-group structure of the navigation model
to yield autonomous noise-free error dynamics, see \eqref{eq:xi_deterministic}.
The stochastic noise injection nevertheless depends on the estimated state.
Both filters are initialized with the same $30^\circ$ pitch error, while position and velocity are initialized
at their true values. The initial covariance is specified as $P_{0,\mathrm{MEKF}} = I_9$
in the MEKF error coordinates and mapped to the invariant coordinates according to
\[
\begin{aligned}
P_{0,\mathrm{InvEKF}} &= T_0 P_{0,\mathrm{MEKF}} T_0^\top,\quad
T_0 &=
\begin{bmatrix}
I_3 & 0_3 & 0_3\\
\hat v_0^\wedge & -I_3 & 0_3\\
\hat p_0^\wedge & 0_3 & -I_3
\end{bmatrix}.
\end{aligned}
\]
Here $\hat v_0$ and $\hat p_0$ are the common initial velocity and position estimates.
The matrix $T_0$ is the Jacobian of the change from MEKF to logarithmic invariant
error coordinates, evaluated at zero error about the common initial estimate.
This transformation matches the initial physical covariance to first order.
The comparison uses 50 paired noise realizations.

Figure~\ref{fig:example_circular} reports the evolution of the attitude, velocity, and position
error norms, together with the estimated trajectories, for two sensing configurations.
In the absence of magnetometer aiding (top row), attitude information is indirect and relies
on the coupling between inertial measurements and translational motion.
In this regime, the MEKF exhibits larger transient attitude and velocity errors,
while the difference in position error is more modest.
{
This behavior is consistent with the sensitivity of a trajectory-dependent
linearization when the estimate departs significantly from the true
trajectory, in which case the local approximation underlying
$A_{\mathrm{MEKF}}(t)$ may become inaccurate.}

By contrast, the invariant EKF exhibits
{smaller median transient errors} under the same
limited sensing configuration.
{
Its deterministic propagation benefits from the autonomous log-linear invariant error
dynamics illustrated previously, while corrections are performed directly
on the Lie group through the multiplicative retraction
$\hat X^+ = \exp(-(Kz)^\wedge)\hat X$.}
When magnetometer measurements are included (bottom row), additional attitude information is available and
the performance of both filters improves significantly.
Nevertheless, the invariant EKF continues to exhibit smaller errors during the initial transient,
particularly in attitude, while the long-term performance of the two filters becomes similar.
\end{example}

\section{Recent and Ongoing Developments in the Field}
{Recent years have witnessed substantial progress in the
theoretical foundations and practical deployment of invariant filtering
techniques for navigation and robotics
\cite{Barrau2018,Brossard2020,hartley2020contact}.
A long-standing question concerns the distinction between left- and
right-invariant formulations of the InvEKF and the classical recommendation
to match the invariant error definition to the symmetry of the measurements
\cite{Potokar2024}. Such matching exposes particularly simple invariant
output errors and, for invariant measurements, can yield measurement
Jacobians that are independent of the estimated state.}

{Recent work by Ge et al.~\cite{Ge2026} provides a complementary perspective.
When the appropriate reset operation is included, the left- and
right-InvEKF are shown to be stochastically equivalent, with their state
estimates and covariance representations related through the corresponding
change of invariant coordinates. This result shows that measurement--error
matching is not required for equivalence of the complete filters with reset.
It should not, however, be interpreted as implying that the choice of
invariant error, or its compatibility with the measurement symmetry, is
structurally irrelevant. In particular, while some properties of the
classical InvEKF carry over to the reset formulation, the consistency
properties associated with symmetry-induced unobservable directions are
not established in general and may be affected by the reset
\cite{Ge2026}.}

In the following, we briefly review several recent theoretical developments that aim to provide stronger stability guarantees than those offered by the classical InvEKF. Our aim is not to catalogue observer designs tailored to specific sensing modalities, but rather to focus on frameworks that accommodate a broad class of measurement models within a unified, symmetry-based formulation, in a spirit similar to that of the InvEKF. Such approaches emphasize structural properties over sensor-specific constructions, and thus provide a flexible foundation for multi-sensor fusion and extended state representations.
\subsection{Observers on the Extended Pose--Velocity Group $\SEE$}
\label{subsec:se53}
\begin{table*}[t]
\centering
\caption{\textbf{Measurements compatible with the $\SEE$ linear error model.}
All outputs admit an innovation that has the exact linear form $z_i =y_i-\hat y_i= C_i\,\xi$ with
$C_i = -(r_i^\top \otimes H_i)$. Note the difference between the actual measurement and the processed output $y_i$ used in the filter.}
\renewcommand{\arraystretch}{1.25}
\setlength{\tabcolsep}{5pt}
\begin{threeparttable}
\rowcolors{2}{CSMgray}{white}
\begin{tabular}{p{2.8cm}p{3.0cm}p{3cm}p{3cm}p{3cm}}
\toprule
\rowcolor{CSMblue}
\textbf{Sensor} &
\textbf{Measurement} &
\textbf{Output $y_i$} &
\textbf{Matrix $H_i$} &
\textbf{Vector $r_i^\top$} \\
\midrule

GPS position (lever arm $b$) &
$p_{\mathrm{gps}}=p+Rb$ &
$b=R^\top(p_{\mathrm{gps}}-p)$ &
$I_3$ &
$\left[\begin{array}{ccc}
0 & -1 & p_{\mathrm{gps}}^\top
\end{array}\right]$ \\
Landmarks &
$y=R^\top(p_\ell-p)$ &
$y=R^\top(p_\ell-p)$&
$I_3$ &
$\left[\begin{array}{ccc}
0 & -1 & p_\ell^\top
\end{array}\right]$\\

Bearing to landmark &
$b_\ell=\frac{R^\top(p_\ell-p)}{\|p_\ell-p\|}$&
$0=\pi_{b_\ell}R^\top(p_\ell-p)$ &
$\pi_{b_\ell}=I_3-b_\ell b_\ell^\top$ &
$\left[\begin{array}{ccc}
0 & -1 & p_\ell^\top
\end{array}\right]$\\

Tilt-like constraint &
$y=b^\top R^\top a$ &
$y=b^\top R^\top a$ &
$b^\top$ &
$\left[\begin{array}{ccc}0 & 0 & a^\top\end{array}\right]$
\\

Magnetometer &
$y=R^\top a$ &
$y=R^\top a$ &
$I_3$ &
$\left[\begin{array}{ccc}0 & 0 & a^\top\end{array}\right]$
 \\

Pitot tube &
$y=e_1^\top R^\top(v-v_w)$ &
$y=e_1^\top R^\top(v-v_w)$ &
$e_1^\top$ &
$\left[\begin{array}{ccc}1 & 0 & -v_w^\top\end{array}\right]$
\\

DVL &
$y=R^\top v$ &
$y=R^\top v$ &
$I_3$ &
$\left[\begin{array}{ccc}1 & 0 & 0\end{array}\right]$\\

GPS velocity &
$v_{\mathrm{gps}}=v+R(\omega\times b)$ &
$\omega\times b=R^\top(v_{\mathrm{gps}}-v)$ &
$I_3$ &
$\left[\begin{array}{ccc}-1 & 0 &v_{\mathrm{gps}}^\top\end{array}\right]$ \\

Optical flow &
$f=R^\top\frac{v}{\|v\|}$ &
$0=\pi_{f}R^\top v$ &
$\pi_{f}=I_3-f\,f^\top$ &
$\left[\begin{array}{ccc}1 & 0 &0\end{array}\right]$ \\

Pseudo–measurement (zero side-slip angle)&$e_2^\top R^\top (v-v_w)\approx 0$&$0=e_2^\top R^\top (v-v_w)$&$e_2^\top$ &
$\left[\begin{array}{ccc}1 & 0 & -v_w^\top\end{array}\right]$ \\
\bottomrule
\end{tabular}
\end{threeparttable}
\label{tab:se53_measurements}
\end{table*}
As discussed throughout this tutorial, invariant observer designs on the pose--velocity
group $\SE$ typically guarantee only local asymptotic stability of the estimation error.
While the invariant extended Kalman filter (InvEKF) has demonstrated strong performance
and robustness to large initial errors in many practical scenarios, several works have
sought observer designs achieving almost global or global asymptotic stability.
Due to the well-known topological obstruction to global asymptotic stability on compact
manifolds such as $\SO$ \cite{BhatBernstein2000}, continuous time-invariant observer designs on $\SO$ and its
product manifolds (e.g., $\SE$) can achieve at most almost global asymptotic stability,
with convergence guaranteed from all initial conditions except for a set of Lebesgue
measure zero.

One route toward almost global asymptotic stability consists in enlarging the group
structure so as to embed additional geometric information directly into the state space,
at the price of stricter observability requirements.

\begin{pullquote}
One route toward almost global asymptotic stability consists in enlarging the group
structure so as to embed additional geometric information directly into the state space,
at the price of stricter observability requirements.
\end{pullquote}

To this end, we consider the extended Lie group $\SEE$, whose elements are written as
\begin{equation}\nonumber
X = \Gamma(R,Z) =
\begin{bmatrix}
R & Z\\
0_{5\times 3} & I_5
\end{bmatrix},
\qquad
Z = \begin{bmatrix} v & p & Q \end{bmatrix} \in \mathbb{R}^{3\times 5},
\label{eq:se53_state}
\end{equation}
where $R\in\SO$, $v,p\in\mathbb{R}^3$, and
$Q\in\mathbb{R}^{3\times3}$.
The physical navigation state is embedded by fixing
$Q=[e_1\ e_2\ e_3]=I_3$.
This auxiliary state does not correspond to an additional physical quantity.
Rather, it serves a purely geometric role by lifting the system to a
higher-dimensional Lie group, thereby modifying the symmetry structure of the
estimation error dynamics and enabling observer designs with improved global
convergence properties.

The inertial navigation dynamics on $\SEE$ retain a group-affine structure and are written as
\begin{equation}
\dot X = XU + [X,D],
\label{eq:se53_dynamics}
\end{equation}
where
$
U= (\omega,\,a,\,0,\,0,\,0,\,0)^\wedge\in\mathfrak{se}_5(3)
$. Here, the $\wedge$ operator is overloaded in the natural way to map vectors in
$\mathbb{R}^{18}$ to elements of $\mathfrak{se}_5(3)$, following the same convention used
throughout the tutorial for $\SE$. The constant coupling matrix $D$ has the form
\begin{equation}
D =
\begin{bmatrix}
0_{3\times 3} & 0_{3\times 5}\\
0_{5\times 3} & A
\end{bmatrix},
\qquad
A=
\begin{bmatrix}
0 & 1 & 0_{1\times 3}\\
0 & 0 & 0_{1\times 3}\\
g & 0_{3\times 1} & 0_{3\times 3}
\end{bmatrix}.
\end{equation}

The matrix $A$ captures the kinematic coupling induced by gravity and the velocity-position integrator.
Interestingly, the state augmentation has allowed to remove the extra gravity term $GX$ that was present in \eqref{eq:ins_SE}, which is essential for the observer construction hereafter.
Consider now the observer
\begin{equation}
\dot{\hat X} = \hat XU + [\hat X,D] + \Delta\,\hat X,
\qquad
\Delta=(\Delta_R^\vee,\Delta_Z)^\wedge\in \mathfrak{se}_5(3),
\label{eq:se53_observer}
\end{equation}
and define the right-invariant estimation error
\begin{equation*}
\tilde X = \hat X X^{-1}
=
\begin{bmatrix}
\tilde R & \tilde Z\\
0 & I_5
\end{bmatrix},
\qquad
\tilde R = \hat R R^\top,
\quad
\tilde Z = \hat Z - \tilde R Z .
\end{equation*}
A direct computation shows that the error evolves autonomously according to
\begin{equation}
\dot{\tilde X} = [\tilde X,D] + \Delta\,\tilde X,
\label{eq:se53_error}
\end{equation}
which yields the \textit{decoupled} subsystems
\begin{align}
\dot{\tilde R} &= \Delta_R \tilde R, \label{eq:se53_rot_error}\\
\dot{\tilde Z} &= \Delta_R \tilde Z + \tilde Z A + \Delta_Z. \label{eq:se53_trans_error}
\end{align}
Consider the right-invariant translational error $\tilde Z$ governed by
\eqref{eq:se53_trans_error}. Expressing this error in the body frame and
vectorizing, define the translational error state
\begin{equation}
\xi := \mathrm{vec}\!\left(\hat R^\top \tilde Z\right) \in \mathbb{R}^{15}.
\label{eq:def_xi}
\end{equation}
Using the identity $\mathrm{vec}(AXB)=(B^\top\!\otimes A)\mathrm{vec}(X)$ together
with $\dot{\hat R}=\hat R\omega^\wedge+\Delta_R\hat R$ and
\eqref{eq:se53_trans_error}, all terms involving $\Delta_R$ cancel exactly,
yielding the closed-form linear time-varying dynamics
\begin{equation}
\dot\xi(t)
=
F(t)\,\xi(t) + u_\Delta(t),
\label{eq:LTV_xi}
\end{equation}
where
\begin{equation}
F(t)
:=
A^\top\!\otimes I_3 - I_5\!\otimes \omega(t)^\wedge,
\qquad
u_\Delta(t)
:=
\mathrm{vec}\!\left(\hat R^\top\Delta_Z\right).
\label{eq:F_udelta}
\end{equation}
System \eqref{eq:LTV_xi} is obtained without any linearization and is independent
of the state estimate, depending only on measured angular velocity through
$\omega(t)^\wedge$.

A key feature of the $\SEE$ construction is that a broad class of inertial and
exteroceptive measurements leads to innovations that are linear in $\xi$.
Specifically, consider outputs of the form
\begin{equation}
y_i = H_i R^\top Z r_i,
\label{eq:generic_output}
\end{equation}
where $H_i\in\mathbb{R}^{m_i\times 3}$ and $r_i\in\mathbb{R}^5$ are known matrices and vectors,
respectively. Many sensors commonly used for inertial navigation aiding can be expressed
in this form, either directly or after simple algebraic pre-processing that yields an
equivalent output structure; representative examples are summarized in
Table~\ref{tab:se53_measurements}. The associated measurement noise covariance must, in
general, be mapped consistently from the original sensor noise for each particular case.
For clarity of exposition, this aspect is omitted here.

Thanks to this output structure, the corresponding innovation can be written as
\begin{align}
z_i
&:= y_i - \hat y_i \nonumber= H_i R^\top Z r_i - H_i \hat R^\top \hat Z r_i \nonumber\\
&= -\,H_i \hat R^\top \tilde Z r_i \nonumber= C_i\,\xi,
\label{eq:linear_output}
\end{align}
where we defined
$C_i := -\,r_i^\top \otimes H_i$. Stacking all available measurements yields the LTV output equation
\begin{equation}
z(t) = C(t)\,\xi(t),
\label{eq:LTV_output}
\end{equation}
where $C(t)$ is formed by stacking the individual matrices $C_i$.
The translational error dynamics \eqref{eq:LTV_xi}–\eqref{eq:LTV_output} thus form a
standard linear time-varying system.

Next, a Riccati-based correction law is chosen as
\begin{equation}
u_\Delta(t) = -K(t)\,z(t),
\qquad
K(t)=P(t)C^\top(t)Q(t),
\label{eq:riccati_gain}
\end{equation}
where $P(t)$ solves the continuous Riccati equation
\begin{equation}
\dot P = FP + PF^\top - PC^\top QCP + V(t),
\label{eq:riccati}
\end{equation}
with $P(0)=P_0$ positive definite. The symmetric weighting
matrices $Q(t)$ and $V(t)$ are continuous, uniformly bounded,
and uniformly positive definite.

\begin{theorem}[UGES of the translational subsystem]
\label{thm:uges_translation}
Suppose that $F(t)$ and $C(t)$ defined by \eqref{eq:F_udelta}
and \eqref{eq:LTV_output} are bounded and continuous, and that
the pair $(F(t),C(t))$ is uniformly observable.
Then,
the origin $\xi=0$ of the noise-free closed-loop system
\begin{equation}
\dot\xi = \big(F(t)-K(t)C(t)\big)\xi
\label{eq:error_dynamics_xi}
\end{equation}
is uniformly globally exponentially stable.
\end{theorem}
Theorem~\ref{thm:uges_translation} establishes uniform global exponential convergence
of the translational error state $\xi=\mathrm{vec}(\hat R^\top \tilde Z)$.
Uniform global exponential stability of $\xi$ implies
\[
\hat R^\top \hat Z \;\longrightarrow\; R^\top Z .
\]
As a result, the estimates $\hat R^\top \hat v$ and $\hat R^\top \hat p$
converge exponentially to the true body-frame velocity $R^\top v$ and position
$R^\top p$, respectively.
The remaining degree of freedom is governed by the attitude error subsystem
\eqref{eq:se53_rot_error}, where the correction $\Delta_R\in\so$ is constructed
using the fixed embedded frame $Q=[e_1\ e_2\ e_3]=I_3$ and its estimate
$\hat Q=[\hat e_1\ \hat e_2\ \hat e_3]$ according to
\begin{equation}
\Delta_R
:=
\sum_{i=1}^3 \rho_i\,(\hat e_i \times e_i)^\wedge,
\qquad \rho_i>0.
\label{eq:deltaR}
\end{equation}

The choice of \eqref{eq:deltaR} is motivated by its intrinsic geometric structure.
Indeed, using the identity
\[
\hat R^\top(\hat e_i \times e_i)
=
(\hat R^\top \hat e_i)\times(\hat R^\top e_i),
\]
and noting that $\hat R^\top \hat e_i$ converges exponentially to $R^\top e_i$ as a
direct consequence of the translational error convergence, it follows that the
innovation $\Delta_R$ is asymptotically driven by the cross product between the
estimated and true body-frame representations of the inertial axes.
This structure coincides with that of the nonlinear complementary filter on
$\SO$~\cite{Mahony2008}, for which almost global asymptotic convergence is well
established.

Accordingly, the proposed attitude correction inherits the same geometric
properties, with the translational error $\xi$ acting as a vanishing perturbation.
To make the cascade structure explicit, the attitude error dynamics can be written
in the perturbed form~\cite{Wang2021}
\begin{equation}
\dot{\tilde R}
=
-2\mathbb{P}_a(\tilde R\Lambda)\,\tilde R
+
\big[\Gamma(t)\,\xi\big]^\wedge\,\tilde R,
\label{eq:att_perturbed}
\end{equation}
where $\Lambda=\mathrm{diag}(\rho_1,\rho_2,\rho_3)$,
$\mathbb{P}_a(M):=\tfrac12(M-M^\top)$, and $\Gamma(t)$ is a continuous bounded matrix
depending on the attitude estimate $\hat R$.
The autonomous part of \eqref{eq:att_perturbed} is almost globally asymptotically
stable, while the perturbation term vanishes exponentially as $\xi\to 0$.
Consequently, the overall estimation error dynamics inherit an almost global
asymptotic stability property through the cascade with
\eqref{eq:error_dynamics_xi}.

\begin{theorem}[Almost global convergence on $\SEE$]
\label{thm:agas_se53}
Assume the hypotheses of Theorem~\ref{thm:uges_translation} so that $\xi$ is UGES.
Pick positive and distinct gains $\rho_i>0$.
Then the equilibrium $\tilde X=I_8$ of the full error dynamics \eqref{eq:se53_error}
is almost globally asymptotically stable.
\end{theorem}
While the proposed $\SEE$ observer yields almost global asymptotic stability
under the stated observability conditions, an important limitation concerns the
treatment of sensor biases.
In particular, extending the present framework to simultaneously estimate IMU
biases while preserving almost global convergence remains a challenging open
problem.
Addressing bias estimation within an AGAS framework therefore constitutes an
interesting direction for future research.
\begin{remark}
An alternative viewpoint, closely related to the present state-extension approach on $\SEE$,
is to note that the remaining columns of $\hat R^\top \hat Z$, \emph{i.e.},
$\hat R^\top \hat Q$, converge exponentially to the rotation matrix
$R^\top Q = R^\top$ (recall that $Q\equiv I_3$).
At this stage, the attitude can be recovered from the estimate $\hat R^\top \hat Q$,
which is not necessarily an element of $\SO$.
A valid estimate of $R$ may then be obtained by transposing the projection of
$\hat R^\top\hat Q$ onto $\SO$, for instance via the nearest-rotation mapping
based on singular value decomposition (SVD);
see~\cite{Benahmed2024LTV,Alnahhal2025Scalar}.
Since this convergence property is independent of the attitude innovation $\Delta_R$,
one may alternatively implement the observer directly in the body frame by defining
$\hat Z^{\mathcal B} := \hat R^\top \hat Z$.
In this case, the dynamics of $\hat Z^{\mathcal B}$ can be shown to be independent of
$\hat R$ and $\Delta_R$, resulting in a reduced-order implementation (without $\hat R$ dynamics) with lower
memory requirements. A potential limitation of this approach is that the projection onto $\SO$ is not, in general, a smooth operation, and may therefore introduce discontinuities in the reconstructed attitude estimate.\end{remark}

\subsection{Synchronous Observer Design}

An alternative approach to almost global asymptotic stability (AGAS) is based on
\emph{synchronous observer design}, originally introduced via the notion of
$E$-synchrony \cite{LagemanTrumpfMahony2009,vanGoorMahony2021} and recently applied to inertial navigation systems with AGAS
guarantees \cite{vanGoor2025Automatica}. The central idea is to augment the observer with an
auxiliary state evolving on an automorphism group of the navigation state space,
such that a suitably defined error is constant in the absence of innovation
terms.

For inertial navigation, the relevant automorphism group is the \textit{extended
similarity group} $\mathbb{SIM}_2(3)$ acting on $\SE$. However, as shown in
\cite[Remark~6.3]{vanGoor2025Automatica}, it is sufficient to track only the translational
$3\times 2$ component of the auxiliary $\mathbb{SIM}_2(3)$ state while preserving the almost global
convergence properties. This reduced construction significantly simplifies the
observer design. We adopt this
reduced synchronous formulation in what follows.

We consider the extended Lie group $\SE$, whose elements are written as
\begin{equation}
X = \Gamma(R,Z) =
\begin{bmatrix}
R & Z\\
0_{2\times 3} & I_2
\end{bmatrix},
\qquad
Z = \begin{bmatrix} v & p \end{bmatrix} \in \mathbb{R}^{3\times 2}.
\end{equation}
With this representation, the translational component $Z$ evolves according to
\begin{equation}
\dot Z = ZS + (g + Ra)B,
\end{equation}
where $B=[1\;\;0]\in\mathbb{R}^{1\times 2}$ and the constant matrix $S$ is given in
Table~\ref{tab:constant-matrices}. Let $\hat X=\Gamma(\hat R,\hat Z)\in\SE$ denote the estimated state. Following the
standard synchronous construction \cite{vanGoor2025Automatica}, the pre-observer dynamics are chosen as
\begin{equation}
\dot{\hat X}
= \hat X U + G\hat X + [\hat X,D] + \Delta \hat X,
\qquad \hat X(0)\in\SE,
\end{equation}
with $U$, $G$, and $D$ defined in~\eqref{eq:ins_SE} and $\Delta=(\Delta_R,\Delta_Z)^\wedge\in\se$. At the level of the translational
block, this yields
\begin{equation}
\dot{\hat Z}
= \hat ZS + \Delta_R^\wedge \hat Z + (g+\hat R a)B + \Delta_Z .
\end{equation}
If one were to use the standard right-invariant error
$\hat X X^{-1}=\Gamma(\tilde R,\hat Z-\tilde RZ)$ with the associated translational
error $\tilde Z := \hat Z - \tilde R Z,$
it can be observed that the evolution of $\tilde Z$ is intrinsically coupled to
the rotational error $\tilde R$ due to the gravity term. Indeed, one obtains
\[
\dot{\tilde Z}
= \tilde Z S + (I-\tilde R)gB
+ \Delta_R^\wedge \tilde Z + \Delta_Z ,
\]
which prevents a direct decoupled analysis of the translational dynamics.
This coupling can be compensated by introducing an auxiliary state
$\Psi\in\mathbb{R}^{3\times 2}$ governed by
\begin{equation}
\dot\Psi = \Psi S + gB + \Gamma .
\end{equation}
Using this auxiliary variable, we define the \textit{coupled} error signal
\begin{equation}
\xi := \tilde R^\top \tilde Z + (I-\tilde R^\top)\Psi .
\end{equation}
The use of coupled error signals to remove coupling between translational and rotational dynamics have been used in different context related to the inertial navigation problem such as in \cite{RobertsTayebi2011,BerkaneTayebi2017,Berkane2021}.
A straightforward computation then gives the dynamics
\begin{equation}
\dot\xi
= \xi S
+ \tilde R^\top \Delta_R^\wedge \Psi
+ (I-\tilde R^\top)\Gamma
+ \tilde R^\top \Delta_Z .
\end{equation}
We now assume inertial position measurements of the form
$
y = ZC, \;C=(0\;1)^\top .
$
The innovation terms are selected as
\begin{align}
\Delta_Z &= -\Delta_R^\wedge\Psi + (y-\hat ZC)L,\\
\Gamma &= (y-\Psi C)L,
\end{align}
where $L=[l_v\;l_p]\in\mathbb{R}^{1\times 2}$ is a constant gain. With this choice, all coupling
terms cancel and the error dynamics reduce to
\begin{equation}
\dot\xi = \xi(S-CL).
\end{equation}
If the entries of $L$ are positive, the matrix $(S-CL)$ is Hurwitz
\cite{vanGoorHamelMahony2023}, and therefore $\xi$ converges globally exponentially
to zero, independently of the rotational error dynamics. Furthermore,
the rotational correction is chosen as
\begin{equation}
\Delta_R
= \rho\,(\hat ZC-\Psi C)\times (y-\Psi C).
\end{equation}
Assume that the inertial-frame specific acceleration \(Ra\) is bounded, uniformly continuous, and directionally persistently exciting. If \(l_p>0\), \(0<l_v<l_p^2/4\), and \(\rho>0\), then the equilibrium
\[
(\tilde R,\xi)=(I_3,0_{3\times2})
\]
is almost globally asymptotically stable~\cite[Theorem~4.1]{vanGoorHamelMahony2023}.

Compared with the full $\SEE$ construction (see ``\nameref{subsec:se53}''), this reduced formulation
requires fewer auxiliary states ($3\times 2$ instead of $3\times 3$), admits constant
gains, and achieves convergence under the excitation condition stated above. At the same
time, its extension to other sensing modalities, such as bearing-only measurements
or pitot-tube sensing, remains limited. Incorporating IMU bias estimation into this synchronous construction while
preserving almost global asymptotic stability also remains an open problem.

\subsection{Alternative Constructions and Bias Handling}
Several alternative observer constructions have been proposed in the literature
for inertial navigation systems equipped with inertial-frame position measurements.
These designs typically achieve what may be described as almost semi-global
exponential stability, while relying on simpler internal structures than the
ones discussed above.
In particular, they require a reduced number of auxiliary states
\cite{Bryne2017,Berkane2021}, and in some cases no auxiliary states at all
\cite{BerkaneTayebi2021}. A notable advantage of these approaches is their explicit treatment of gyro biases,
which are incorporated directly into the observer dynamics.
However, these observers are largely tailored to inertial-frame position measurements,
and extending them to more general sensing modalities remains an open and nontrivial problem.

{
A complementary line of work addresses biased inertial navigation through
equivariant filtering. Recent developments have shown that several modern
geometric INS filters can be interpreted within a common EqF framework,
with different choices of symmetry leading to different error coordinates
and linearization properties \cite{Fornasier2025}. This perspective is
particularly useful in the presence of IMU biases, for which the
group-affine structure of the bias-free INS is generally lost under
conventional state augmentation.}

{Among the resulting constructions, the tangent-group EqF
(TG-EqF) employs the semi-direct product
$\SE\ltimes\se$ to geometrically couple the navigation and bias states.
With this symmetry, the navigation-state error dynamics admit an exact
linearization, with the remaining nonlinearities confined to the bias
subsystem \cite{Fornasier2025}. Other geometric treatments also exploit
coupling between navigation and bias states. In particular, the
two-frames group construction of Barrau and Bonnabel
\cite{Barrau2023} incorporates the IMU biases within the symmetry
structure, improving the resulting error linearization compared with
a direct-product bias augmentation, although state dependence remains
in the velocity, position, and bias error dynamics. More recent
semi-direct constructions, such as the DP-EqF and SD-EqF
\cite{Fornasier2025}, retain part of the tangent-group bias coupling
while avoiding the additional virtual bias state required by the
TG-EqF. These different choices of symmetry therefore provide different
trade-offs between state dimension and the structure of the resulting
error dynamics.}

{Beyond these propagation-side considerations, an important
practical advantage of the EqF framework arises from exploiting symmetry
in the measurement model. As discussed in ''\nameref{sidebar-InEKF}", the
EqF$^\star$ construction improves the standard first-order output
approximation by eliminating its second-order error term, yielding an
approximation with an $\mathcal O(\|\xi\|^3)$ remainder \cite{vanGoorHamelMahony2022}. This higher-order
measurement approximation can significantly improve transient filtering
performance and provides a complementary benefit to the improved error
dynamics obtained through suitable choices of symmetry.}

\section{Conclusion}

This tutorial presented a unified, control-oriented treatment of aided inertial
navigation systems grounded in Lie-group theory.
Rather than surveying estimation methods in isolation, the exposition was
deliberately organized around a single geometric backbone—the pose–velocity
group $\SE$ and its extensions—through which modeling, propagation, uncertainty
analysis, and measurement fusion were developed in a coherent and structurally
transparent manner.

A central message of this work is that many results scattered across the inertial
navigation, robotics, and estimation literature admit a common interpretation
once the intrinsic symmetries of the problem are made explicit.
By formulating both the system dynamics and the estimation error directly on Lie
groups, the tutorial clarifies why invariant error definitions lead to autonomous
(or state-independent) linearized dynamics, constant transition matrices, and
simplified observability analysis.
This perspective explains, in a unified way, the robustness and consistency
properties observed in invariant extended Kalman filters and related nonlinear
observers.

\begin{pullquote}
A central message of this work is that many results scattered across the inertial
navigation, robotics, and estimation literature admit a common interpretation
once the intrinsic symmetries of the problem are made explicit.
\end{pullquote}

Beyond classical propagation and filtering on $\SE$, the tutorial showed how the
same geometric principles naturally extend to higher-order constructions such as
$\SEE$, synchronous observer designs, and equivariant filtering frameworks.
These extensions were presented not as separate methodologies, but as structural
refinements that address specific limitations—such as convergence guarantees or
global stability—within the same symmetry-based design philosophy.
In particular, the $\SEE$ construction illustrates how state augmentation can be
used to recover almost global asymptotic stability while preserving a linear
error structure suitable for rigorous analysis.

From a practical standpoint, the tutorial emphasized implementation-ready
formulations.
Exact discrete-time propagation on the group, consistent covariance evolution,
and reusable IMU integration structures were derived within a single framework
applicable to both filtering and optimization-based estimators.
Likewise, a broad range of aiding sensors were shown to fit naturally into the
same correction architecture, avoiding ad hoc linearizations and
coordinate-dependent reasoning.

Several important challenges remain open.
In particular, the simultaneous estimation of inertial sensor biases while
preserving almost global convergence guarantees within the observer architectures
discussed here remains a challenge.
Addressing bias estimation within invariant and equivariant frameworks without
sacrificing stability, consistency, or geometric clarity remains an important
direction for future research.

Overall, the intent of this tutorial is not to introduce a new estimator, but to
consolidate a mature body of work into a clear geometric narrative that highlights
the role of symmetry, invariance, and structure in aided inertial navigation.
By making these principles explicit, the tutorial aims to serve both as a
reference for practitioners and as a conceptual bridge between classical inertial
navigation and modern nonlinear observer theory.

\section*{Acknowledgment}
The author would like to thank his collaborators for valuable discussions.
Special thanks are due to Prof.~Abdelhamid Tayebi and Prof.~Tarek Hamel for their
support and insightful feedback during the development of many of the results
on $\SEE$ and synchronous observers presented in this paper.

\section{Author Information}

\begin{IEEEbiography}{{S}oulaimane Berkane}{\,}(soulaimane.berkane@uqo.ca)
received his Ph.D. degree in Electrical Engineering from the University of Western Ontario, Canada, in 2017. He held postdoctoral positions at the University of Western Ontario, Canada, and at KTH Royal Institute of Technology, Sweden, between 2018 and 2019. He is currently an Associate Professor with the Department of Computer Science and Engineering, Université du Québec en Outaouais, Canada, where he leads the Robotics and Autonomous Systems Laboratory (LaRSA). He is a Senior Member of IEEE and a Professional Engineer (P.Eng.) in Ontario. He serves as an Associate Editor for the IEEE Control Systems Society Conference Editorial Board. His research interests include nonlinear control theory, state estimation, and autonomous robotic systems.
\end{IEEEbiography}

\bibliographystyle{ieeetr}    
\bibliography{references}

\begin{thebibliography}{99}

\bibitem{S1}
F.~Bullo and A.~D.~Lewis,
\emph{Geometric Control of Mechanical Systems: Modeling, Analysis, and Design for Simple Mechanical Control Systems},
vol.~49.
Springer, 2005.

\end{thebibliography}

\endarticle


\clearpage
\onecolumn
\pagestyle{plain}

\setcounter{section}{0}
\setcounter{subsection}{0}
\setcounter{equation}{0}
\setcounter{figure}{0}
\setcounter{table}{0}

\renewcommand{\thesection}{S\arabic{section}}
\renewcommand{\thesubsection}{\thesection.\arabic{subsection}}
\renewcommand{\theequation}{S\arabic{equation}}
\renewcommand{\thefigure}{S\arabic{figure}}
\renewcommand{\thetable}{S\arabic{table}}

\begin{center}
{\LARGE\bfseries Supplementary Material}\\[1em]
{\large\bfseries Aided Inertial Navigation on Lie Groups}
\end{center}

\vspace{1em}

\noindent
This supplementary material provides detailed proofs and derivations
supporting the results presented in the main tutorial. It is intended
to complement the technical-report version of the article; the main
tutorial is self-contained and does not rely on this material.

\vspace{1em}

\subsection{Proof of Lemma \ref{lem:per-sample-inc}}
\label{sec:supp-per-sample-increment}

We provide here the derivation of the compact group-level expression stated
in Lemma~\ref{lem:per-sample-inc}. Over the sampling interval
$[t_k,t_{k+1}]$, let
\[
\hat U_k:=\hat U(t_k),
\qquad
\Delta t:=t_{k+1}-t_k,
\]
and assume the zero-order-hold model
$\hat U(t_k+\tau)=\hat U_k$ for
$\tau\in[0,\Delta t]$.

The IMU increment satisfies
\begin{equation}
\dot{\hat M}_{t_k}(\tau)
=
\hat M_{t_k}(\tau)\hat U_k
+
[\hat M_{t_k}(\tau),D],
\qquad
\hat M_{t_k}(0)=I_5 .
\label{eq:supp-M-ode}
\end{equation}
Using
\[
[\hat M_{t_k},D]
=
\hat M_{t_k}D-D\hat M_{t_k},
\]
equation~\eqref{eq:supp-M-ode} may be written as
\begin{equation}
\dot{\hat M}_{t_k}
=
\hat M_{t_k}(\hat U_k+D)
-
D\hat M_{t_k}.
\label{eq:supp-M-ode-expanded}
\end{equation}

Introduce the transformed variable
\begin{equation}
Y(\tau)
:=
\exp(\tau D)\hat M_{t_k}(\tau).
\label{eq:supp-Y-def}
\end{equation}
Differentiating~\eqref{eq:supp-Y-def} and using
\eqref{eq:supp-M-ode-expanded} gives
\begin{align}
\dot Y(\tau)
&=
D\exp(\tau D)\hat M_{t_k}
+
\exp(\tau D)\dot{\hat M}_{t_k}
\nonumber\\
&=
DY
+
Y(\hat U_k+D)
-
D Y
\nonumber\\
&=
Y(\hat U_k+D).
\label{eq:supp-Y-ode}
\end{align}
Since $Y(0)=I_5$, its solution is
\begin{equation}
Y(\tau)
=
\exp\!\big(\tau(\hat U_k+D)\big).
\label{eq:supp-Y-sol}
\end{equation}

For the matrices $\hat U_k$ and $D$ used in the $\SE$ representation of
the INS dynamics, one has
\begin{equation}
D^2=0,
\qquad
D\hat U_k=0.
\label{eq:supp-D-properties}
\end{equation}
Let
\[
A:=\tau\hat U_k,
\qquad
B:=\tau D.
\]
Then $B^2=0$ and $BA=0$. Consequently,
\begin{equation}
(A+B)^n
=
A^n+A^{n-1}B,
\qquad n\geq1,
\label{eq:supp-power-id}
\end{equation}
and hence
\begin{align}
\exp(A+B)
&=
I_5
+
\sum_{n=1}^{\infty}
\frac{A^n+A^{n-1}B}{n!}
\nonumber\\
&=
\exp(A)
+
\left(
\sum_{n=1}^{\infty}
\frac{A^{n-1}}{n!}
\right)B
\nonumber\\
&=
\exp(A)+\mathbf J(A)B.
\label{eq:supp-exp-identity}
\end{align}
Using~\eqref{eq:supp-exp-identity} in~\eqref{eq:supp-Y-sol} yields
\begin{equation}
Y(\tau)
=
\exp(\tau\hat U_k)
+
\tau\,\mathbf J(\tau\hat U_k)D.
\label{eq:supp-Y-expanded}
\end{equation}

Finally, from~\eqref{eq:supp-Y-def},
\[
\hat M_{t_k}(\tau)
=
\exp(-\tau D)Y(\tau).
\]
Since $D^2=0$,
\[
\exp(-\tau D)=I_5-\tau D.
\]
Moreover, $D\hat U_k=0$ implies
\[
D\exp(\tau\hat U_k)=D,
\qquad
D\,\mathbf J(\tau\hat U_k)=D.
\]
Therefore,
\begin{align}
\hat M_{t_k}(\tau)
&=
(I_5-\tau D)
\left[
\exp(\tau\hat U_k)
+
\tau\,\mathbf J(\tau\hat U_k)D
\right]
\nonumber\\
&=
\exp(\tau\hat U_k)
+
\tau\,
\big(
\mathbf J(\tau\hat U_k)-I_5
\big)D .
\label{eq:supp-M-final}
\end{align}
Setting $\tau=\Delta t$ gives
\[
\boxed{
\hat M_{t_k}(\Delta t)
=
\exp\!\big(\Delta t\,\hat U_k\big)
+
\Delta t\,
\Big(
\mathbf J\!\big(\Delta t\,\hat U_k\big)-I_5
\Big)D
}
\]
which proves Lemma~\ref{lem:per-sample-inc}.

\subsection{Lie-Algebra Representation of the Automorphism}
\label{sec:supp-Phi-lie-algebra}

Consider the automorphism
\[
    \Phi_t(X)=\exp(-tD)X\exp(tD).
\]
For $X=\exp(\xi^\wedge)$, the conjugation property of the matrix
exponential gives
\begin{equation}
    \Phi_t\!\big(\exp(\xi^\wedge)\big)
    =
    \exp\!\left(
        \exp(-tD)\xi^\wedge\exp(tD)
    \right).
\end{equation}
Define
\[
    Z(t):=\exp(-tD)\xi^\wedge\exp(tD).
\]
Differentiating yields
\begin{align}
    \dot Z(t)
    &=
    -DZ(t)+Z(t)D \\
    &=
    [Z(t),D].
\end{align}
Since $Z(t)\in\se$, write $Z(t)=z(t)^\wedge$. Defining
$\bar D\in\mathbb{R}^{9\times9}$ through
\[
    (\bar D z)^\wedge=[z^\wedge,D],
\]
the corresponding vector coordinates satisfy
\[
    \dot z(t)=\bar D z(t),
    \qquad
    z(0)=\xi.
\]
Therefore,
\[
    z(t)=\exp(t\bar D)\xi,
\]
and consequently
\[
    \Phi_t\!\big(\exp(\xi^\wedge)\big)
    =
    \exp\!\left(
        \big(\exp(t\bar D)\xi\big)^\wedge
    \right).
\]
Defining $F_t:=\exp(t\bar D)$ gives
\[
    \Phi_t\!\big(\exp(\xi^\wedge)\big)
    =
    \exp\!\big((F_t\xi)^\wedge\big).
\]

\subsection{Proof of Lemma \ref{lem:imu-increment-noise}}
\label{sec:supp-imu-increment-noise}

We provide here the derivation of the first-order IMU-increment noise
perturbation used in Lemma~\ref{lem:imu-increment-noise}.

Consider a bias-free IMU over the sampling interval
$[t_k,t_{k+1}]$, with $\Delta t=t_{k+1}-t_k$. The measured
per-sample increment is
\[
\hat M_k
=
\Gamma\!\left(
\exp(\omega_m^\wedge\Delta t),\,
J_\ell(\omega_m\Delta t)a_m\Delta t,\,
Q_\ell(\omega_m\Delta t)a_m\Delta t^2
\right).
\]
Using the logarithm on $\SE$ recalled in
``\nameref{sidebar-SE23}'', its exponential coordinates are
\begin{equation}
\log(\hat M_k)^\vee
=
\begin{bmatrix}
\omega_m\Delta t\\[2pt]
a_m\Delta t\\[2pt]
J_\ell^{-1}(\omega_m\Delta t)
Q_\ell(\omega_m\Delta t)a_m\Delta t^2
\end{bmatrix}.
\label{eq:supp-log-Mhat}
\end{equation}

Let
$n_k=(n_{\omega,k},n_{a,k})$ denote the discrete IMU noise sample
over the same interval. The corresponding true bias-free inputs are
$\omega_m-n_{\omega,k}$ and $a_m-n_{a,k}$, and hence
\begin{equation}
\log(M_k)^\vee
=
\begin{bmatrix}
(\omega_m-n_{\omega,k})\Delta t\\[2pt]
(a_m-n_{a,k})\Delta t\\[2pt]
J_\ell^{-1}\!\big((\omega_m-n_{\omega,k})\Delta t\big)
Q_\ell\!\big((\omega_m-n_{\omega,k})\Delta t\big)
(a_m-n_{a,k})\Delta t^2
\end{bmatrix}.
\label{eq:supp-log-Mtrue}
\end{equation}

To obtain a compact first-order noise model, the Jacobian factors in the
last block are evaluated at the measured angular increment
$\omega_m\Delta t$. Neglecting their first-order variation with respect to
$n_{\omega,k}$ gives
\begin{equation}
\log(\hat M_k)^\vee-\log(M_k)^\vee
\approx
C_k n_k,
\label{eq:supp-log-difference}
\end{equation}
where
\begin{equation}
C_k
:=
\begin{bmatrix}
\Delta t\,I & 0\\[2pt]
0 & \Delta t\,I\\[2pt]
0 &
J_\ell^{-1}(\omega_m\Delta t)
Q_\ell(\omega_m\Delta t)\Delta t^2
\end{bmatrix}.
\label{eq:supp-Ck}
\end{equation}

We now seek a left-multiplicative perturbation $\eta_k$ satisfying
\begin{equation}
M_k
=
\exp(-\eta_k^\wedge)\hat M_k.
\label{eq:supp-M-perturb}
\end{equation}
Let
\[
\hat m_k:=\log(\hat M_k)^\vee.
\]
Using the first-order perturbation identity
\[
\exp(\epsilon^\wedge)\exp(\hat m_k^\wedge)
\approx
\exp\!\left(
\big(
\hat m_k+\mathcal J_\ell^{-1}(\hat m_k)\epsilon
\big)^\wedge
\right),
\]
with $\epsilon=-\eta_k$, equation~\eqref{eq:supp-M-perturb} gives
\[
\log(M_k)^\vee
\approx
\hat m_k-\mathcal J_\ell^{-1}(\hat m_k)\eta_k.
\]
Consequently,
\begin{equation}
\eta_k
\approx
\mathcal J_\ell(\hat m_k)
\left(
\log(\hat M_k)^\vee-\log(M_k)^\vee
\right).
\label{eq:supp-eta-from-log}
\end{equation}
Combining~\eqref{eq:supp-log-difference} and
\eqref{eq:supp-eta-from-log} yields
\begin{equation}
\eta_k
\approx
\bar L_k n_k,
\label{eq:supp-eta-linear}
\end{equation}
with
\begin{equation}
\boxed{
\bar L_k
=
\mathcal J_\ell\!\big(\log(\hat M_k)^\vee\big)
\begin{bmatrix}
\Delta t\,I & 0\\[2pt]
0 & \Delta t\,I\\[2pt]
0 &
J_\ell^{-1}(\omega_m\Delta t)
Q_\ell(\omega_m\Delta t)\Delta t^2
\end{bmatrix}.
}
\label{eq:supp-Lbar}
\end{equation}

For completeness, if $n_\omega(t)$ and $n_a(t)$ are continuous-time
white processes with
\[
\mathbb E[n_\omega(t)n_\omega(s)^\top]
=
Q_\omega\delta(t-s),
\qquad
\mathbb E[n_a(t)n_a(s)^\top]
=
Q_a\delta(t-s),
\]
their piecewise-constant discrete representations over the sampling
interval satisfy
\begin{equation}
\mathbb E[n_kn_k^\top]
=
Q_k,
\qquad
Q_k
=
\operatorname{diag}\!\left(
\frac{1}{\Delta t}Q_\omega,\,
\frac{1}{\Delta t}Q_a
\right).
\label{eq:supp-Qk}
\end{equation}
Therefore, to first order,
\begin{equation}
\mathbb E[\eta_k\eta_k^\top]
=
\bar L_kQ_k\bar L_k^\top.
\label{eq:supp-eta-cov}
\end{equation}

\subsection{Higher-Order Statistics of the Propagated Invariant Error}
\label{sec:supp-higher-order-propagation}

We now derive the higher-order statistical properties underlying the
bias-free covariance propagation discussed in the main tutorial.

Let the right-invariant state error and the IMU-increment perturbation at
$t_k$ be defined by
\begin{equation}
X_k
=
\exp(-\xi_k^\wedge)\hat X_k,
\qquad
M_k
=
\exp(-\eta_k^\wedge)\hat M_k.
\label{eq:supp-state-increment-errors}
\end{equation}
The exact state propagation is
\[
X_{k+1}
=
G_{\Delta t}\,
\Phi_{\Delta t}(X_k)\,
M_k.
\]
Substituting~\eqref{eq:supp-state-increment-errors} gives
\begin{align}
X_{k+1}
&=
G_{\Delta t}\,
\Phi_{\Delta t}
\!\left(
\exp(-\xi_k^\wedge)\hat X_k
\right)
\exp(-\eta_k^\wedge)\hat M_k
\nonumber\\
&=
G_{\Delta t}\,
\exp\!\big(-(F_{\Delta t}\xi_k)^\wedge\big)
\Phi_{\Delta t}(\hat X_k)
\exp(-\eta_k^\wedge)\hat M_k,
\label{eq:supp-Xnext-1}
\end{align}
where we used
\[
\Phi_{\Delta t}\!\big(\exp(\xi^\wedge)\big)
=
\exp\!\big((F_{\Delta t}\xi)^\wedge\big),
\qquad
F_{\Delta t}
=
\exp(\Delta t\,\bar D).
\]

Using
\[
Y\exp(\zeta^\wedge)
=
\exp\!\big((\Ad_Y\zeta)^\wedge\big)Y,
\]
equation~\eqref{eq:supp-Xnext-1} can be rearranged as
\begin{equation}
X_{k+1}
=
\exp(-\alpha_k^\wedge)
\exp(-\beta_k^\wedge)
\hat X_{k+1},
\label{eq:supp-Xnext-errors}
\end{equation}
where
\begin{align}
\alpha_k
&:=
\Ad_{G_{\Delta t}}F_{\Delta t}\xi_k,
\label{eq:supp-alpha}\\
\beta_k
&:=
\Ad_{G_{\Delta t}}
\Ad_{\Phi_{\Delta t}(\hat X_k)}
\eta_k.
\label{eq:supp-beta}
\end{align}
Since the propagated right-invariant error is defined by
\[
X_{k+1}
=
\exp(-\xi_{k+1}^\wedge)\hat X_{k+1},
\]
we obtain
\begin{equation}
\exp(-\xi_{k+1}^\wedge)
=
\exp(-\alpha_k^\wedge)
\exp(-\beta_k^\wedge).
\label{eq:supp-error-product}
\end{equation}

Applying the Baker--Campbell--Hausdorff expansion to
\eqref{eq:supp-error-product} and expressing the result in vector
coordinates yields
\begin{align}
\xi_{k+1}
={}&
\alpha_k+\beta_k
-\frac{1}{2}\ad_{\alpha_k}\beta_k
\nonumber\\
&+
\frac{1}{12}
\left(
\ad_{\alpha_k}^{2}\beta_k
+
\ad_{\beta_k}^{2}\alpha_k
\right)
\nonumber\\
&+
\frac{1}{24}
\ad_{\beta_k}\ad_{\alpha_k}^{2}\beta_k
+
\mathcal O(5).
\label{eq:supp-BCH}
\end{align}
Here, $\mathcal O(5)$ collects terms of total degree five and higher in
$(\alpha_k,\beta_k)$.

Suppose that $\xi_k$ and $\eta_k$ are mutually independent and zero-mean.
Since $\alpha_k$ and $\beta_k$ are linear transformations of
$\xi_k$ and $\eta_k$, respectively, they are also independent and
zero-mean. Consequently,
\[
\mathbb E[\alpha_k]
=
\mathbb E[\beta_k]
=
0,
\]
and independence implies
\[
\mathbb E[\ad_{\alpha_k}\beta_k]=0,
\qquad
\mathbb E[\ad_{\alpha_k}^{2}\beta_k]=0,
\qquad
\mathbb E[\ad_{\beta_k}^{2}\alpha_k]=0.
\]
Hence all contributions up to third order vanish in expectation, and
\begin{equation}
\mathbb E[\xi_{k+1}]
=
\frac{1}{24}
\mathbb E\!\left[
\ad_{\beta_k}\ad_{\alpha_k}^{2}\beta_k
\right]
+
\mathcal O(5).
\label{eq:supp-error-mean}
\end{equation}
Thus, the invariant error remains zero-mean up to third order, while a
fourth-order mean correction may in general remain.

At first order,
\begin{equation}
\xi_{k+1}
\approx
\alpha_k+\beta_k.
\label{eq:supp-first-order-error}
\end{equation}
The deterministic part satisfies
\begin{equation}
\Ad_{G_{\Delta t}}F_{\Delta t}
=
A_k,
\label{eq:supp-A-identity}
\end{equation}
where $A_k$ is the transition matrix given in
\eqref{eq:Ak_explicit}. Furthermore, using
$\eta_k\approx\bar L_kn_k$ gives
\begin{equation}
\beta_k
\approx
\Ad_{G_{\Delta t}}
\Ad_{\Phi_{\Delta t}(\hat X_k)}
\bar L_k n_k.
\label{eq:supp-beta-noise}
\end{equation}
The matrix
\begin{equation}
L_k^{\mathrm{grp}}
:=
\Ad_{G_{\Delta t}}
\Ad_{\Phi_{\Delta t}(\hat X_k)}
\bar L_k
\label{eq:supp-Lgroup}
\end{equation}
therefore provides the group-consistent first-order injection of the
per-sample IMU-increment perturbation.

The simplified injection matrix $L_k$ used in the main tutorial is
obtained by freezing the continuous perturbation injection over the
sampling interval and retaining its leading-order terms. Accordingly,
\begin{equation}
L_k^{\mathrm{grp}}
=
L_k+\mathcal O(\Delta t^2)
\label{eq:supp-L-comparison}
\end{equation}
under the adopted small-sampling-interval approximation.

Using the first-order relation~\eqref{eq:supp-first-order-error} and the
independence of $\alpha_k$ and $\beta_k$ gives the familiar second-order
covariance propagation
\begin{equation}
\Sigma_{k+1}
\approx
A_k\Sigma_kA_k^\top
+
L_kQ_kL_k^\top.
\label{eq:supp-cov-second-order}
\end{equation}

Retaining the higher-order BCH terms in~\eqref{eq:supp-BCH} produces
additional fourth-order contributions to
$\mathbb E[\xi_{k+1}\xi_{k+1}^\top]$.
Collecting these contributions into the correction matrix
$S_{\mathrm{4th}}$ gives
\begin{equation}
\boxed{
\Sigma_{k+1}
\approx
A_k\Sigma_kA_k^\top
+
L_kQ_kL_k^\top
+
S_{\mathrm{4th}}.
}
\label{eq:supp-cov-fourth-order}
\end{equation}
The explicit fourth-order expressions follow from the BCH covariance
expansion and are discussed in
\cite{barfoot2014associating,brossard2022uncertainty}.
The main tutorial adopts the second-order approximation obtained by
setting $S_{\mathrm{4th}}=0$.

\subsection{Proof of Lemma \ref{lem:group-preint}}
\label{sec:supp-group-preintegration}

\begingroup

Write $h=\Delta t$ and $T_k=(k-i)h$. The proof uses
\[
\Phi_t(XY)=\Phi_t(X)\Phi_t(Y),
\qquad
\Phi_t\circ\Phi_s=\Phi_{t+s},
\qquad
G_t\Phi_t(G_s)=G_{t+s}.
\]
The last identity follows directly from
$G_t=\Gamma(I_3,gt,\tfrac12gt^2)$ and the block action of
$\Phi_t$.

The case $k=i$ holds because $G_0=I_5$, $\Phi_0$ is the
identity map, and $\Delta\hat M_{ii}=I_5$. Assuming
\[
\hat X_k
=
G_{T_k}\Phi_{T_k}(\hat X_i)\Delta\hat M_{ik},
\]
one obtains
\begin{align*}
\hat X_{k+1}
&=G_h\Phi_h(\hat X_k)\hat M_k\\
&=G_h\Phi_h\!\left(
G_{T_k}\Phi_{T_k}(\hat X_i)\Delta\hat M_{ik}
\right)\hat M_k\\
&=G_h\Phi_h(G_{T_k})\,
\Phi_{T_k+h}(\hat X_i)\,
\Phi_h(\Delta\hat M_{ik})\hat M_k\\
&=G_{T_k+h}\Phi_{T_k+h}(\hat X_i)\,
\Delta\hat M_{i(k+1)}\\
&=G_{T_{k+1}}\Phi_{T_{k+1}}(\hat X_i)\,
\Delta\hat M_{i(k+1)}.
\end{align*}
This proves~\eqref{eq:Xj_batch} by induction.

Repeatedly using the same automorphism identities
in~\eqref{eq:recursion:Mij} gives
\begin{equation}
\Delta\hat M_{ij}
=
\prod_{k=i}^{j-1}
\Phi_{(j-1-k)h}(\hat M_k),
\label{eq:supp-DeltaM-product}
\end{equation}
with factors ordered from left to right with increasing $k$.
For example,
\[
\Delta\hat M_{i(i+3)}
=
\Phi_{2h}(\hat M_i)\,
\Phi_h(\hat M_{i+1})\,
\hat M_{i+2}.
\]

All subscripts of $G$ and $\Phi$ denote elapsed physical time.
Each increment is therefore transported over the time from
the end of its sample interval to the terminal keyframe.

\endgroup
\end{document}